\documentclass[times,twocolumn, final]{elsarticle}

\usepackage{framed,multirow}

\usepackage{amssymb}
\usepackage{latexsym}

\usepackage{url}
\usepackage[dvipsnames]{xcolor}
\usepackage{lineno}
\modulolinenumbers[5]

\definecolor{newcolor}{rgb}{.8,.349,.1}

\usepackage{graphicx}
\usepackage{wrapfig}
\usepackage[tableposition=top]{caption}
\usepackage[caption=false]{subfig}
\usepackage{amsmath}
\usepackage{mathtools}
\usepackage{bm}
\usepackage{natbib}
\usepackage{placeins}
\usepackage{algpseudocode}
\usepackage[section]{algorithm} 
\algnewcommand\And{\textbf{and} }
\usepackage{diagbox}
\usepackage{booktabs}
\usepackage[T1]{fontenc}
\usepackage[automake]{glossaries-extra}
\glsdisablehyper
\usepackage{tabularx}
\usepackage{makecell} 
\usepackage{booktabs}
\usepackage{graphics}
\usepackage[export]{adjustbox}

\usepackage[colorlinks]{hyperref}

\newcolumntype{Y}{>{\centering\arraybackslash}X}

\makeglossaries
\setabbreviationstyle{long-short}
\newabbreviation{mic}{MIC}{medical image computing}
\newabbreviation{ml}{ML}{machine learning}
\newabbreviation{mse}{MSE}{mean squared error}
\newabbreviation{dl}{DL}{deep learning}
\newabbreviation{nn}{NN}{neural network}
\newabbreviation{mue}{MuE}{method under evaluation}
\newabbreviation{rl}{RL}{reinforcement learning}
\newabbreviation{qs}{QS}{questioning strategy}
\newabbreviation{vqa}{VQA}{visual question answering}
\newabbreviation{vtt}{VTT}{visual Turing test}
\newabbreviation{dme}{DME}{diabetic macular edema}
\newabbreviation{pomdp}{POMDP}{partially observable Markov decision process}
\newabbreviation{mdp}{MDP}{Markov decision process}
\newabbreviation{pmf}{PMF}{probability mass function}
\newabbreviation{pdf}{PDF}{probability density function}
\newabbreviation{mc}{MC}{Monte Carlo}
\newabbreviation{kl}{KL divergence}{Kullback-Leibler divergence}
\newabbreviation{gp}{GP}{gaussian process}

\DeclarePairedDelimiterX{\infdivx}[2]{(}{)}{%
	#1\;\delimsize\|\;#2%
}
\newcommand{\KLdiv}{D_{KL}\infdivx}
\DeclareMathOperator*{\argmax}{arg\,max}

\def\equationautorefname~#1\null{%
	Eq.~(#1)\null
}
\newcommand{\eg}{e.\,g.,\ } 
\newcommand{\ie}{i.\,e.,\ }	

\graphicspath{{figures/}}

\usepackage{geometry}
\begin{document}


\begin{frontmatter}

\title{A reinforcement learning approach for VQA validation: an application to diabetic macular edema grading }
\tnotetext[abbr]{Abbreviations: \textbf{QS:} Questioning Strategy, 
	\textbf{MuE:} Method under Evaluation} 
\author{Tatiana Fountoukidou}\corref{cor1}
\cortext[cor1]{Corresponding author}
\ead{tatiana.fountoukidou@gmail.com}
\author{Raphael Sznitman}
\address{Artificial Intelligence in Medical Imaging, ARTORG Center, University of Bern, Murtenstrasse 50, 3008 Bern, Switzerland}

\begin{abstract}
Recent advances in machine learning models have greatly increased the performance of automated methods in medical image analysis. However, the internal functioning of such models is largely hidden, which hinders their integration in clinical practice. Explainability and trust are viewed as important aspects of modern methods, for the latter's widespread use in clinical communities. As such, validation of machine learning models represents an important aspect and yet, most methods are only validated in a limited way. In this work, we focus on providing a richer and more appropriate validation approach for highly powerful Visual Question Answering (VQA) algorithms. To better understand the performance of these methods, which answer arbitrary questions related to images, this work focuses on an automatic visual Turing test (VTT). That is, we propose an automatic adaptive questioning method, that aims to expose the reasoning behavior of a VQA algorithm. Specifically, we introduce a reinforcement learning (RL) agent that observes the history of previously asked questions, and uses it to select the next question to pose. We demonstrate our approach in the context of evaluating algorithms that automatically answer questions related to diabetic macular edema (DME) grading. The experiments show that such an agent has similar behavior to a clinician, whereby asking questions that are relevant to key clinical concepts. 
\end{abstract}


\end{frontmatter}

\section{Introduction}
\begin{figure*}[!t]
	\centering
\subfloat[\label{fig:VQA}]{\includegraphics[height=0.19\textheight]{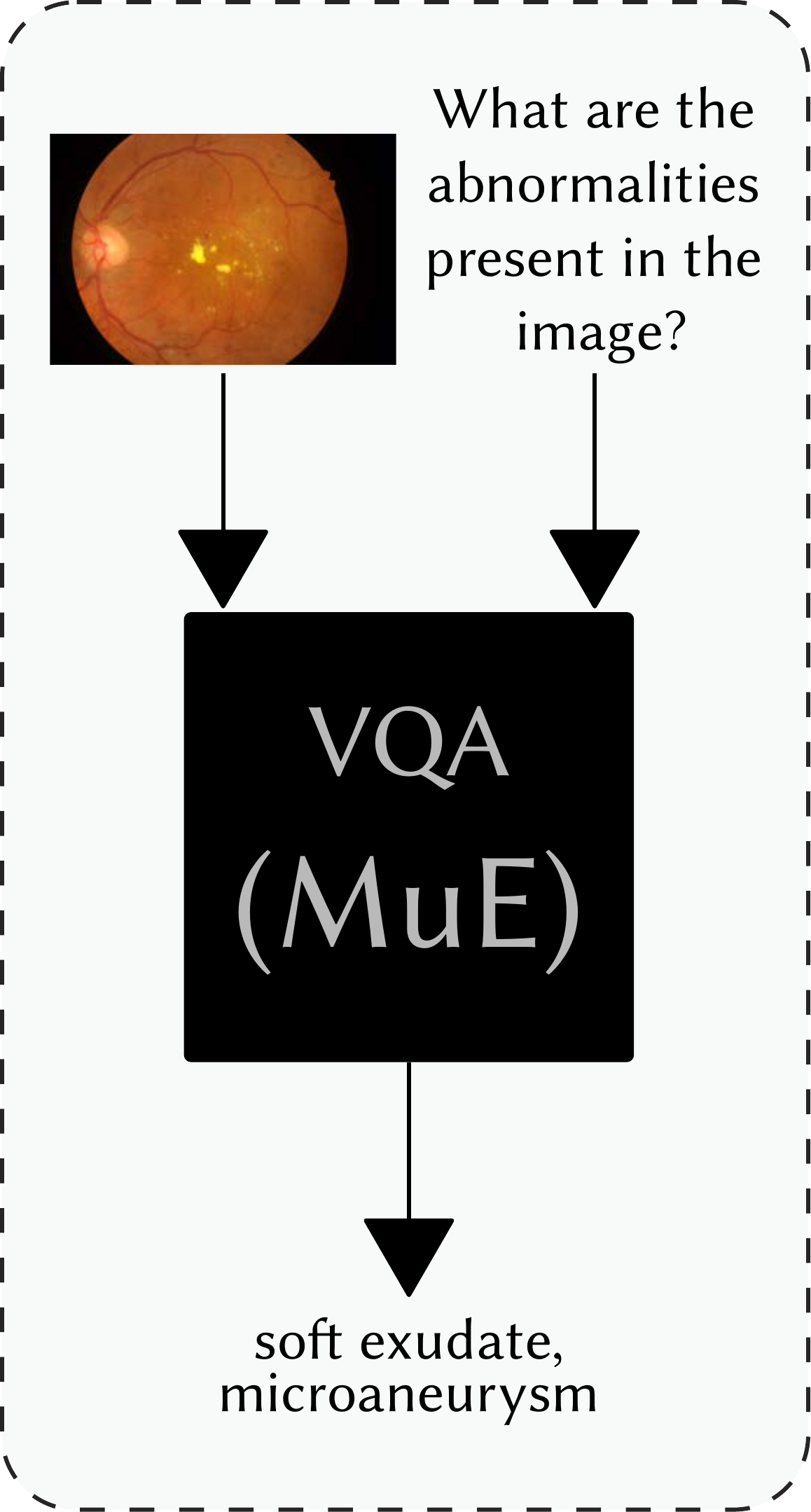}}
\qquad
\subfloat[\label{fig:concept_VTT}]{\includegraphics[height=0.19\textheight]{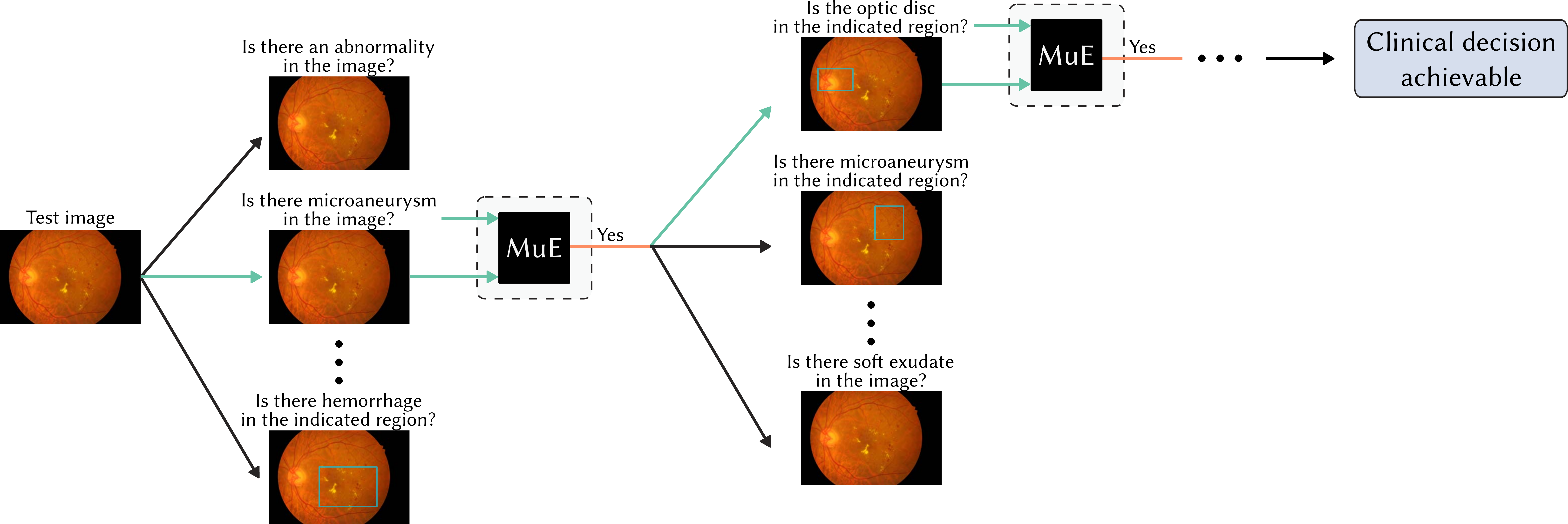}}
	\caption{(a) Example of MuE's inputs and outputs. (b) Conceptual illustration of a visual Turing test (VTT)  for fundus screening. The VTT selects the questions to  pose (green arrows) from all possible questions. Answers provided by the MuE are shown in orange. Questioning continues until a clinical decision (\eg diagnosis) can be made. A MuE that correctly answers questions leading to a clinical decision resembles a trained clinician and is therefore more trustworthy.}
\label{fig:VTT}
\end{figure*}
Recent advances in computer vision and medical image analysis have shown remarkable performances for numerous diagnostic and intervention applications. With the emergence of large neural networks, or \gls{dl}, tasks that were long considered extremely challenging are now performed with human-level skill. 

At the same time, comprehensive validation is increasingly critical as these methods move on towards translation into clinical practice. Yet, as methods have become increasingly powerful, the overall methodology to validate them has largely remained intact. For instance, many challenge competitions~\citep{sun2020multi, allan20192017} compare different methods on a common dataset, using metrics most often borrowed from the computer vision literature. Such competitions have been criticized, as final rankings and outcomes are very often highly skewed to the dataset or metrics used~\citep{MeierHein2019}. 

More generally, good performances, expressed by high metric values, are desirable but not enough to trust a system in healthcare. There are a number of other qualitative factors hidden from strictly quantitative metrics, such as when a system would fail and what are its limitations (\eg exposed by adversarial attacks,~\cite{papernot2016limitations}), what evidence is used to infer decisions, or how well does the method really understand its input (\ie interpretability and explainability). Given the growing complexity of \gls{dl} methods, the answers to the above are especially difficult, if not impossible, to determine. This is one reason why clinical adoption of new machine learning based methodology remains challenging.

For this reason, this work focuses on {\it evaluating and assessing} how a trained \gls{ml} model (in particular one trained to answer questions related to images) makes its decisions.  
Specifically, we consider that this \gls{ml} model has been trained  to answer questions related to images, linked to a specific medical task, and that we do not know anything about its internal structure, how it was trained or what data was used to train it. Our goal is to design a method that can assess how this trained model, or \emph{\gls{mue}}, is able to reason when evaluating unseen test data. Here, we refer to ``reasoning'' as the ability of the \gls{mue} to correctly answer questions relevant to the specified medical task. In this context, we propose here an evaluation method that provides insight that goes beyond common evaluation metrics (\eg accuracy, precision, etc). As such, our method would not directly be used to tackle a clinical task but rather help in choose which methods should be used in clinical practice.

\subsection{Related works}
Enhancing explanatory power has gathered strong recent interest from the medical imaging community. \Gls{vqa}~\citep{Antol_2015_ICCV,ImageCLEFVQA-Med2018,lau2018dataset,wu2017visual} is one category of models that provide enhanced explainability. In the typical \gls{vqa} setting a model takes as input both a test image and a question in text form and must predict the correct answer to the question posed (\autoref{fig:VQA}). Questions could be open-ended (\ie demand for a free text answer), or closed-ended (\ie ``Yes/No'' answer). The assumption is that a model's capability to answer questions with respect to an image or set of images can make it easier to reveal its inner functioning. This makes \gls{vqa} particularly attractive for medical applications, since explainability is a necessary asset for a \gls{ml} model to be integrated in healthcare. There are several remarkable attempts to tackle the task of \gls{vqa} for medical datasets, proposing different combinations and merging techniques for image and text processing~\citep{vu2020question,ren2020cgmvqa,lubna2019mobvqa,gupta2021hierarchical,zhan2020medical,lin2021medical,tascon2022consistency}. Here again, however, current metrics used to evaluate \gls{vqa} methods remain inadequate (\ie accuracy of correctly answered questions or BLEU scores). It is such \gls{vqa} models that play the role of \gls{mue}s in this work, and it is their trustworthiness that we aim to assess.

In the early 1950's, Alan Turing devised a test to assess if a machine exhibits human like behavior~\citep{machinery1950computing}. In a ``Turing test'', an interrogator questions a responder by asking sequential questions, in order to discern if the responder is human or a machine.
Turing tests have been used in medical imaging to evaluate the quality of adversarial attacks, by seeing if an expert can distinguish between a real and an adversarial example~\citep{chuquicusma2018fool,schlegl2019f}. Another use of the Turing test paradigm was to assess the interpretability of methods that infer semantic information from images (\ie classification, segmentation etc.). In~\cite{geman2015visual}, an automated \gls{vtt} algorithm adaptively selects images and questions to pose to a model such that the answers can not be predicted from the history of answers. While this approach has increased explanatory power, it is limited to manually fabricated story lines to guide questioning. This makes it ill suited for medical applications where such story lines are hard to formalize. In~\cite{fountoukidou2019concept}, questions are posed to a \gls{mue}, with the aim to examine whether a concept of interest exists in an entire image. The answers for all images are then used to update a \gls{gp} that reveals the biases of both the answers and the dataset. The \gls{gp} is consequently used to indicate the subset of questions from which the next question should be sampled. While this method helps reduce uncertainty with fewer questions, it does not guarantee that the chosen questions are appropriate to assess the \gls{mue}'s reasoning. This is due to questions designed to reduce the uncertainty over the entire dataset, rather than exposing the different elements of an image that play a role in a MuE output. Since the question selection criteria are not related to any specific medical task, the \gls{mue}'s reasoning over complex medical decisions can not be assessed. In this work we opt for a finer level of detail, selecting questions related to each specific image in order to expose the \gls{mue}'s reasoning.
\subsection{Contributions}
Following the \gls{vtt} of \cite{geman2015visual},  the present work proposes an automatic, learned interrogator that sequentially selects questions to pose to a black-box \gls{mue}. We achieve this by training a \gls{rl} agent to act as the interrogator that selects questions (see \autoref{fig:concept_VTT} for an illustration). That is, we do not focus on how to train or optimize a specific \gls{mue}, that responds to questions, but rather on training an evaluation method that can effectively assess whether the answers produced by the \gls{mue} reveal a desirable behavior. Once trained, our interrogator does not simply indicate a static sequence of questions, but adapts and dynamically chooses from a pool of questions, based on the task context. The \gls{rl} agent is therefore able to produce an arbitrary number of question sequences, depending on its interaction with the \gls{mue}.

In this work, we focus on a specific diagnostic application for our \gls{vtt}: \gls{dme} grading. Using prior knowledge of this task, we construct a set of questions that are clinically relevant and train our \gls{rl} agent to select questions that are necessary and adequate to perform this medical task. The question selection is dynamic, as they depend on both the image and the answers from the \gls{mue}. We show in our experiments that the trained RL agent questions a \gls{mue} in much the same way as a clinician would for this task. To the best of our knowledge this work is the first to learn such an interrogator with the objective of focusing on core concepts related to a medical task. The contributions of this work are thus the following:
\begin{enumerate}
	\item The proposal of a pipeline to train a questioning strategy for method validation, that is able to simulate the decision process of an expert and thus enhance a model's evaluation of explainability.
	\item Development of novel and appropriate evaluation metrics to assess questioning strategies.
\end{enumerate}
It should be noted that this problem is very hard to solve in a universal way, since the very notion of understanding a topic cannot be disconnected from its particularities. The specific solution we propose here may need to be adjusted for a different clinical application\footnote{In particular, the set of questions should be designed for the medical task, and depending on the later's complexity the question selection network might need to be altered.}, but the general idea, training pipeline, and evaluation metrics for a \gls{vtt} questioning strategy hold.

The remainder of the paper is organized as follows: in the next section we provide a detailed description of our method and the \gls{rl} method we propose. In \autoref{sec:exp}, we detail our experimental setup and report results in \autoref{sec:results}. We conclude with final remarks in \autoref{sec:conc}.

\section{Automated visual Turing test}
\label{sec:method}


Posing the right questions is an important step in learning and understanding a given topic. This is one reason why \gls{vqa} models can link understanding to the questions and associated answers they evaluate at test time. However the development of a \gls{vqa} model focuses primarily on the answering part, while the questions are fixed and given beforehand. We claim that the question posing component, however, plays a tremendous role in the \gls{vqa} evaluation. In order to define whether a responder (in our case a \gls{mue}) correctly understands a topic, it does not suffice that they correctly answer some questions. Which questions they answer correctly is critical, as there are questions that are more appropriate than others to expose a ``cheating'' or incompetent responder. Our hypothesis is that a set of appropriate questions has more discriminative power than a larger set of all possible questions, a hypothesis that is confirmed by our experiments and results. 

To this end, this work focuses not on how to answer questions, but on \emph{how to choose which questions to ask}.We do not devise a \gls{vqa} model that answers questions, we propose a dynamic way to evaluate such models, treated as black-box responders, in an insightful way. In a clinical scenario, a medical task is usually linked to questions, that serve as intermediate steps to solve the specific task. That is, clinicians look at an image, and consecutively look for elements in it that allow them to make a diagnosis. Depending on what elements they observe, they adjust their focus on what to look for next (they do not perform an exhaustive search of all possible elements a medical image contains). For instance, to grade \gls{dme} severity, a number of elements must be observed in a patient's fundus photograph. Thus, the ability to answer the clinical question of what is the \gls{dme} grade of an eye depends directly on answering questions about these visible elements. Such elements represent concepts that clinicians are experts in interpreting. \gls{vqa} algorithms can therefore be compared more in detail to a clinician. However, not all questions are equally important for any given medical case. Also, the complete number of all possible questions could be overwhelming, thus hindering an insightful \gls{vqa} evaluation. Trained clinicians adjust their reasoning process to the pathology they are dealing with, but also to the medical history of the patient, the imaging modality they are observing, what they have already observed in the image before etc. 

In this work, we assume that the questions the experts choose to pose are the most relevant for a clinical task, and therefore the most appropriate to judge if such a task is well understood. We therefore aim to devise a \emph{\gls{qs}} that would ``simulate'' an expert, such that the \gls{qs} can be used to select which questions should posed to a \gls{mue}, in order to expose if the \gls{mue} is reliable. For example, a \gls{mue} that can correctly infer the \gls{dme} grade from fundus images, but fails to answer correctly regarding the aforementioned concepts, would not be considered trustworthy, despite its potentially high performance according to standard metrics.

Assuming a \gls{vqa} model trained on answering questions about fundus images, our \gls{qs} selects which question will be asked first. The \gls{vqa}, or MuE, provides a response that is fed back to the \gls{qs}, which selects the next question to be posed, and so on. The questioning stops when the sequence of question-response pairs for a specific image provides enough information for this image to be graded. The same process can be repeated for several images in the same way a clinician screens images from several patients. We formalize our approach below.

\subsection{Problem formulation}
We first define the set of all possible text questions relevant to a clinical task, $\mathcal{A}$\footnote{Note that we choose to use the symbol $\mathcal{A}$ for the question set, as the questions will consist of actions following the reinforcement learning convention.}, of size $N$.
We specify the subset 
$\mathcal{A}_{\text{asked},t}\subseteq\mathcal{A}$ as the set of asked questions at time $t$. We assume that $\mathcal{A}$ contains closed-ended (``Yes/No'') questions indicated by an expert or explicitly defined by a medical textbook. Note that we limit $\mathcal{A}$ to closed-ended questions so that datasets and methods that are not specifically designed for question answering can be evaluated too. Also,  most open-ended questions can be reformulated into a series of ``Yes/No'' questions. 

For each available test image, we further specify a set of $N_L$ locations, or image regions $\mathcal{L} = \{ l_i ~|~ i=0, \dots, N_L-1\}$, and a set of $N_C$ clinically relevant concepts $\mathcal{C} = \{ c_i ~|~ i=0, \dots, N_C-1\}$. Let
\begin{equation}
\mathcal{A} = \mathcal{L} \times \mathcal{C} = \{a_c^l ~|~ c \in \mathcal{C}, l\in \mathcal{L} \},
\label{eq:questionset}
\end{equation}
be the set of all possible questions, where $a_c^l$ is of the form \emph{"Is concept $c$ present in region $l$?"}.

We denote the response of the black box \gls{mue} (\ie the \gls{vqa} model) to question $a_c^l$, as $r_c^l \in \{\text{``N/A'', ``No'', ``Yes''} \}$\footnote{Meaning, the \gls{vqa} model that serves as our \gls{mue} gets a fundus image and $a_c^l$ as input, and gives $r_c^l$ as output.}, and let ${z: \{\text{``N/A'', ``No'', ``Yes''} \} \rightarrow \{0, 0.5, 1\}}$ be a mapping of the form,
\begin{equation}
z(r) =
\begin{cases}
0, & r=\text{``N/A'' (the question is not asked)} \\
0.5, & r=\text{``No'' (the response is ``No'')} \\
1, & r=\text{``Yes'' (the response is ``Yes'')} \\
\end{cases}.
\end{equation} 

For clarity we refer to $a_c^l$ and $r_c^l$ at time $t$ as $a_t$ and $r_t$ respectively. We then let $(a_i, r_i)$ be a question-response pair and
\begin{equation}
\label{eq:history}
\bm{H}_t = \left \{ (a_0, r_0), (a_1, r_1), \dots ,(a_t, r_t) \right\}, \quad \bm{H}_t \in \mathcal{H}, 
\end{equation}
be the history of question-response pairs at time $t$, where $\mathcal{H}$ is the space of all possible history sequences. 
We then relate the question set $\mathcal{A}$ to the $z$ mapping by defining a transformation $\phi: \mathcal{H} \rightarrow \{0, 0.5, 1\}^{N_C \times N_L}$, where 
\begin{equation}
\phi(\bm{H}_t) =
\begin{bmatrix}
z(r^0_0) & \dots & z(r^{N_L}_0)\\
\vdots & \ddots & \vdots\\
z(r^0_{N_C}) & \dots & z(r^{N_L}_{N_C})\\
\end{bmatrix}.
\label{eq:fv_history}
\end{equation} 
Note that the order of questions in $\bm{H}_t$ is not preserved in $\phi(\bm{H}_t)$ and that we assume that an expert can confirm if a specific instance of $\mathcal{H}$ is adequate to assess the clinical task (\ie enough information for a clinical decision to be made).

\begin{figure}[!t]
	\centering
	\includegraphics[width=0.5\textwidth]{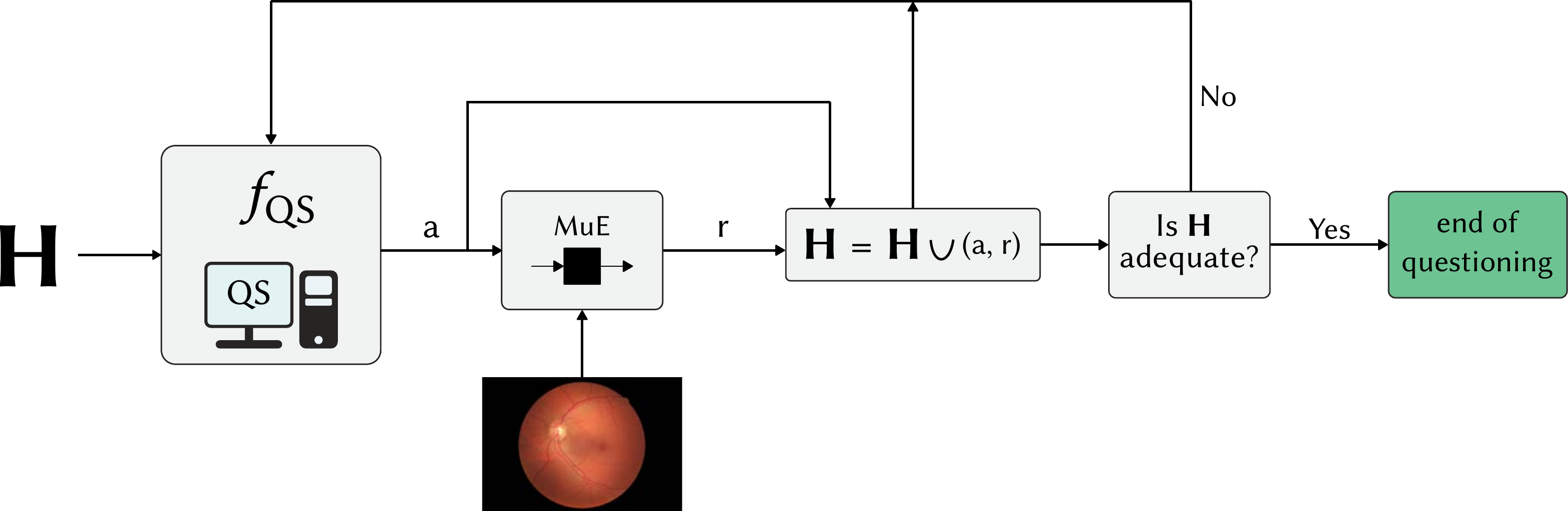}
	\caption{Overview of questioning process for a single image.}
	\label{fig:qs_overview}
\end{figure}

We are interested in generating questioning strategies that use a minimum number of questions necessary to ascertain a clinically relevant task. We thus define a \glsentrylong{qs} as a function $f_\text{QS}: \mathcal{H} \rightarrow \mathcal{A}$, that given a history, selects which question should be posed next. An overview of the questioning process is presented in \autoref{fig:qs_overview}. 
In the next section, we describe our \glsentrylong{rl} method and how we train an agent to select questions with the aim of maximizing a scalar reward. 

\subsection{Questioning strategy generation}
\label{sec:RL}
Our goal is to learn a function that can adaptively select questions to pose to a black-box \gls{mue}, and 
we aim to do so such that our method yields questions that are ``reasonable''  with respect to the task the \gls{mue} is attempting to perform. To this end, we use a \glsentrylong{rl} (RL) approach~\citep{sutton2018reinforcement}, as our aim is to outline a decision process that maximizes a reward which we define below.

\gls{rl} is a branch of \gls{ml}, with the particularity of \emph{learning from interactions}. An \emph{agent} has to learn what \emph{actions} to take, by interacting with an \emph{environment}, and observing how different actions affect it.
Specifically, we propose to utilize \gls{rl} to construct an agent to model the questioning function $f_{QS}$. That is, we treat the \gls{qs} as an \emph{agent} that selects \emph{actions} (\ie questions to pose to the \gls{mue}) at every step of the \gls{mue} evaluation. The \emph{environment} is the way the \gls{mue} perceives and understands the medical image.
Based on the history $\bm{H}_t$, the agent selects the question that is likely to lead to a clinical decision sooner (\ie via less questions), whereby simulating a doctor's reasoning process. The agent-environment interaction is represented by the question posing, the observation of the \gls{mue} response and a produced \emph{reward} signal. This observation helps update the agent's view of the environment, meaning the agent's state, and therefore affects the selection of the next action (\ie question). In our setup, we establish episodic tasks that are completed in a finite number of steps. Terminal states are those where sufficient questions are asked for a clinical diagnosis to be established. Such states are specific to the medical task, and can be inferred by published medical criteria (as is the case in our \gls{dme} application), or indicated by experts. Since our goal is to ask the questions that are relevant for diagnosis, the reward is defined so as to encourage questions in that direction. The \gls{rl} elements that are used in our approach are described below.

\subsubsection{State, actions and observations}
We specify the state that the agent observes at timestep $t$ to be $s_t = \phi(\bm{H}_t)$. That is, the state reflects the history of the questions posed by the \gls{qs} (interrogator) and the responses given by the \gls{mue} (responder) for a given image. The action corresponds to the question $a_t$, that is posed to the \gls{mue}. The agent's observation is the answer, $r_t$, that the \gls{mue} provides, which is treated as stochastic in nature. The agent then uses this observation to update its internal state, by updating all the values related to the given question-response pair. 

\subsubsection{Reward}
We define the immediate reward after a transition from state $s$ to $s'$ following action $a$ as
\begin{equation}
R^a_{s, s'} = 
\begin{cases}
0 & \parbox[t]{.3\textwidth}{if $s'$ is not terminal and $a$ has never been chosen}\\
1 & \parbox[t]{.3\textwidth}{if $s'$ is terminal and $a$ has never been chosen}\\
-1 & \text{if $a$ has been chosen before}
\end{cases}.
\end{equation}
We then compute the discounted reward for an episode as
\begin{equation}
G_{\text{episode}} = \sum_{t=0}^{t=T-1}R^{a_t}_{s_t, s_{t+1}}\cdot\gamma^{t},
\label{eq:episode_reward}
\end{equation}
where $T$ is the length of the episode and $\gamma \in (0,1)$ is a discount factor, such that higher values of $\gamma$ emphasize future reward.

\subsubsection{Action-value function approximation}
As in most \gls{rl} settings, the action-value function, $Q(s, a)$, estimates the discounted future reward when being in state $s$, after taking action $a$, and the policy $\pi(s,a) = p(a_t{=}a | s_t{=}s)$ is a probability function that maps state-action pairs to probabilities. Our goal then is to learn an optimal policy, which maximizes $Q(s, a)$,
\begin{equation}
Q_*(s,a) \dot{=} \max_{\pi} Q_{\pi}(s,a), \quad \forall s \in \mathcal{S}, a \in \mathcal{A}.
\end{equation}
An agent following an optimal policy $\pi_*$ is our desired \gls{qs}. 
For a given $Q(s,a),$ a greedy policy is defined as
\begin{equation}
\pi(s, a) = 
\begin{cases}
1 & \text{if } a = \argmax\limits_{a} Q(s,a)\\
0 & \text{if } a \neq \argmax\limits_{a} Q(s,a)
\end{cases},
\end{equation}
and a policy that is greedy with respect to an optimal value function $Q_*(s,a)$ is an optimal policy. Our task therefore consists in finding $Q_*(s,a)$. 

\begin{figure}[t]
	\centering
	\includegraphics[width=.48\textwidth]{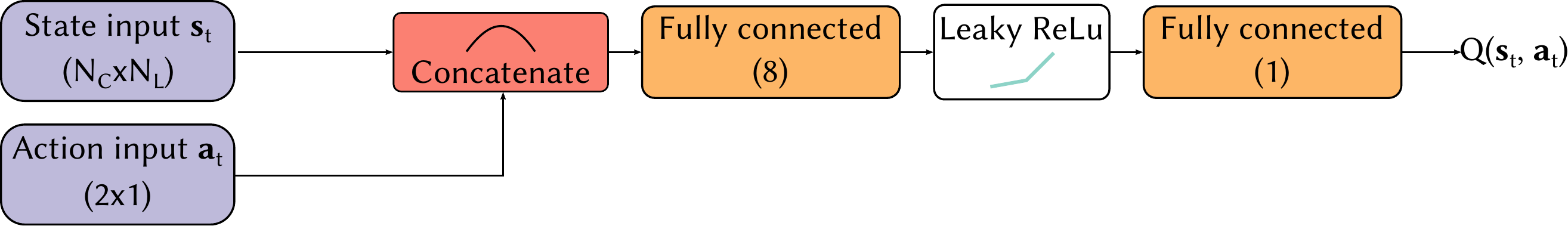}
	\caption{Function approximation network. The numbers in parentheses refer to the layer size.}
	\label{fig:FAmodel}
\end{figure}

As in recent trends, we model $Q(s, a)$ with a parameterized function $\tilde{Q}(s,a, \bm\theta)$ and model this using a \gls{nn}, $\tilde{Q}(\cdot)$. \autoref{fig:FAmodel} depicts the architecture of our proposed network. 

\subsubsection{Masked policy}
Intuitively one can assume that there is no value in asking the same question twice. However, this creates a complexity in the learning process of the agent, since exactly the same action can yield very different rewards (\ie high if it has not been chosen before and assists diagnosis, and low if it has been chosen before, regardless of its importance for diagnosis, low if in this state of the questioning there is no new information added to the agent's state). We tackle this issue explicitly by imposing a very low return to actions that have already affected the agent's state, whereby forcing the policy to assign them zero probability. Formally, we impose $\pi(s,a_t)=0 \text{ if } a_t\in\mathcal{A}_{\text{asked}, t-1}$ as in \cite{kucur2019patient}.

\subsubsection{Training}
To learn an optimal approximation of $Q_*$, we utilize the \gls{mse} to measure the distance between $\tilde{Q}(s,a, \bm\theta)$ and $Q_*(s,a)$,
\begin{equation}
\mathcal{L}_{\bm{\theta}} = \mathbb{E}_{a \sim \mathcal{A}, s \sim \mathcal{S}}\left[\left( \tilde{Q}(s,a, \bm\theta) - Q_*(s,a) \right)^2\right].
\end{equation}

In addition, we propose two different training schemes in this work: (1) \gls{mc} learning and (2) Q-learning~\citep{watkins1992q}.
 We show the performance and behavior of both training schemes in our results section.

In \gls{mc} learning, a batch update is performed after an entire episode is finished. The update sample batch consists of the state-action pair at each episode step, and the target for each such pair is the discounted reward from that step onward. The \gls{mse} loss is computed as
\begin{align}
\mathcal{L}_{\text{MC}}  &= \mathbb{E}_{\mathcal{E}}\left[\mathcal{L}_{\text{MC}}^{\mathcal{E}}\right]  \nonumber\\ 
&= \mathbb{E}_{\mathcal{E}}\left[\frac{1}{T}\sum_{s_t, a_t \in \mathcal{E}}\left(G_t - \tilde{Q}(s_t,a_t, \bm\theta)\right)^2\right],
\end{align}
where $\mathcal{E}$ is an episode of length $T$, and $\mathcal{L}_{\text{MC}}^{\mathcal{E}}$ is the average loss calculated for a single episode.

Conversely, in Q-learning an update is performed after every episode step (in our case, after every question), with the update target being the sum of the observed immediate reward $R_{s,s'}^a$ and the maximum predicted discounted reward from then on. 
We also use {\it experience replay} with a replay memory~\citep{lin1993reinforcement, mnih2013playing}, where we store the agent's experiences at every step in a replay memory $\mathcal{M} = \{e_0, \dots, e_{N_{\text{replay}}-1}\}$ of size $N_{\text{replay}}$, with $e_i = (s_t, a_t, R_{s_t,s_{t+1}}^a, s_{t+1})$ being an experience tuple. When the time for the update comes, instead of updating with the step that was just taken, we sample a minibatch of size $N_b$ from the replay memory, $(e_{i,0}, \dots, e_{i,N_b-1}) \sim \mathcal{M}$. This way, the update batch has less chances to include samples with strong correlations. The lack of such strong correlations is desirable because it reduces the variance of the updates. In addition, each experience tuple can be used in several updates, therefore making better use of the dataset. The \gls{mse} loss for this case is
\begin{equation}
\mathcal{L}_{\text{Q}} = \mathbb{E}_{s,a, R, s'\sim \mathcal{M}} \left[\left(R + \gamma\max_{a'} \tilde{Q}(s',a', \bm\theta) - \tilde{Q}(s,a, \bm\theta)\right)^2 \right],
\end{equation}
and we make use of an $\epsilon-$greedy policy to generate episodes,
\begin{equation}
\pi(s_t, a_t) = 
\begin{cases}
\frac{\epsilon}{|\mathcal{A}|} + 1-\epsilon & \text{if } a_t = \argmax\limits_{a} Q(s_t, a)\\
\frac{\epsilon}{|\mathcal{A}|} & \text{if } a_t \neq \argmax\limits_{a} Q(s_t, a)\\
\end{cases}
.
\label{eq:egreedy}
\end{equation}
It can be easily noticed that the greedy policy is a special case of $\epsilon$-greedy with $\epsilon=0$. We use an $\epsilon$ decay scheme, whereby training starts with $\epsilon=1$ and progressively reduces by a factor of $\epsilon_\text{decay}$. This scheme allows for early exploration when the agent is unaware of the environment, and progressively moves in exploring more the high reward state-action pairs as the agent's confidence increases. The questioning stops when a terminal state is reached, or a maximum number of questions is asked (\eg 20 questions in our experiments).

Algorithms~\ref{alg:MCtrain} and ~\ref{alg:Qltrain} in the Supplementary Material (\ref{app:RL_training_algs}) outline the \gls{mc} learning and Q-learning with function approximation, respectively.

\subsection{Questioning strategy evaluation}
\label{sec:res_evaluation}
In this setting, quantifying how well questions are posed is not obvious and we propose a number of ways to do so. In general, we are interested in how quickly a \gls{qs} can reach a clinically adequate state and how well a \gls{qs} can differentiate \gls{mue}s exhibiting different behaviors. 

\paragraph{\textbf{Episode rewards}}
First, we consider the achieved episode reward on a number of testing images as an indication of the \gls{qs} ability to ask proper questions.
 
\paragraph{\textbf{MuE separation}}
We also wish to see how questioning strategies can be used to differentiate \gls{mue}s. Specifically, we are interested in examining whether a good \gls{qs} can differentiate between \gls{mue}s with the same overall performance. That is, we wish to promote appropriate \gls{qs}s that can ask relevant questions, instead of just asking all possible questions. Several \gls{mue}s may have the same accuracy over all the questions, but some of those \gls{mue}s are more appropriate than others for clinical integration. This is yet another reason why asking all possible questions might not be enlightening enough when evaluating a \gls{mue}. To identify the more reliable \gls{mue}s, we look at the rate of correct answers for each \gls{qs}. The overall performance of a \gls{mue} is based on the correctly answered questions on the entire question set $\mathcal{A}$. Note however that the \gls{mue} performance induced by a \gls{qs} depends on the questions $\mathcal{A}_{\text{asked}}$ that the strategy posed. Since the different \gls{mue}s represent different reasoning behaviors, we would like to identify a \gls{qs} that can separate \gls{mue}s despite a similar average accuracy.

\paragraph{\textbf{MuE accuracy approximation with beta distribution}}
To examine a questioning strategy's ability to distinguish between \gls{mue}s, we treat every question-response pair produced by a \gls{qs}-\gls{mue} pair as a Bernoulli trial. 
Over a test set then, the probability mass function of these trials can be computed as,
\begin{equation}
P_{\text{Bernoulli}}(k, p) = p^k(1-p)^{1-k}, \quad \text{for } k \in \{0,1\},~p \in [0,1],
\end{equation}
where $k=1$ if the \gls{mue} answers correctly, and $k=0$ otherwise.

We then approximate the accuracy achieved by each \gls{mue} with a beta distribution, which we update through Bayesian inference. This involves computing,
\begin{equation}
p_{\text{beta}}(x, \alpha, \beta) = \frac{x^{\alpha-1}(1-x)^{\beta-1}}{B(\alpha, \beta)},
\end{equation}
where $\alpha$ and $\beta$ are the beta distribution parameters and $B(\cdot, \cdot)$ is a normalizing factor. Note that we can interpret the numbers $\alpha-1$ and $\beta-1$ as the number of successes and failures of an experiment, respectively. Given that the beta distribution is the conjugate prior of the Bernoulli, 
and that the beta distribution describes a distribution over probabilities, we can model the \gls{mue} performance as perceived by a \gls{qs} by computing,
\begin{align}
p_{\text{qs}}^{u}(x, \alpha_{\text{qs}}^{u}, \beta_{\text{qs}}^{u}) = \frac{x^{\alpha_{\text{qs}}^{u}-1}(1-x)^{\beta_{\text{qs}}^{u}-1}}{B(\alpha_{\text{qs}}^u, \beta_{\text{qs}}^u)}
\nonumber \\ 
\quad \forall \text{qs} \in \mathcal{I} \text{ and } \forall u \in \mathcal{U},
\label{eq:beta_mue_qs}
\end{align}
where $\mathcal{I}$ is the set of all questioning strategies (interrogators), and $\mathcal{U}$ is the set of all \gls{mue}s. 
We initialize each one of the $p_{\text{qs}}^u$ with an uninformative prior (\ie $\alpha=\beta=1$), which gives a uniform distribution in the interval $[0,1]$, and we update the parameters after every observed question-response. After the questioning is over, we end up with one beta distribution per \gls{qs} per \gls{mue}, that characterizes the performance of each \gls{mue} as perceived by each \gls{qs}. 

Note that because each \gls{qs} asks a different number of questions, we can anticipate lower variance beta distributions for the strategies that ask more questions, as these will undergo a greater number of Bayesian updates. 
To avoid this bias, we set a limit $N_u$ (for each \gls{mue}) in the number of questions that are used to define parameters $\alpha$ and $\beta$. We then ask $N_u$ questions from each \gls{qs}, and we update the corresponding beta distributions. $N_u$ is set to the number of questions of the most effective \gls{qs}.

\paragraph{\textbf{Information radius}}
To quantify the ability of a \gls{qs} to differentiate between responding \gls{mue}s, we use the information radius measure proposed in~\cite{sibson1969information}. The information radius is a symmetric measure of separation, or dissimilarity coefficient, between distributions. It is inspired by the \gls{kl}, but is symmetric and generalizes to more than two distributions. 
Assuming $N$ distributions with probabilities, $p_i, \forall i \in {1, ..., N}$ defined in the same probability space $\mathcal{X}$, the information radius is calculated as,
\begin{equation}
R = \frac{1}{N}\sum_{i=1}^{N} \KLdiv*{p_i}{\frac{\sum_{j=1}^{N} p_j}{N}}.
\label{eq:info_radius}
\end{equation}
It is therefore the average KL divergence from the mean distribution to each \gls{mue} distribution. 
 Note that it is always finite and bound by $N\text{log}K$, where $\frac{1}{K}$ is the probability at each point in $\mathcal{X}$, if $p$ is uniform. Combining \autoref{eq:beta_mue_qs} and \autoref{eq:info_radius}, we thus compute the information radius for a questioning strategy given a set of \gls{mue}s $\mathcal{U}$ as
\begin{equation}
R_\text{qs} = \frac{1}{|\mathcal{U}|}\sum_{u \in \mathcal{U}} \KLdiv*{p_{\text{qs}}^u}{\frac{\sum_{u \in \mathcal{U}} p_{\text{qs}}^u}{|\mathcal{U}|}}, \quad \forall \text{qs} \in \mathcal{I}.
\label{eq:info_radius_qs}
\end{equation} 
\section{Experimental setup}
\label{sec:exp}
We describe below an overview of the experimental setup we use to validate our approach. Specifically, we propose to validate our method for the task of \gls{dme} grading, where we first describe the data we use in our experiments and then detail a number of comparison methods.
\subsection{DME grading and datasets}
\label{subsec:dme_grading}
\begin{figure}[!t]
	\centering
	\subfloat[\small{DME grade 0}]{\includegraphics[width=0.15\textwidth]{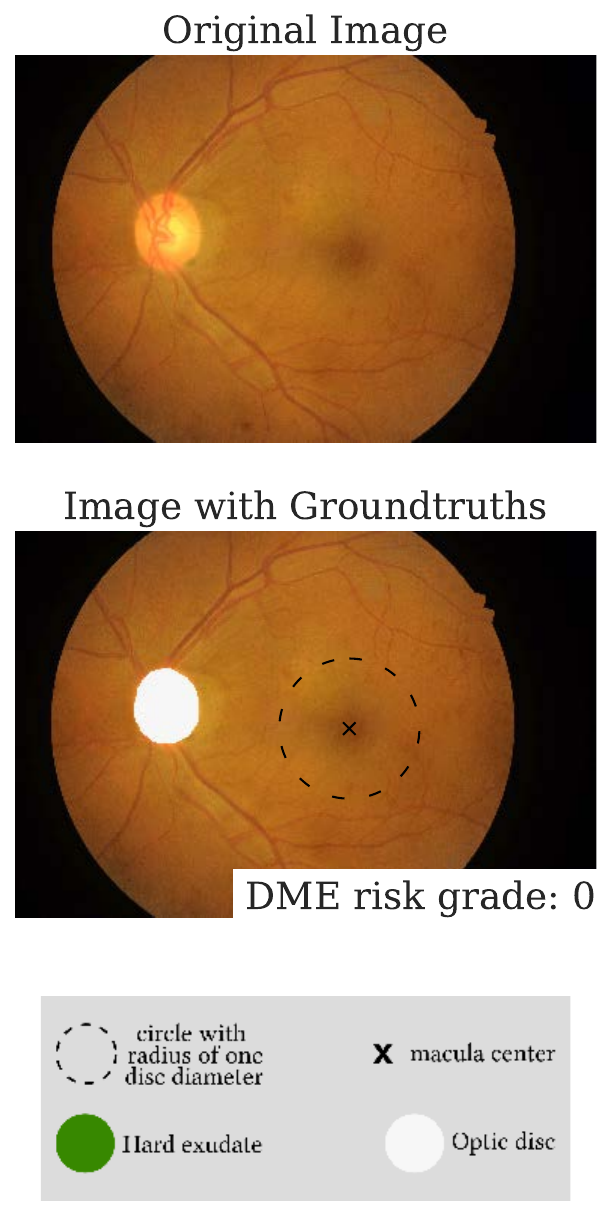}}
	~
	\subfloat[\small{DME grade 1}]{\includegraphics[width=0.15\textwidth]{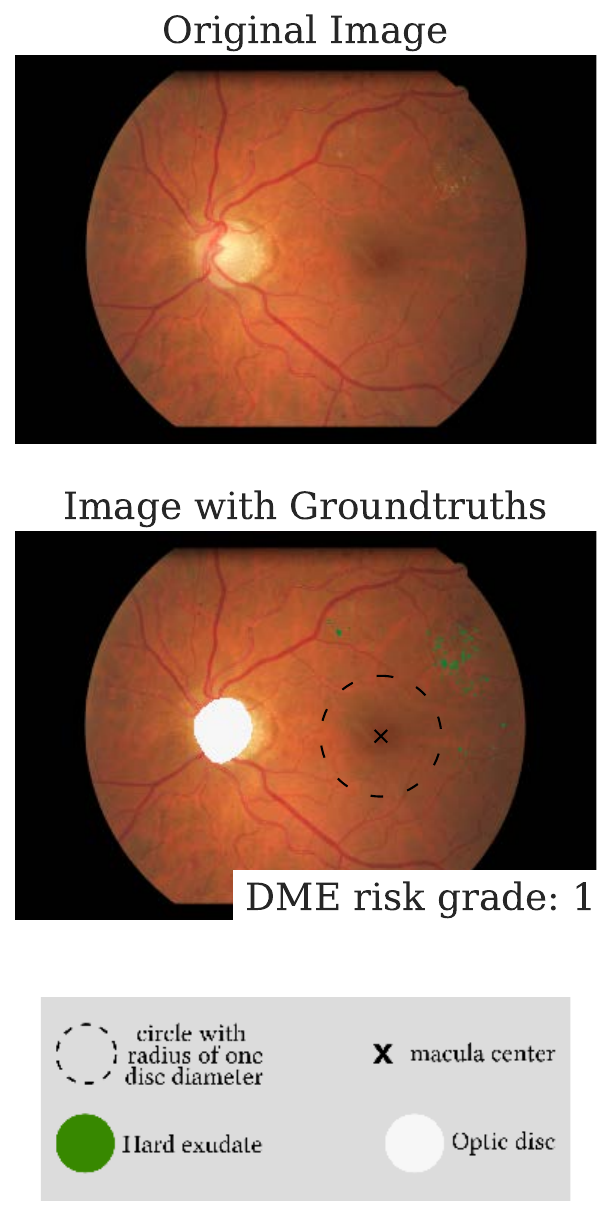}}
	~
	\subfloat[\small{DME grade 2}]{\includegraphics[width=0.15\textwidth]{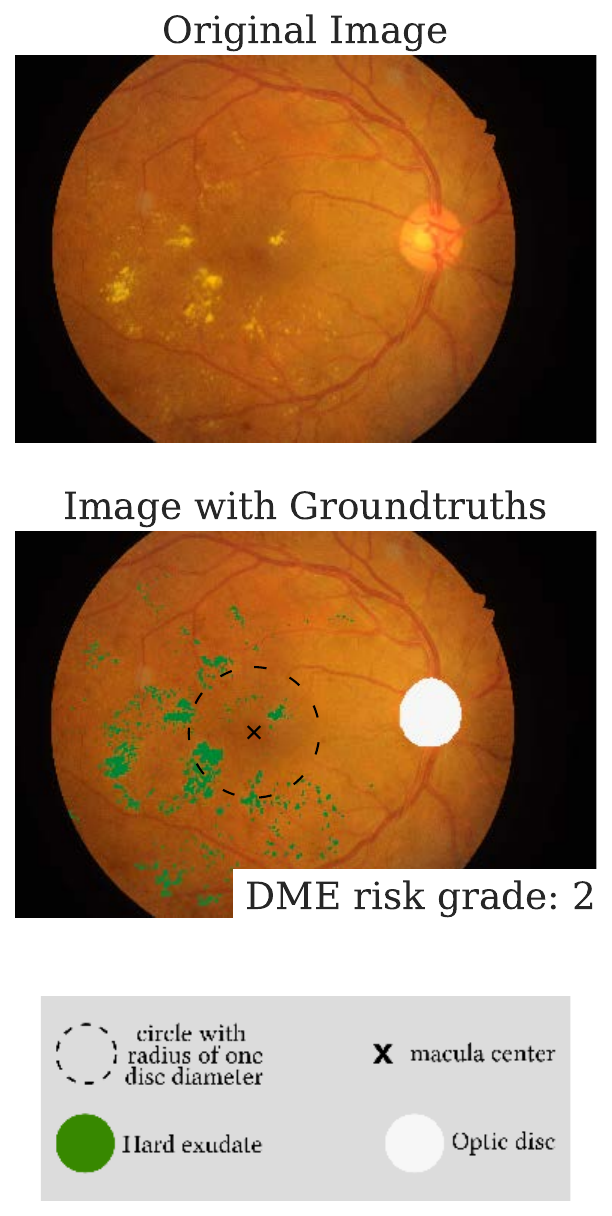}}
	\caption{{Examples of fundus images with groundtruths of lesions, anatomical structures and DME risk gradings.}}
	\label{fig:DME_examples}
\end{figure}

\Gls{dme} is the build-up of fluid in the macula of the retina. This fluid increase leads to  blurry or wavy vision, near or in the center of the visual field~\citep{Bandello17}. Color fundus photography (see ~\autoref{fig:DME_examples}) plays a key role in diagnosing and assessing the risk levels associated with the condition. 
One important type of retinal lesions visible in color fundus images is called {\it hard exudate} and is well known to be linked to the disease. To assess the risk level of \gls{dme}, the following guidelines of the ETDRS grading scale~\citep{early1991fundus} are typically used:
\begin{description}
	\item{\textbf{Grade 0}:} No apparent hard exudates,
	\item{\textbf{Grade 1}:} Presence of hard exudates outside the radius of one disc diameter from the macula center (fovea),
	\item{\textbf{Grade 2}:} Presence of hard exudates within the radius of one disc diameter from the macula center (fovea).	
\end{description}

Despite the apparent simplicity of this task, any automated method that grades for \gls{dme} must implicitly be able to: (1) classify and identify hard exudate lesions, (2) localize the fovea, (3) segment the optic disk and (4) compare relative distances and sizes between the fovea and hard exudates. We hence consider a \gls{mue} that can correctly answer questions on the aforementioned points more explainable and trustworthy.

Given that our goal is to train a questioning strategy that can assess if black-box trained \gls{mue}s are performing \gls{dme} grading according to the correct clinical reasoning,  we propose to build a dedicated \gls{vqa} dataset for our experiments. We do so because, despite the existence of a few medical \gls{vqa} datasets~\citep{lau2018dataset, he2020pathvqa}, they are not appropriate, as we require questions and responses that relate to a specific medical task, where the medical outcome is also known. For this reason, we use datasets that were not initially designed for \gls{vqa}, but contain all necessary annotations for our purpose.
 
To this end, we make use of two different color fundus photograph datasets\footnote{For each dataset, we use a subset of images that contain  annotations relating to hard exudates and this precludes us from using fundus datasets with only \gls{dme} annotations, such as MESSIDOR~\citep{decenciere2014feedback}.} and summarize the number of images per grade in \autoref{table:dataset} :

{\bf Indian Diabetic Retinopathy image Dataset (IDRiD)~\citep{h25w98-18}:} {148 color fundus images} from both healthy and diabetic retinopathy subjects, with optic disc and hard exudate segmentation masks. Fovea localization was manually performed. The dataset is split in a 60\%-10\%-30\% training, validation and test set respectively. 

{\bf eOphtha Dataset~\citep{decenciere2013teleophta}:} {62 color fundus images} from both healthy and diabetic retinopathy subjects, with hard exudate segmentation masks. Optic disc segmentation and fovea localization were manually performed. The dataset is split in a 60\%-10\%-30\% training, validation and test set respectively. 
\begin{table}[!t]
	\small
	\begin{center}
		\caption{\label{table:dataset}Number and percentage of images per grade in the datasets.}
		\begin{tabular}{lccc}
			\toprule
			& \textbf{IDRiD} & \textbf{e-Ophtha} & \textbf{Total} \\
			\midrule
			\textbf{DME grade 0} & 71 (48\%) & 23 (37\%) & 93 (44\%) \\  
			\textbf{DME grade 1} & 3 (2\%) & 10 (16\%) & 13 (6\%) \\
			\textbf{DME grade 2} & 74 (50\%) & 29 (47\%) & 105 (50\%) \\
			\bottomrule
		\end{tabular}
	\end{center}
\end{table}

The training, validation, and test sets of the two datasets are respectively concatenated, and used in the \gls{qs} generation and evaluation.


\subsection{Question set}
\label{subsec:question_set}
\begin{figure}[!t]
	\centering
	\subfloat[\small{Split of image in quadrants.}\label{fig:quadrants}]{\includegraphics[width=0.21\textwidth]{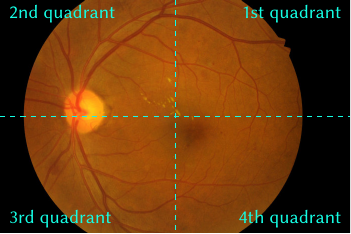}}
	\quad
	\subfloat[\small{Example of concepts.}\label{fig:concepts}]{\includegraphics[width=0.21\textwidth]{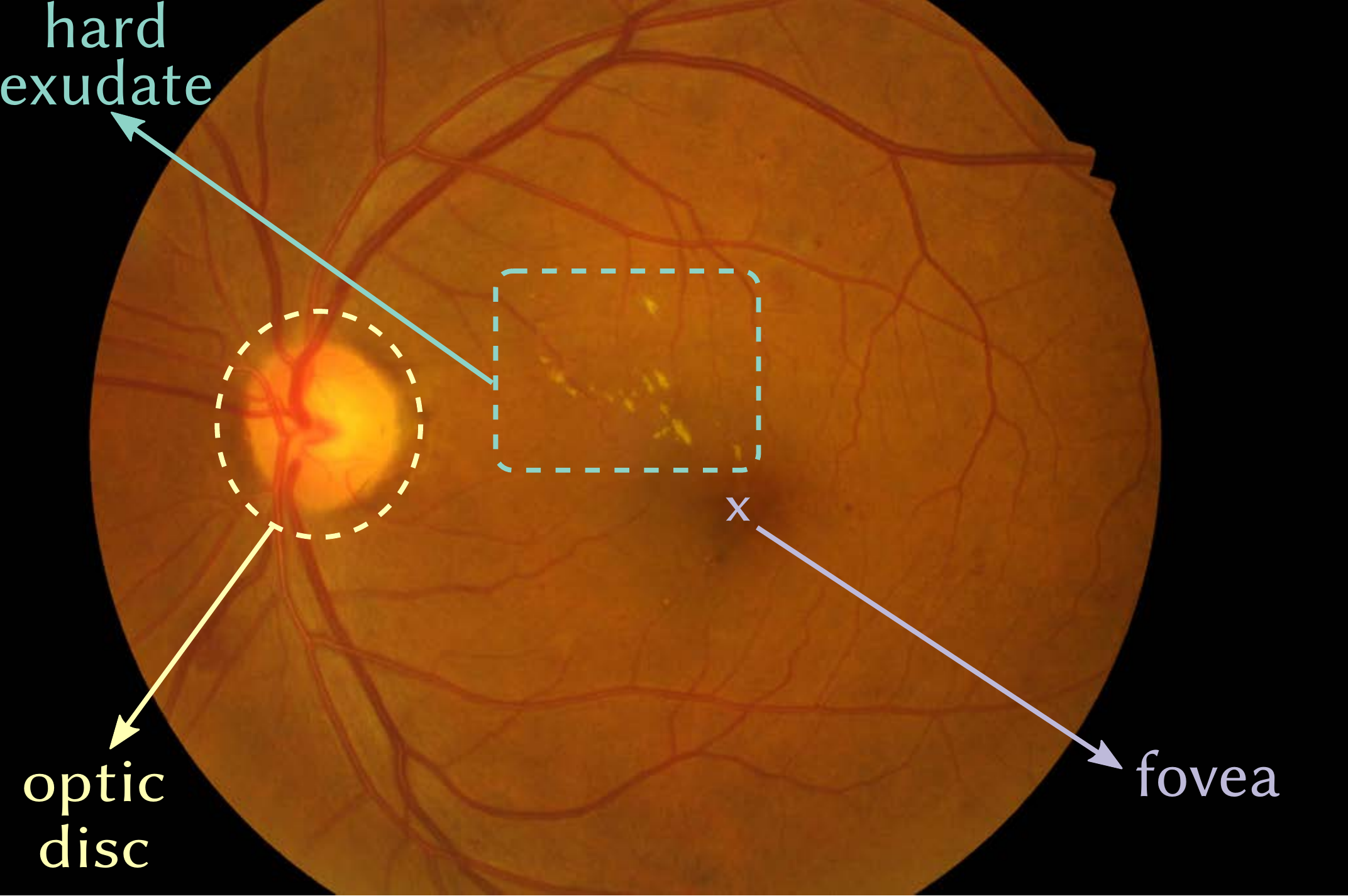}}
	\caption{{Regions and concepts used in questioning.}}
	\label{fig:questionSet}
\end{figure}

We now specify the set of questions $\mathcal{A}$ as defined in \autoref{eq:questionset} for the task of DME grading. We let, 
\begin{align}
\mathcal{L} =  &\{\text{whole image, 1st quadrant, 2nd quandrant,} \nonumber\\
&\text{3rd quadrant, 4th quadrant}\}, \nonumber
\end{align} 
such that a fundus image is divided into 4 non-overlapping quadrants. This division is made so that different image regions can be questioned separately, allowing not only the presence but also the localization of a concept to be determined. That is, we treat the quadrant division as a proxy for the localization of a structure in a closed-ended questioning setup\footnote{We chose to divide the image to quadrants because they are relevant for DME grading, and fundus image inspection.  Extending to finer or different grids for applications that require it is trivial.}. 
We also define the concepts to test by the set of clinically relevant structures, 
$$\mathcal{C} = \{\text{hard exudate, fovea, optic disc}\}.$$
Thus the entire question set is $\mathcal{A} = \mathcal{L} \times \mathcal{C}$ and we depict examples of the regions and concepts \autoref{fig:questionSet}.

\subsection{Generating clinically relevant question streams}
\label{subsec:question_set}
Using our question set $\mathcal{A}$, we are now able to generate sets of closed-ended questions for any image by querying different image locations and structures of interest. That is, we can simulate streams of questions, with some being appropriate for diagnosis inference (\ie corresponding to terminal states). In \autoref{subsec:QSs}, we also outline different questioning strategies that produce different such streams. Specifically, we can simulate questioning strategies that pose similar questions to that of an expert. 

To do so, we make the following assumptions: (1) if the optic disc is localized, its size is assumed to be known, (2) localization of the optic disc is a necessary condition for diagnosis if exudates are present, (3) if the fovea is in the same quadrant as hard exudates, their distance is assumed to be lower than one optic disc diameter and (4) the fovea and the optic disc need to be localized.
\label{assumptions}
Although (3) does not hold for the entire dataset, we confirmed that it is valid for 96\% of the images. While (4) is not strictly necessary for diagnosis, we add it as an extension, since a \gls{mue} that properly understands a fundus image should be able to identify the optic disc and the fovea.  

We henceforth refer to the use of assumptions (1-3) as {\bf simple-A}, and to the use of assumptions (1-4) as {\bf extra-U-A} (from extra-Understanding). In the Supplementary Material (\ref{app:DME_textbook_trees}) we provide an illustration for the \gls{dme} grading decision process described in~\autoref{subsec:dme_grading}, for both {\bf simple-A}  and {\bf extra-U-A} in the form of decision trees, as well as give detailed statistics and examples as to the validity of these assumptions.

\subsection{Questioning strategies}\label{subsec:QSs}
We now describe the different questioning strategies we will compare in our experiments. These include:
\begin{description}
	\item[Random QS:] The next question posed is randomly chosen from the set of \emph{not asked questions} $\mathcal{A} \setminus \mathcal{A}_{\text{asked}} $. Note that this reflects the most common form of \gls{mue} evaluation found in the literature.
	\item[Textbook QS:] Considered our gold standard as it follows a clinical-reasoning and approximates clinical thinking, as described in~\autoref{subsec:dme_grading}. \autoref{fig:DME_textbookTrees} in the Supplementary Material shows two such questioning strategies.
	\item[Decision Tree QS, Random Budget (DT-RB):] This strategy is generated by traversing a classification tree that is trained to perform the \gls{dme} grading task. Specifically, we train the classification tree to correctly assess the grade of an image using a limited budget of randomly selected history sequences (see \autoref{eq:history}). Each node of the trained tree corresponds to a question, and each edge to a response. We then use the tree splits to select the next question that should be posed (see Supplementary Material,~\ref{app:decision_tree_learning} for more details).
	\item[Decision Tree QS Textbook Budget (DT-TB):] Similar to the above but where the classification tree is trained on a budget of history sequences that are selected for each image according to the textbook criteria described in \autoref{subsec:dme_grading}.
	\item[Reinforcement learning QS (RL QS):] Our proposed strategy as described in \autoref{sec:RL}. We show results when our method is trained using \gls{mc} learning, denoted \textbf{(MC)}, and when trained with Q-learning, denoted as \textbf{(Q)}.
\end{description}

We train our RL methods with an Adam optimizer \citep{kingma2014adam}. The discount factor was set to $\gamma=0.8$, so as to emphasize the final reward but to achieve it as quickly as possible. We start by generating one episode per training image with a random policy ($\epsilon$-greedy with $\epsilon=1$), and we reduce the value of $\epsilon$ by a factor of $\epsilon_{\text{decay}}=0.9$ after every epoch. We run one episode per training image for each epoch. For Q-learning, we make use of a replay memory of size 500 and we update with minibatches of size 8. We train for 50 epochs and we keep the model with the best validation reward after the 15th epoch.

\section{Results}
\label{sec:results}

We show the results of our experiments in this section and we provide additional results in the Supplementary Material. 

\subsection{Preliminaries}
To first establish the difficulty in inferring coherent DME grades, we evaluate the \gls{dme} classification performance using standard tree classifiers.
 For the scope of this work, we train classification trees as an intermediate step to generate a questioning strategy, and to show that good classification results do not necessarily go hand in hand with relevant clinical criteria. As the \gls{dme} grading task is not the focus of this work, we do not use a powerful or complex training scheme, but one that can easily be used to generate a \gls{qs}.

To this end, we train 50 trees to predict \gls{dme} grades from history instances, $\bm{H}$ (\autoref{eq:history}), and report the classification accuracy in the training, validation and test set in \autoref{table:decision_tree_results}. We see that such a tree-based classifier performs well when looking at usual classification metrics. However, as we show below, this is not an indication that they can successfully be used to generate a good \gls{qs}.

\begin{table}[!t]
	\small
	\begin{center}
		\caption{\label{table:decision_tree_results}\small Mean classification accuracy of decision tree classifier trained with a random budget ({\bf DT-RB}) and a textbook budget ({\bf DT-TB}) of history sequences. Results are averaged over 50 tries with $\mu \pm \sigma$ provided.}
		\label{table:tree_classification_results}
		\begin{tabular}{l c c c }
			\toprule
			& \textbf{Training set} & \textbf{Validation set} & \textbf{Testing set}  \\
			\midrule
			\emph{simple-A} &&& \\
			{\bf DT-RB} & $0.99 \pm 0.003$  & $0.92 \pm 0.024$ & $0.91 \pm 0.030$  \\  
			{\bf DT-TB} & $0.99 \pm 0.005$ & $0.93 \pm 0.015$ & $0.92 \pm 0.029$  \\
			\emph{extra-U-A} &&& \\
			{\bf DT-RB}  & $0.99 \pm 0.003$  & $0.92 \pm 0.022$ & $0.90 \pm 0.034$  \\  
			{\bf DT-TB} & $0.99 \pm 0.006$ & $0.93 \pm 0.015$ & $0.92 \pm 0.030$  \\
			\bottomrule
		\end{tabular}
	\end{center}
\end{table}
\begin{figure*}[!h]
	\centering
	\subfloat[\small{No terminal states in training (simple-A).}\label{fig:val_reward_diagnosis_ntt}]
	{\includegraphics[width=.32\textwidth]{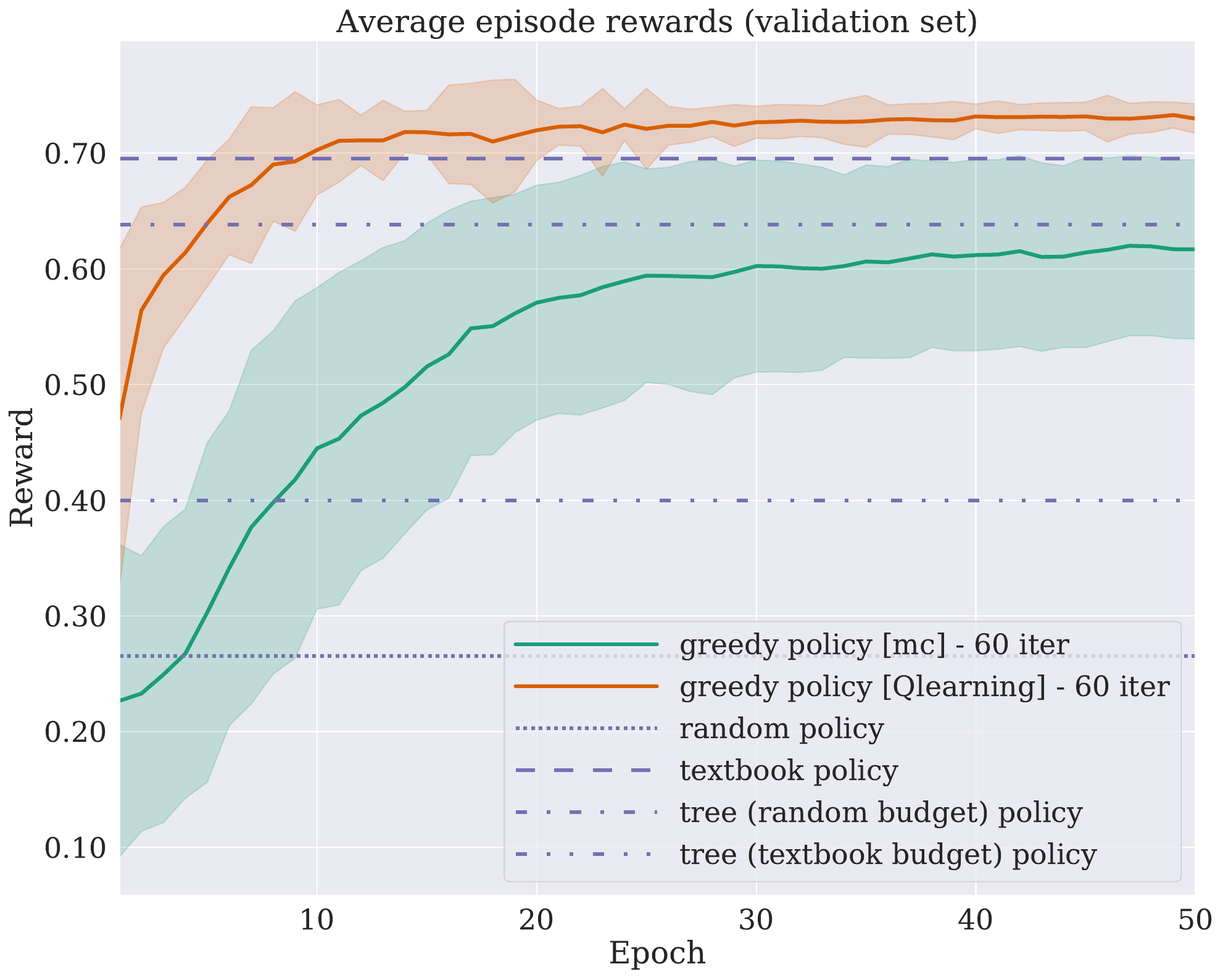}}
	\quad
	\subfloat[\small{Terminal states in training (simple-A).}\label{fig:val_reward_diagnosis_tt}]
	{\includegraphics[width=.32\textwidth]{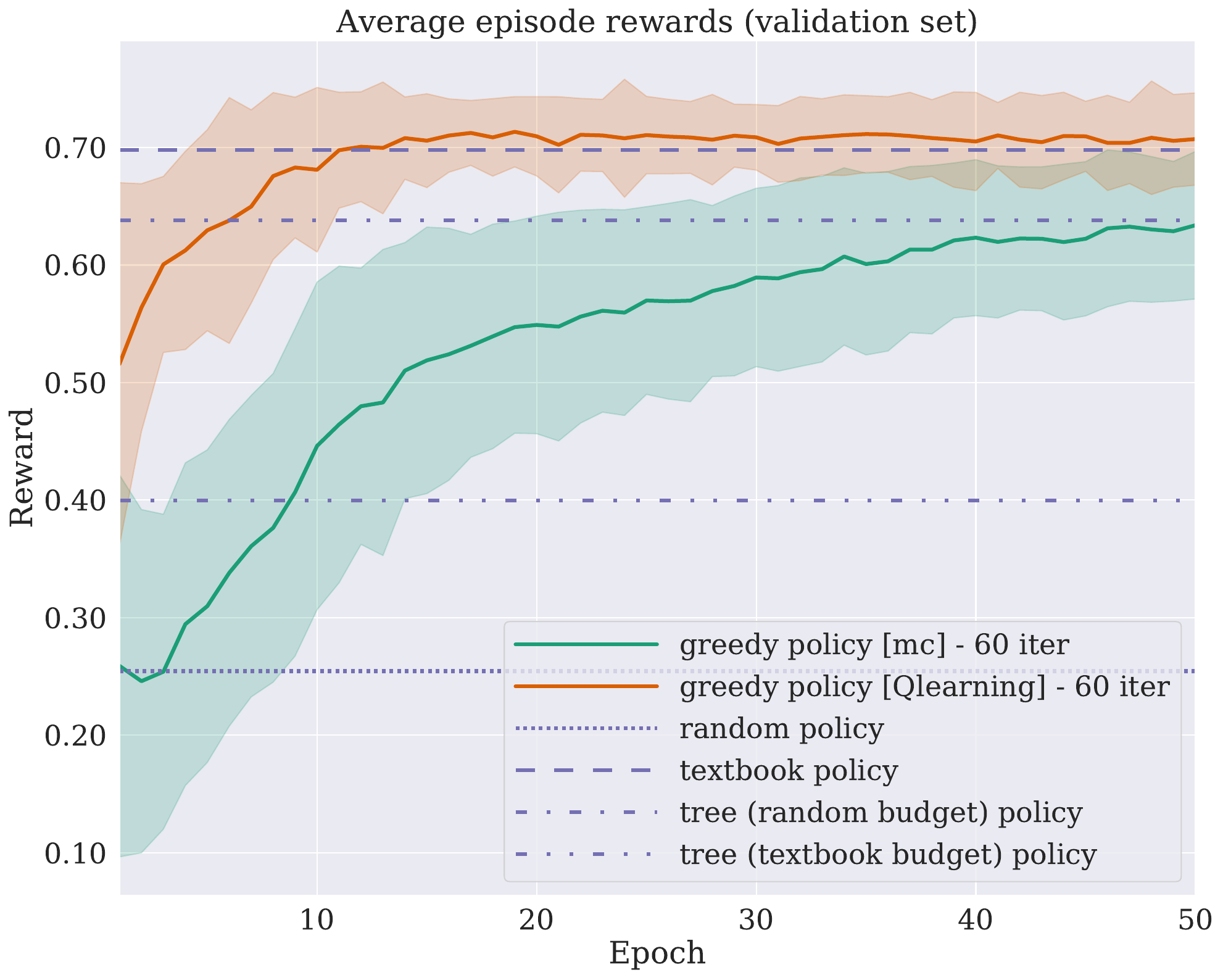}}
	\quad
	\subfloat[\small{\ref{fig:val_reward_diagnosis_ntt} and \ref{fig:val_reward_diagnosis_tt} combined}.]
	{\includegraphics[width=.32\textwidth]{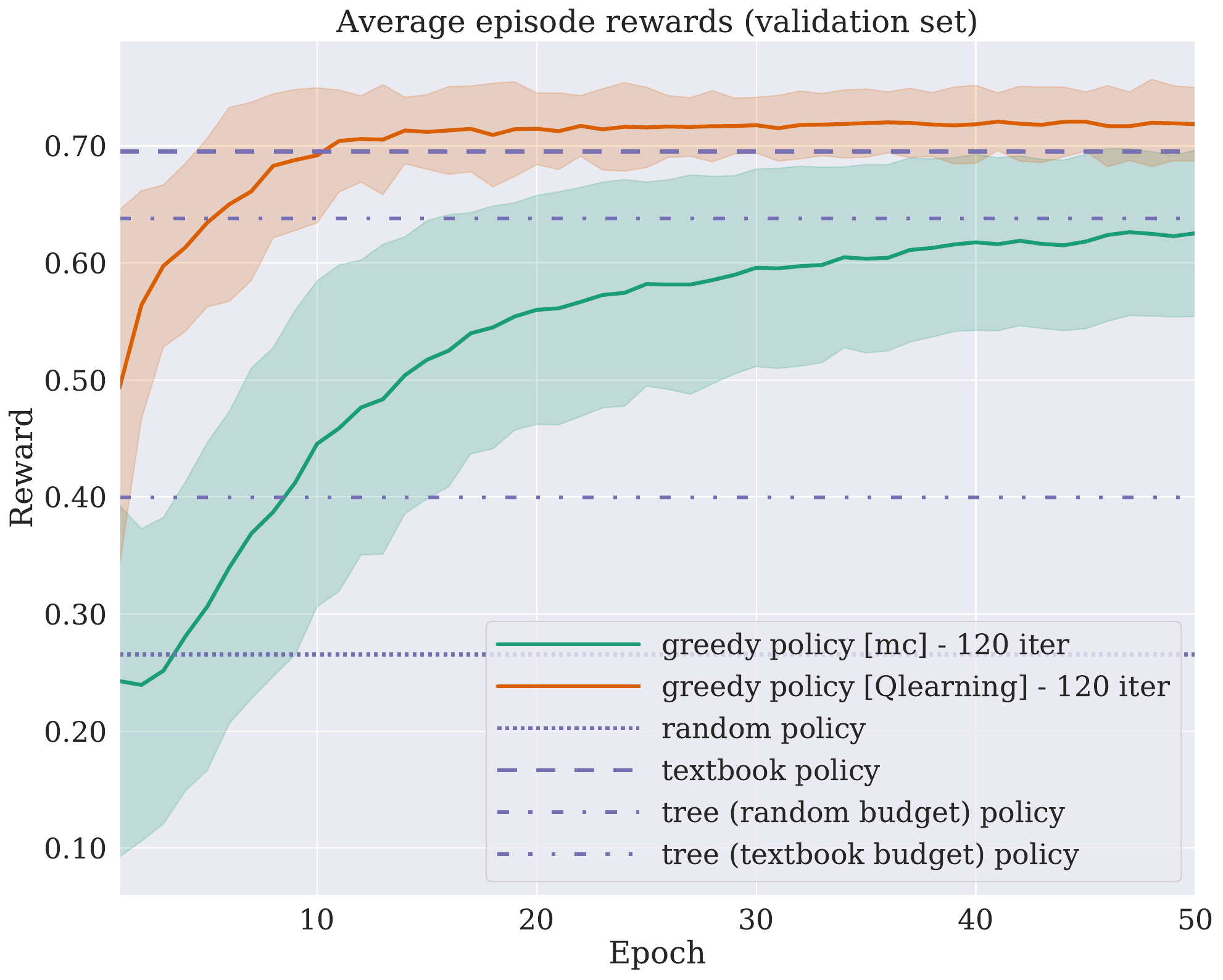}}
	\\
	\subfloat[\small{No terminal states in training (extra-U-A).}\label{fig:val_reward_diagnosis_ntt_extended}]
	{\includegraphics[width=.32\textwidth]{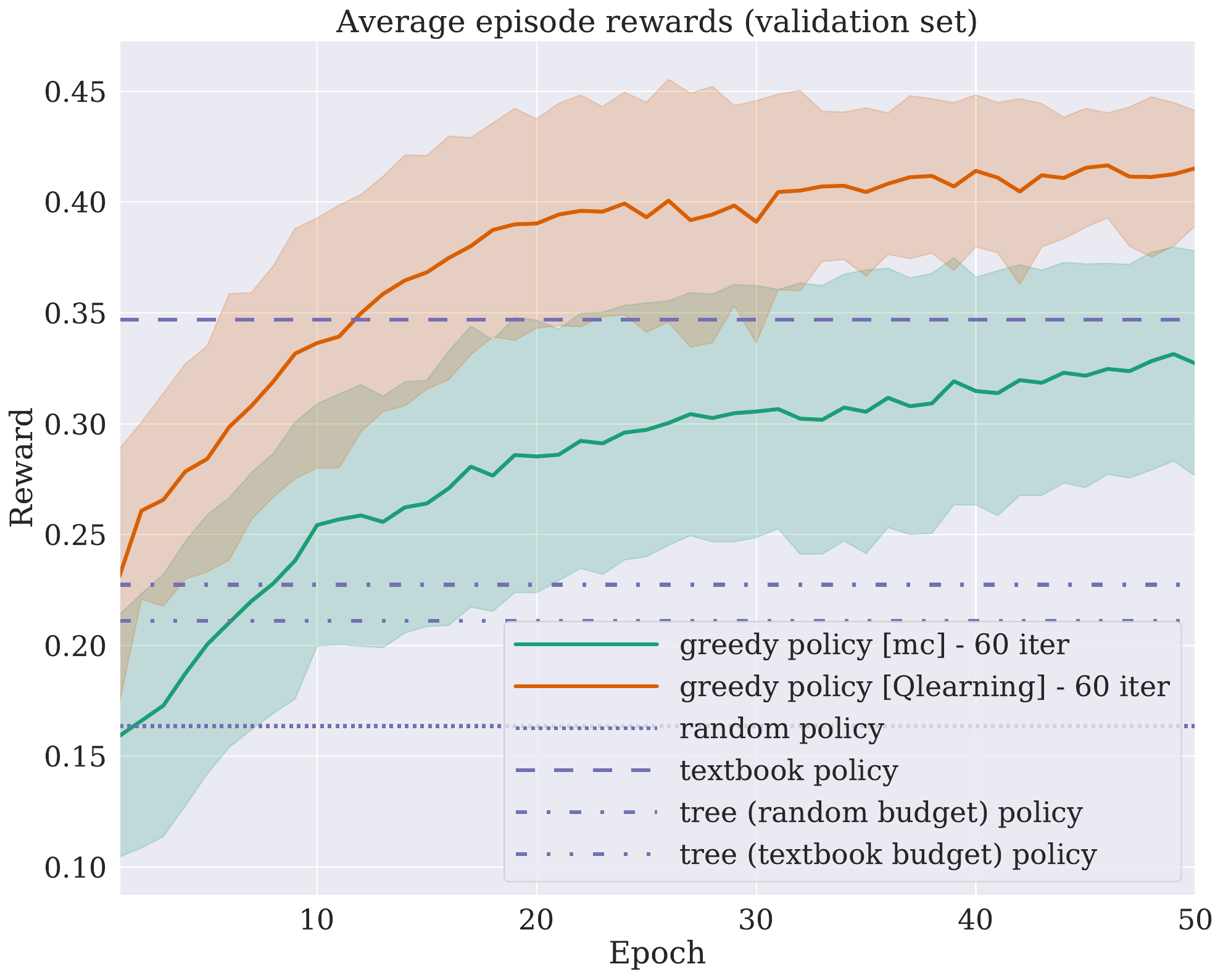}}
	\quad
	\subfloat[\small{Terminal states in training (extra-U-A).}\label{fig:val_reward_diagnosis_tt_extended}]
	{\includegraphics[width=.32\textwidth]{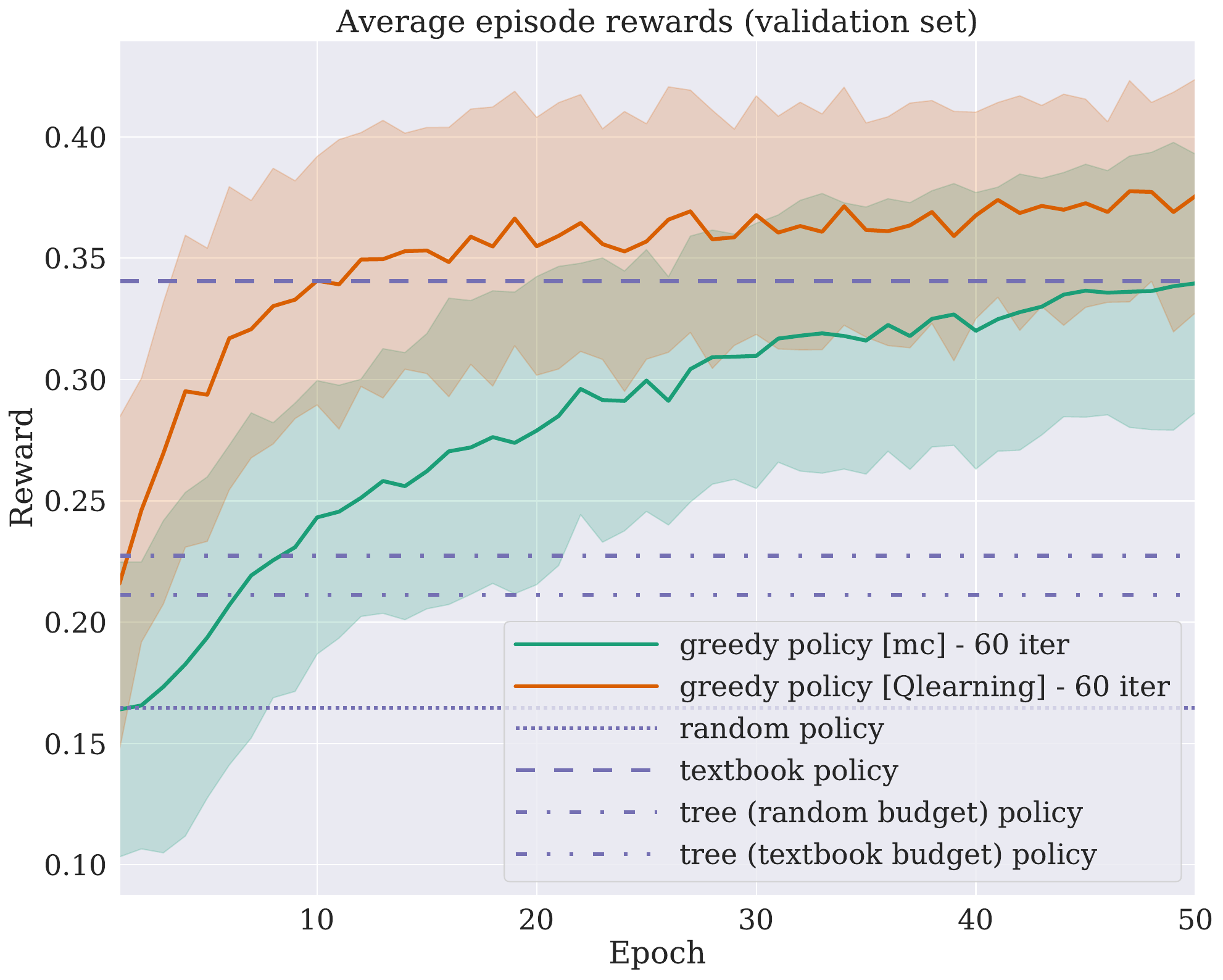}}
	\quad
	\subfloat[\small{\ref{fig:val_reward_diagnosis_ntt_extended} and \ref{fig:val_reward_diagnosis_tt_extended} combined}.]
	{\includegraphics[width=.32\textwidth]{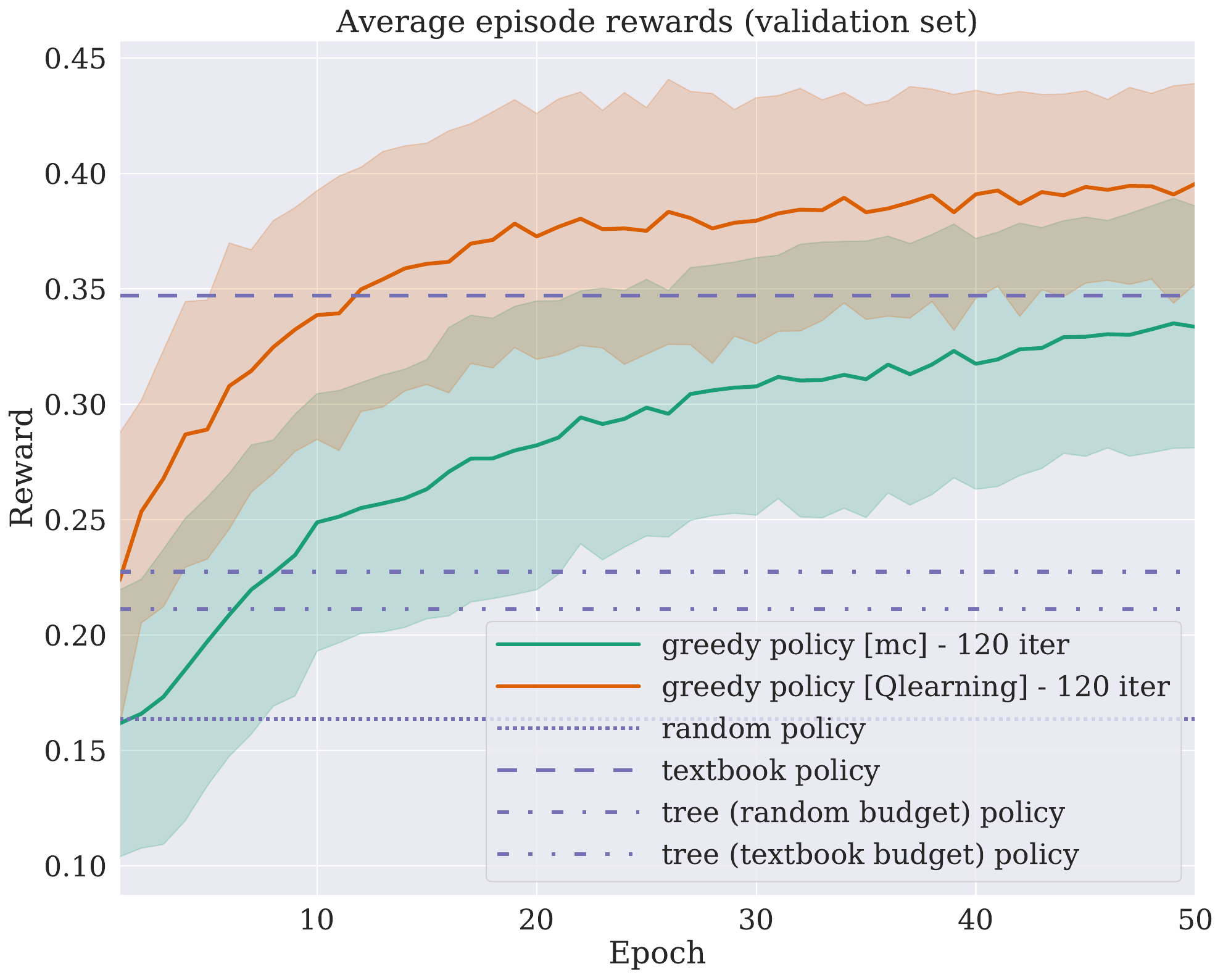}}
	\caption{\small Average validation reward during training. Solid lines show the mean $\mu $ over training iterations, and shaded regions represent $\pm \sigma$. 
	 When terminal states are used, then the tuple $(s_{\text{term}}, a, R_{s_{\text{term}}}^a)$, where $s_{\text{term}}$ is a terminal state, $a$ a random action, and $R_{s_{\text{term}}}^a = 0$, is also used in the updates.
	}
	\label{fig:val_reward_diagnosis}
\end{figure*}
To support this statement, we present in \autoref{fig:val_reward_diagnosis} the average reward during training of the \gls{rl} based \gls{qs}s, on the validation set, as compared to the baselines. The training is performed 120 times and we show here the mean and standard deviation of the reward. From these, we clearly see that the proposed \gls{rl} agents improve their respective policies and achieve higher rewards than the \gls{qs}s that were generated from the classification trees.

\subsection{Comparison of questioning strategies}
We provide in \autoref{fig:qs_trees_diagnosis} a visual representation of each of the \gls{qs} methods in the form of binary trees, for the {\bf simple-A} version (corresponding trees for the {\bf extra-U-A} version are available in the Supplementary Material,~\ref{app:trees_extended}). In addition, we provide an example of the question streams produced by each \gls{qs} for a \gls{dme} grade 2 sample in \autoref{fig:gt_mue_grade2_examples_full} (with more question streams available in the Supplementary Material,~\ref{app:qs_streams_examples}). These illustrate a potential use-case for clinicians to gain insight into a MuE's responses. Examples of question-response streams for different \gls{mue}s can be presented to experts, and the \gls{qs} depiction as a decision tree can help them choose a \gls{qs} that they consider reliable as a validation tool. Subsequently, this insight can then be used to select approriate MuEs.
\begin{figure*}
	\centering
	\subfloat[\small{Textbook QS (gold standard).}]{\includegraphics[width=\textwidth]{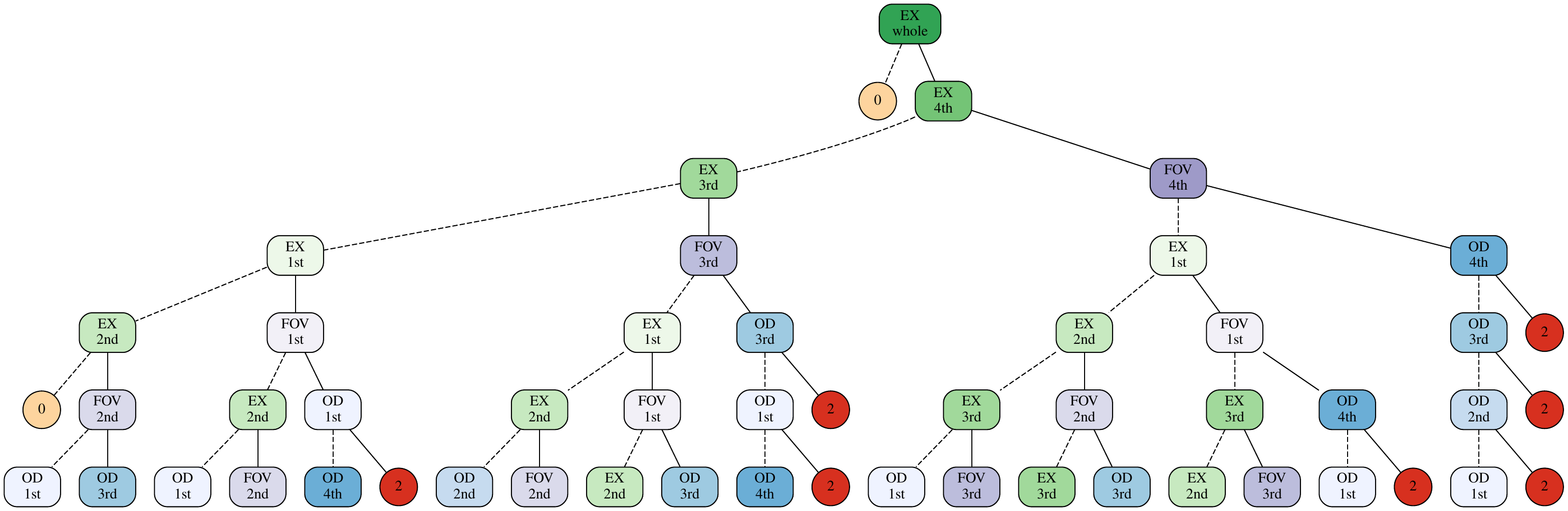}}
	\\
	\subfloat[\small{Random QS.}]{\includegraphics[width=\textwidth]{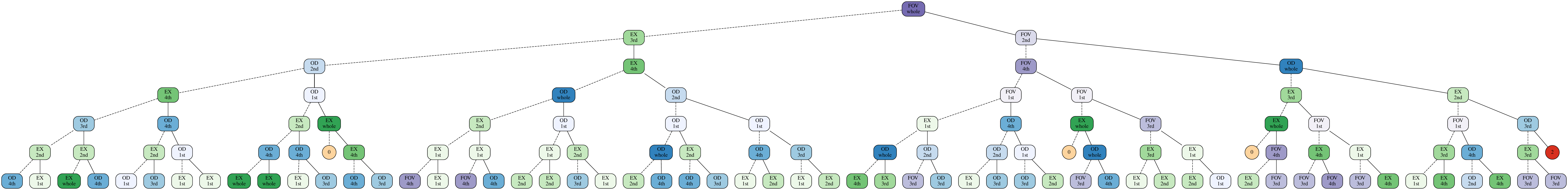}}
	\\
	\subfloat[\small{Decision tree QS (trained on random budget) (DT-RB).}]{\includegraphics[width=\textwidth]{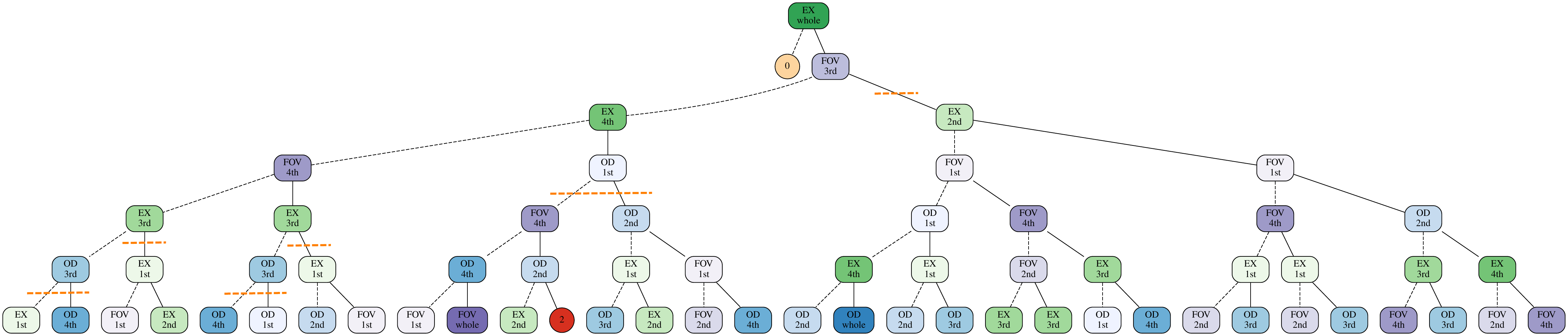}}
	\\
	\subfloat[\small{Decision tree QS (trained on textbook budget) (DT-TB).}]{\includegraphics[width=\textwidth]{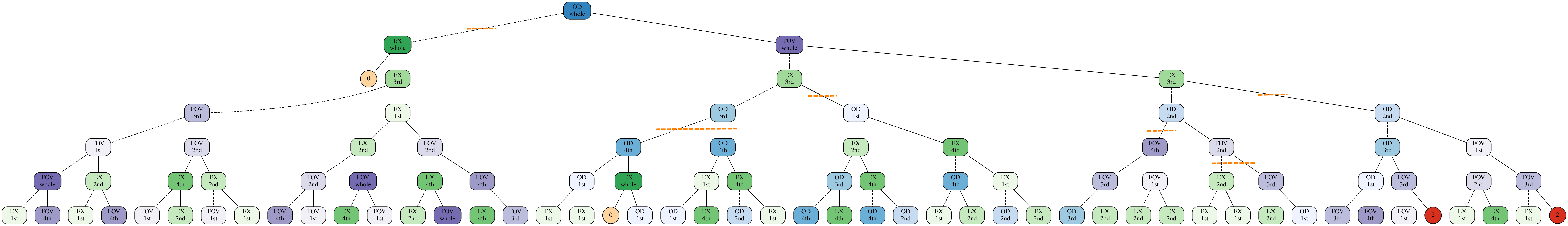}}
	\caption{\small{Decision trees for the questioning strategies for simple-A questioning assumptions (part 1 of 2). For clarity reasons, only the part of the tree up to depth 6 is shown. Notation in the tree nodes is as follows: \textbf{EX:} hard exudate, \textbf{OD:} optic disk, \textbf{FOV:} fovea. The region is specified in the second line. \textbf{Left} child of each node corresponds to answer ``No'' of the parent node question, while \textbf{right} child corresponds to answer ``Yes''. Circles correspond to terminal states with the number indicating the DME grade. Orange dashed lines on \textbf{(b)} and \textbf{(c)} correspond to the point that the baseline considers adequate for classification, and therefore random questions are chosen from then on.}}
	\label{fig:qs_trees_diagnosis}
\end{figure*}
\begin{figure*}
	\ContinuedFloat 
	\centering
	\subfloat[\small{RL QS (MC learning).}]{\includegraphics[width=\textwidth]{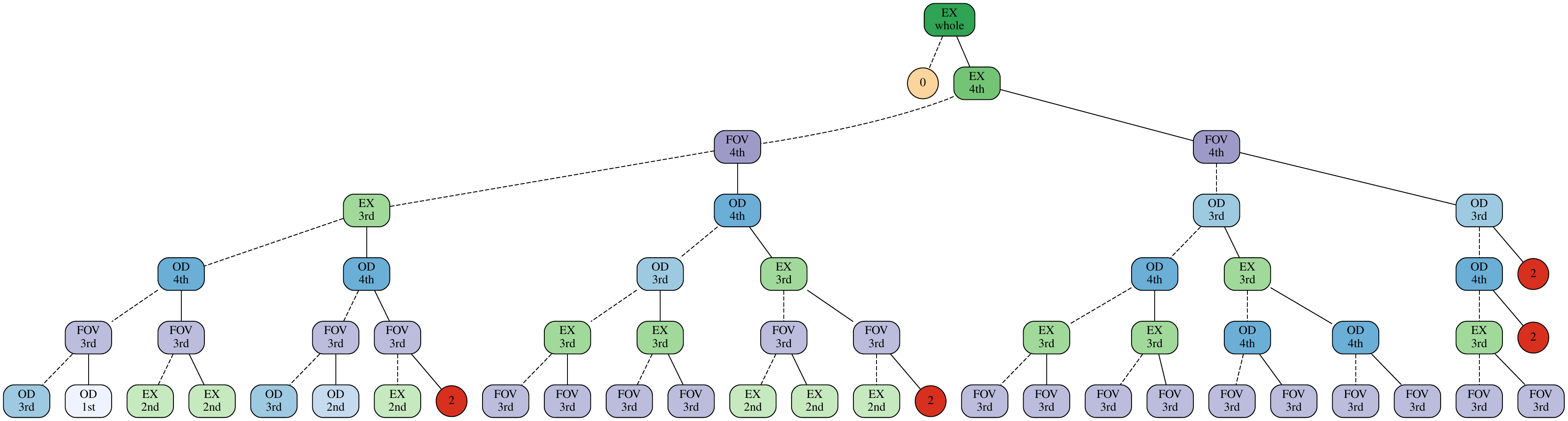}}
	\\
	\subfloat[\small{RL QS (Q-learning).}]{\includegraphics[width=\textwidth]{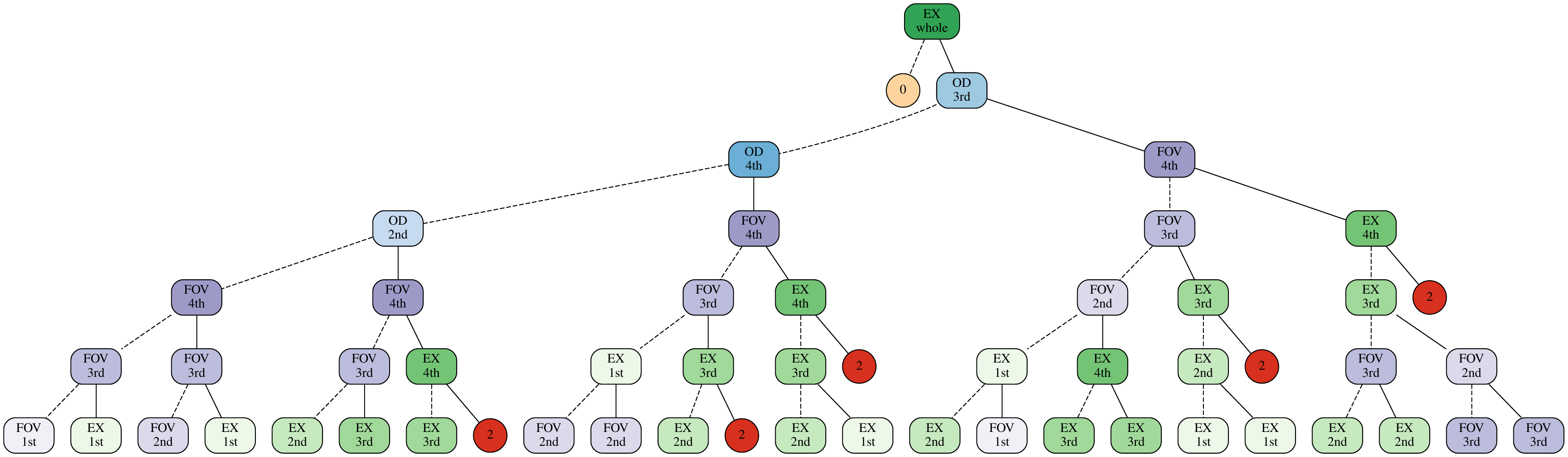}}
	\\
	\subfloat[\small{Legend for tree nodes. }]{\includegraphics[width=.63\textwidth]{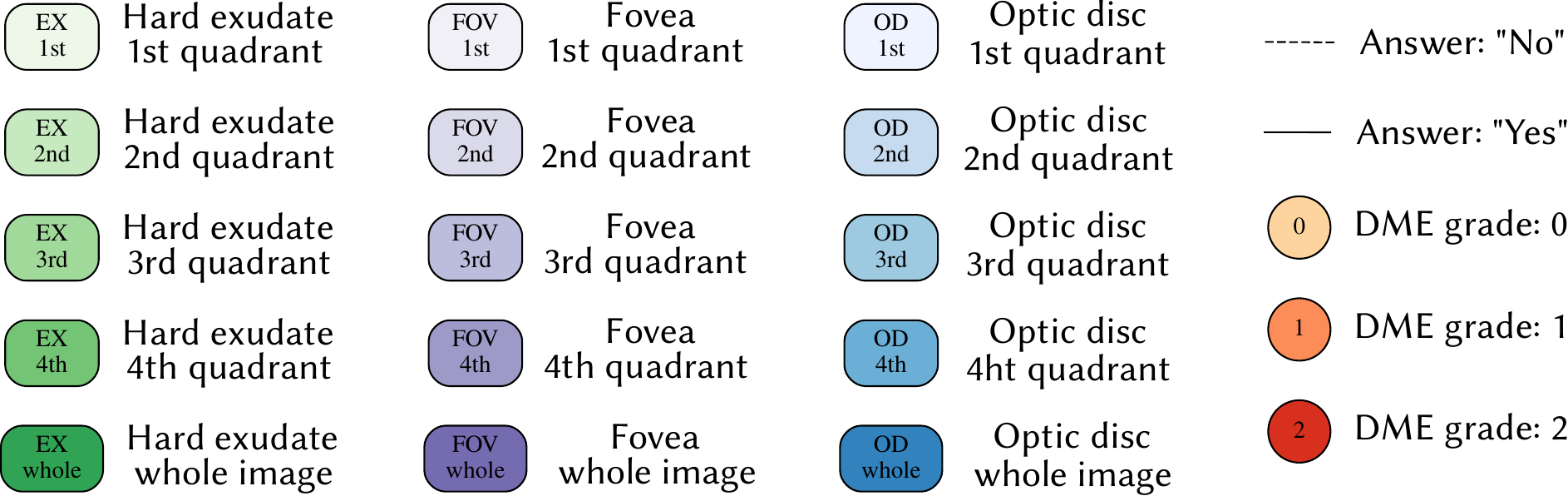}}
	\caption{\small{Decision trees for the questioning strategies for simple-A questioning assumptions (part 2 of 2). For clarity reasons, only the part of the tree up to depth 6 is shown. Notation in the tree nodes is as follows: \textbf{EX:} hard exudate, \textbf{OD:} optic disk, \textbf{FOV:} fovea. The region is specified in the second line. \textbf{Left} child of each node corresponds to answer "No" of the parent node question, while \textbf{right} child corresponds to answer ``Yes''. Circles correspond to terminal states with the number indicating the DME grade. Orange dashed lines on \textbf{(b)} and \textbf{(c)} correspond to the point that the baseline considers adequate for classification, and therefore random questions are chosen from then on.}}
	\label{fig:qs_trees_diagnosis_2}
\end{figure*}

\begin{figure}[!h]
	\centering
	\includegraphics[width=.49\textwidth]{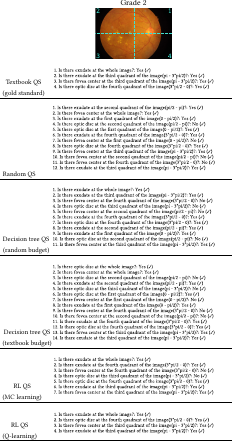}
	\caption{{Example of question stream from all questioning strategies, for a sample with DME grade 2. 
	}}
	\label{fig:gt_mue_grade2_examples_full}
\end{figure}
In \autoref{fig:qs_trees_diagnosis}, we see that some strategies are better than others at reaching terminal states with few questions. 
For instance, the random strategy is represented by a wide tree whose paths reflect no reasoning. Conversely, we can see that the \gls{rl} based questioning strategies are similar to the textbook strategy, indicating that these strategies resemble more closely the clinical decision process.

An interesting observation regarding \textbf{DT-RB QS} and \textbf{DT-TB QS} is that even though the accuracy of the classification trees is very high (see \autoref{table:tree_classification_results}), the generated questioning strategies do not reflect a clinical decision process. This can be explained by the fact that the statistical properties of the dataset are exploited when learning tree splits. However, although an attribute, or combination of attributes, may be common among the samples of the same class, this does not necessarily correspond to an important clinical criterion. For example, samples of the \gls{dme} class 0 (healthy), may happen to have a value of 0 for the question \emph{"Is there optic disc in the whole image?"}, since if there is no exudate, a diagnosis can be achieved without asking about or locating the optic disc. This however is by no means an indication for a diagnosis. A sequence of questions and responses may therefore be considered by the trained tree as sufficient to reach diagnosis, while this sequence might be far from a clinician's reasoning. Such a scenario clearly does not enhance the explainability of the \gls{mue}. 

One can also observe that \textbf{DT-RB} generates a better questioning strategy than \textbf{DT-TB}. This is expected as random episodes contain more feature variability and therefore increase the likelihood of identifying important features that participate in the diagnosis. To illustrate this, the {\bf Textbook \gls{qs}} will never ask about the optic disc for a healthy sample. This can fool the classification tree to consider the absence of the optic disc a sign of health. For the {\bf Random \gls{qs}} however, it could be that questions about the optic disc are asked for some of the healthy samples, and for others not. This way, the classifier can more easily infer that the presence of hard exudates is more important.

\subsubsection{Rewards on test set}
Using \autoref{eq:episode_reward} to compute rewards after running an episode per \gls{qs} per test image, we show the average reward for the entire test set and for the different \gls{dme} grades in \autoref{table:rewards_groundtruth_answers}.
\begin{table}[!h]
	\begin{center}
		\caption{\label{table:rewards_groundtruth_answers} Average reward over the test set (\textbf{[]} show the average number of questions needed to achieve a diagnosis). Here the \textbf{groundtruth} answer is provided to every question.}
		\resizebox{\columnwidth}{!}{
		\begin{tabular}{lrrrrr}
			\toprule
			\textbf{QS} & \textbf{Grade 0} & \textbf{Grade 1} & \textbf{Grade 2} && \textbf{Total} \\
			\toprule
			\emph{simple-A} &&&&&  \\
			\textbf{Textbook (gold standard)} & 1 [1] & 0.21 [8.1] & 0.26 [7.6] && 0.58 [4.7] \\ 
			\cmidrule{1-1}
			\textbf{Random} & 0.36 [6.9] & 0.10 [11.6] & 0.15 [10.4] && 0.24 [8.9]\\  
			\textbf{DT-RB} & \textbf{1 [1]} & 0.09 [12.2] & 0.14 [10.7] && 0.51 [6.6]\\
			\textbf{DT-TB} & 0.61 [4.9] & 0.08 [12.7] & 0.11 [11.5] && 0.33 [8.7] \\
			\textbf{RL (MC)} & \textbf{1 [1]} & 0.10 [11.3] & 0.31 [7.1] && 0.60 [4.6] \\
			\textbf{RL (Q)} & \textbf{1 [1]} & \textbf{0.14 [10]} & \textbf{0.34 [6.5]} && \textbf{0.62 [4.3]} \\
			&&&&& \\
			\emph{extra-U-A} &&&&&  \\
			\textbf{Textbook (gold standard)} & 0.40 [5.4] & 0.21 [8.1] & 0.27 [7.5] && 0.32 [6.6]\\ 
			\cmidrule{1-1}
			\textbf{Random} & 0.17 [9.6] & 0.11 [11.2] & 0.15[10.1] && 0.16 [10.1]\\  
			\textbf{DT-RB} & 0.28 [7.3] & 0.09 [12] & 0.14 [10.8] &&  0.20 [9.2]\\
			\textbf{DT-TB} &  0.28 [7.3] & 0.08 [12.6] & 0.11 [11.4] && 0.18 [9.7] \\
			\textbf{RL (MC)} & 0.41 [5.2] & 0.11 [11.7] & 0.29 [7.1] && 0.34 [6.5]\\ 
			\textbf{RL (Q)} &  \textbf{0.47 [4.6]} &  \textbf{0.12 [10.7]} &  \textbf{0.34 [6.3]} && \textbf{0.40 [5.7]}\\
			\bottomrule
		\end{tabular}
	}
	\end{center}
\end{table}

Here we can see that both RL based questioning strategies perform well and in some cases outperform the gold standard. This occurs because both those \gls{qs}s are trained to exploit the dataset properties so as to quickly attain terminal states and reduce the number of questions they need to pose. This hence increases the episode reward. For example, if the optic disc is observed in a particular quadrant in most images, it is likely that this quadrant will be queried first. This is similar to the use of experience in order to ``know where to look first'', which a real clinician may have, but our implementation of the {\bf Textbook QS} does not. Note that for \gls{dme} grade 1, the available cases are so few that the trained \gls{qs} methods do not outperform the {\bf Textbook QS} in terms of reward.

\subsubsection{Rewards on different MuEs}
To see how different QS methods can differentiate between different MuEs, we generate several synthetic \gls{mue}s. Specifically, each generated MuE has the same accuracy performance in terms of percentage of correctly answered questions over the entire question set $\mathcal{A}$:
\begin{description}
	\item[Random \gls{mue}:] $\mathit{accuracy} \times 100\%$ of randomly selected questions are answered correctly.
	\item[Reasonable \gls{mue}:] Questions that are relevant for diagnosis are answered correctly $95\%$ of the times, while the rest are answered correctly $x\%$ of the times (so that the total accuracy is still $\mathit{accuracy} \times 100\%$).
	\item[Unreasonable \gls{mue}:] Questions that are irrelevant for diagnosis are answered correctly $95\%$ of the times, while the rest are answered correctly $x\%$ of the times (so that the total accuracy is still $\mathit{accuracy} \times 100\%$).
\end{description}
Note that none of the above \gls{mue}s are trained. They are fabricated to intentionally exhibit distinct behaviors, while having a common rate of correct answers. Hence, the aim of this experiment is not to design an optimal \gls{mue} to answer questions, but to explore whether a \gls{qs} can see beyond the common accuracy. This experiment justifies the need for a \gls{qs} in the first place, by exposing that the rate of correct answers over all possible questions is not an adequate quality criterion, and can hide differences in the reasoning behavior of the \gls{mue}s.

\autoref{table:rewards_synthetic_MuEs} shows the average test set rewards for these \gls{mue}s when they have a common  $70\%$ accuracy of correct answers over the entire question set $\mathcal{A}$ (see Supplementary Material,~\ref{app:rewards_per_MuE} for performances on each \gls{dme} grade, for MuEs with common accuracy of 60\%, 70\% and 90\%).
It should be noted that in all cases, the QSs are trained with the groundtruth answers and not with the \gls{mue}'s ones, such that trained \gls{qs}s are exactly the same in all columns of \autoref{table:rewards_synthetic_MuEs}. In general, we expect that the rewards for a reasonable \gls{mue} are closer to the ones of a ``perfect responder'' \gls{mue}, both for the entire test set and for the separate grades, an expectation that is confirmed by the results. 
From these results, we observe that the RL based questioning strategies are consistently better at asking important questions, even if the \gls{mue} responder is not perfect. Those results confirm the value of selecting which questions to pose to a \gls{mue}, instead of posing all possible questions. If we posed all questions, the average acurracy would be the same for the 3 \gls{mue}s, and the one with the more desirable behavior would not stand out.
\begin{table}[!t]
	\begin{center}
			\caption{\label{table:rewards_synthetic_MuEs} 
				Average reward over the test set (\textbf{[]} show the average number of questions needed for diagnosis), for different MuEs with total accuracy $70\%$, compared to a groundtruth MuE (always answering correctly).}
			\resizebox{\columnwidth}{!}{
				\begin{tabular}{lrrrrr}
					\toprule
					\textbf{QS} & \thead{Random \\ MuE} & \thead{Reasonable \\ MuE} & \thead{Unreasonable \\ MuE} && \thead{Groundtruth \\ MuE} \\
					\toprule
					\emph{simple-A} &&&&&  \\
					\textbf{Textbook  (gold standard)} & 0.61 [4.4] & 0.59 [4.4] & 0.55 [5] && 0.58 [4.7] \\ 
					\cmidrule{1-1}
					\textbf{Random} & 0.14 [10.4] & 0.22 [9.1] & 0.04 [12.6] && 0.24 [8.9]\\  
					\textbf{DT-RB} & 0.53 [6.8] & 0.48 [7.1] & 0.39 [8.5] && 0.51 [6.6]\\
					\textbf{DT-TB} & 0.51 [7.3] & 0.49 [6.9] & 0.40 [8.3] && 0.33 [8.7]\\
					\textbf{RL (MC)} & 0.59 [5.4] & 0.57 [5.2] & 0.50 [6.5] && 0.60 [4.6]\\
					\textbf{RL (Q)} & \textbf{0.60 [4.9]} & \textbf{0.60 [4.6]} &  \textbf{0.51 [6.1]} && \textbf{0.62 [4.3]} \\ 
					&&&&& \\
					\emph{extra-U-A} &&&&&  \\
					\textbf{Textbook (gold standard)} & 0.23 [7.7] & 0.35 [6] & 0.11 [8.9] && 0.32 [6.6]\\ 
					\cmidrule{1-1}
					\textbf{Random} & 0.03 [12.4] & 0.15[10.1] & -0.03 [13.6] && 0.16 [10.1]\\ 
					\textbf{DT-RB} & 0.04 [12.3] & 0.21 [9] & -0.01 [13] && 0.20 [9.2] \\
					\textbf{DT-TB} & -0.02 [13.1] & 0.19 [9.5] & -0.06 [13.6] && 0.18 [9.7]\\
					\textbf{RL (MC)} & 0.20 [9.5] & 0.35 [6.5] & 0.15 [10.2] && 0.34 [6.5]\\
					\textbf{RL (Q)} & \textbf{0.22 [9.3]} & \textbf{0.36 [6.5]} & \textbf{0.16 [10.1]} && \textbf{0.40 [5.7]}\\ 
					\bottomrule
				\end{tabular}
			}
		\end{center}
	\end{table}

\subsubsection{MuE separation}

\begin{figure*}[!t]
	\centering
	\subfloat[{Textbook QS (gold standard).}]
	{\includegraphics[width=.4\textwidth]{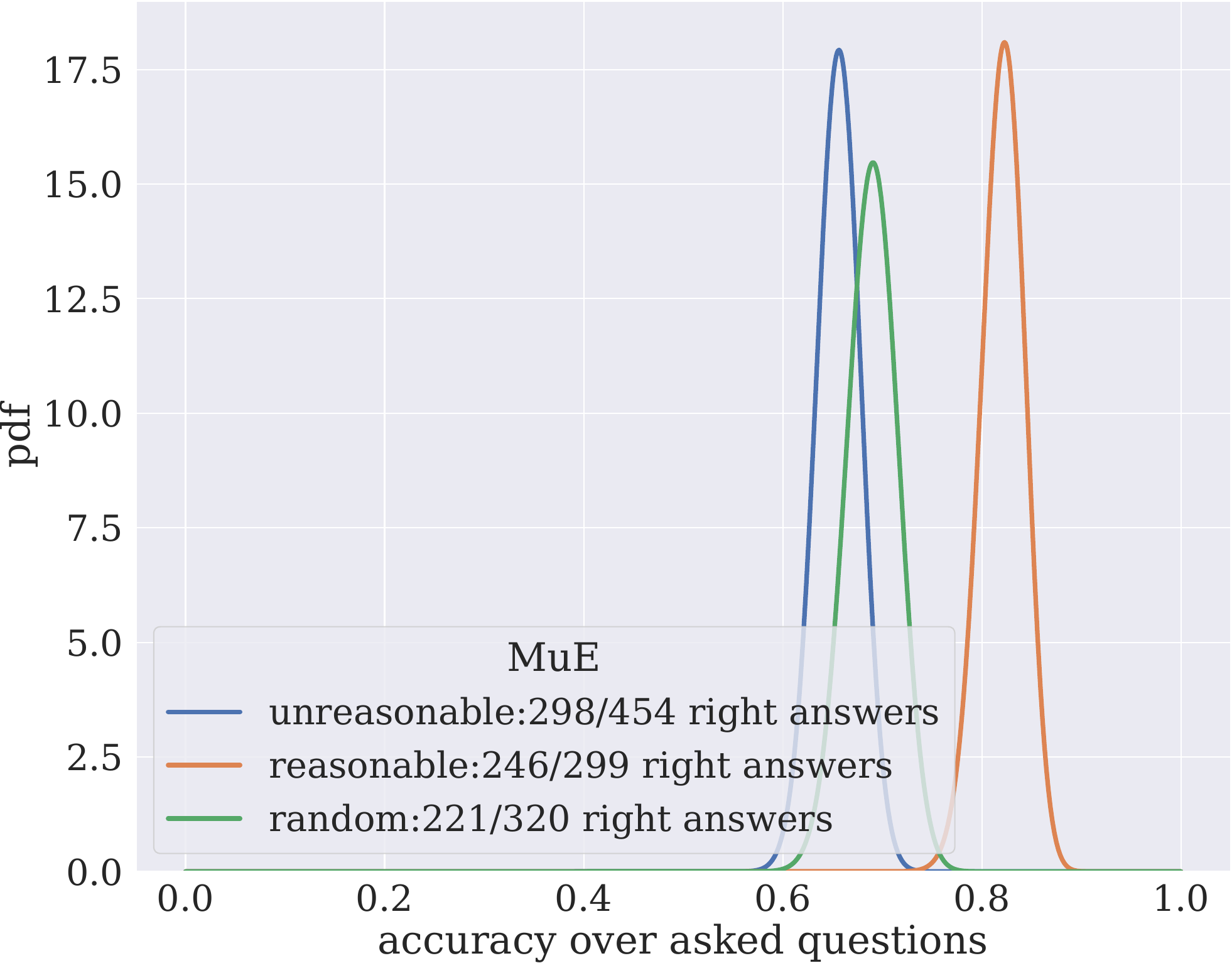}}
	\qquad
	\subfloat[{Random QS}]
	{\includegraphics[width=.4\textwidth]{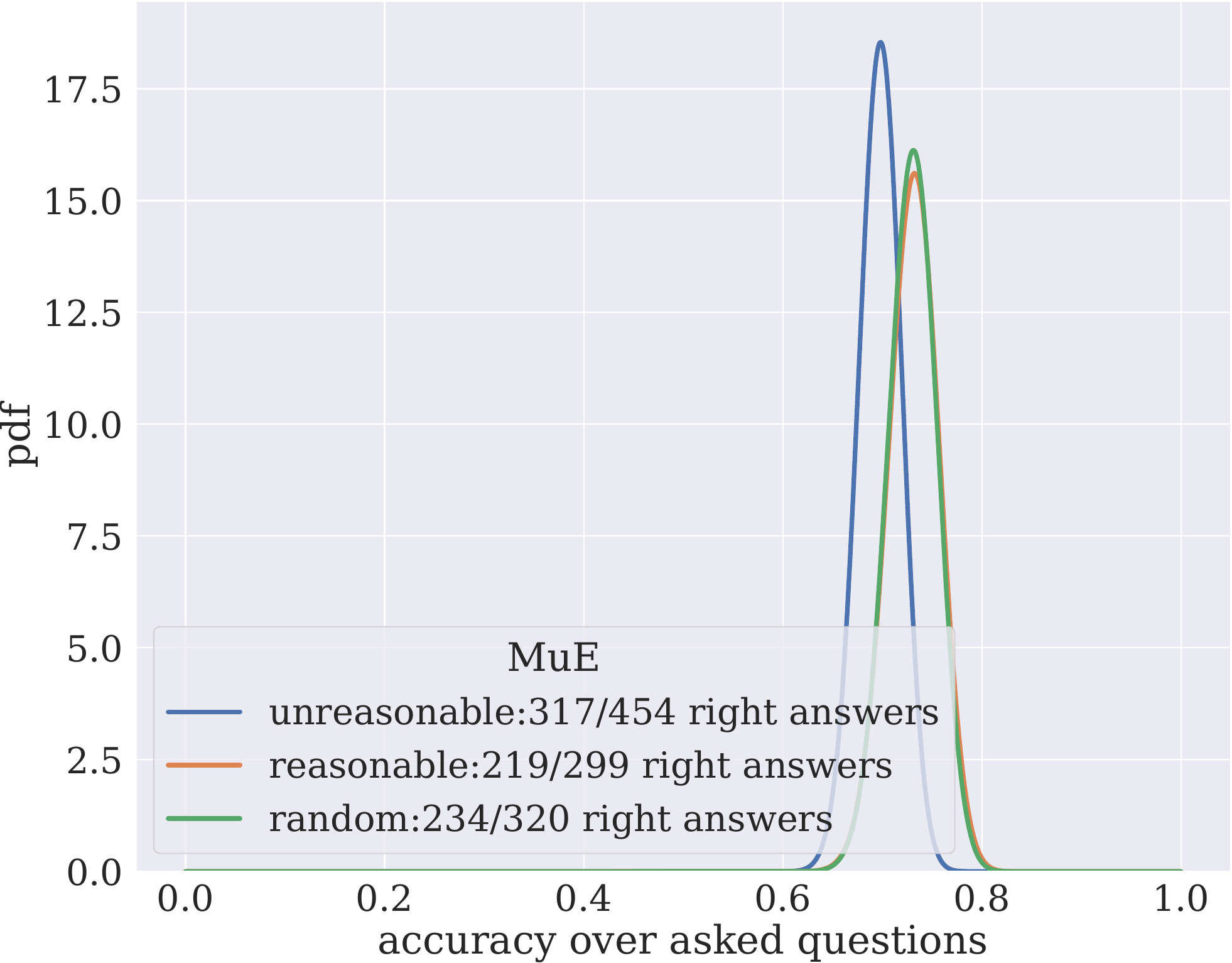}}
	\\
	\subfloat[{Tree (random budget) QS (DT-RB).}]
	{\includegraphics[width=.4\textwidth]{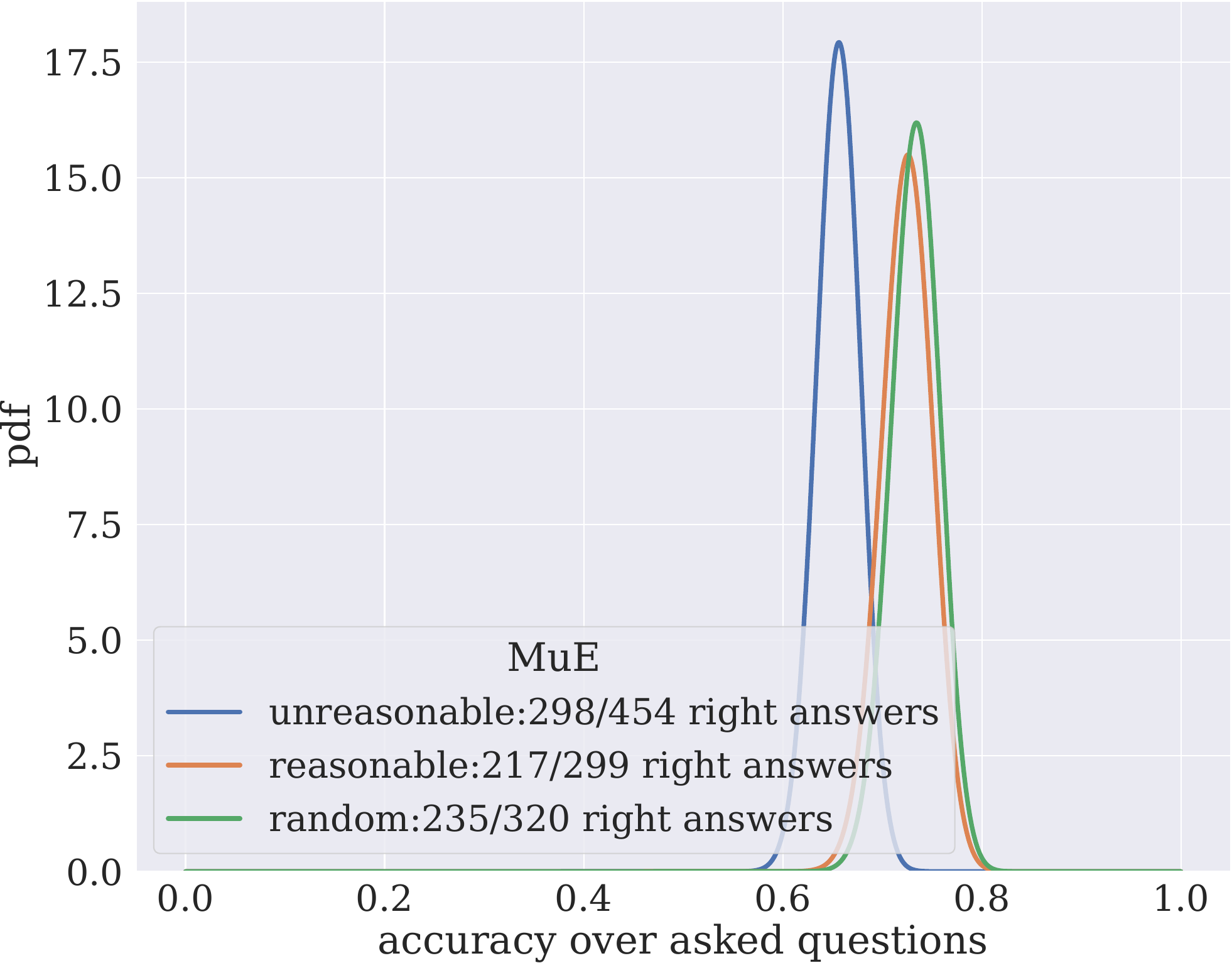}}
	\qquad
	\subfloat[{Tree (textbook budget) QS (DT-TB).}]
	{\includegraphics[width=.4\textwidth]{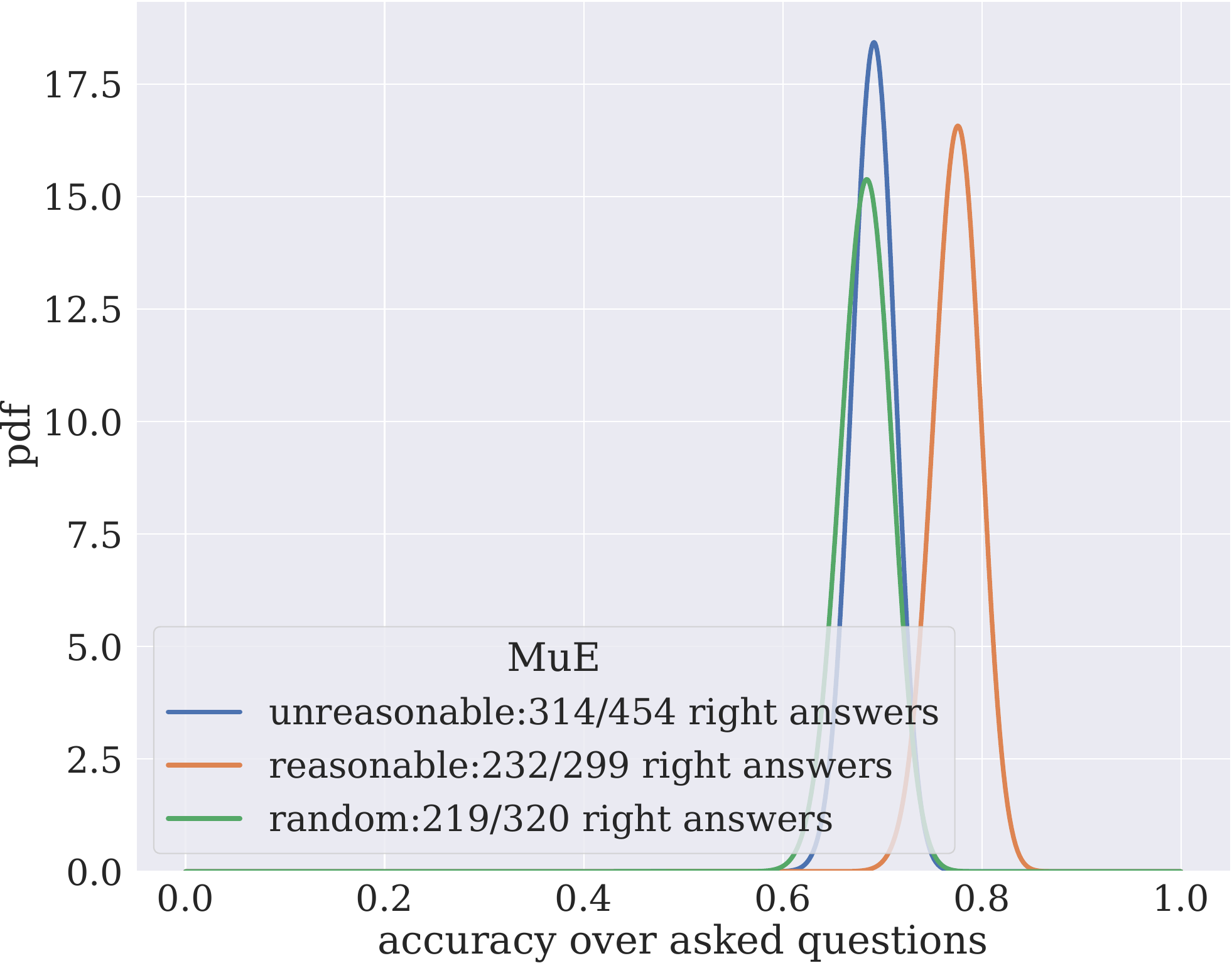}}
	\\
	\subfloat[{MC learning QS.}]
	{\includegraphics[width=.4\textwidth]{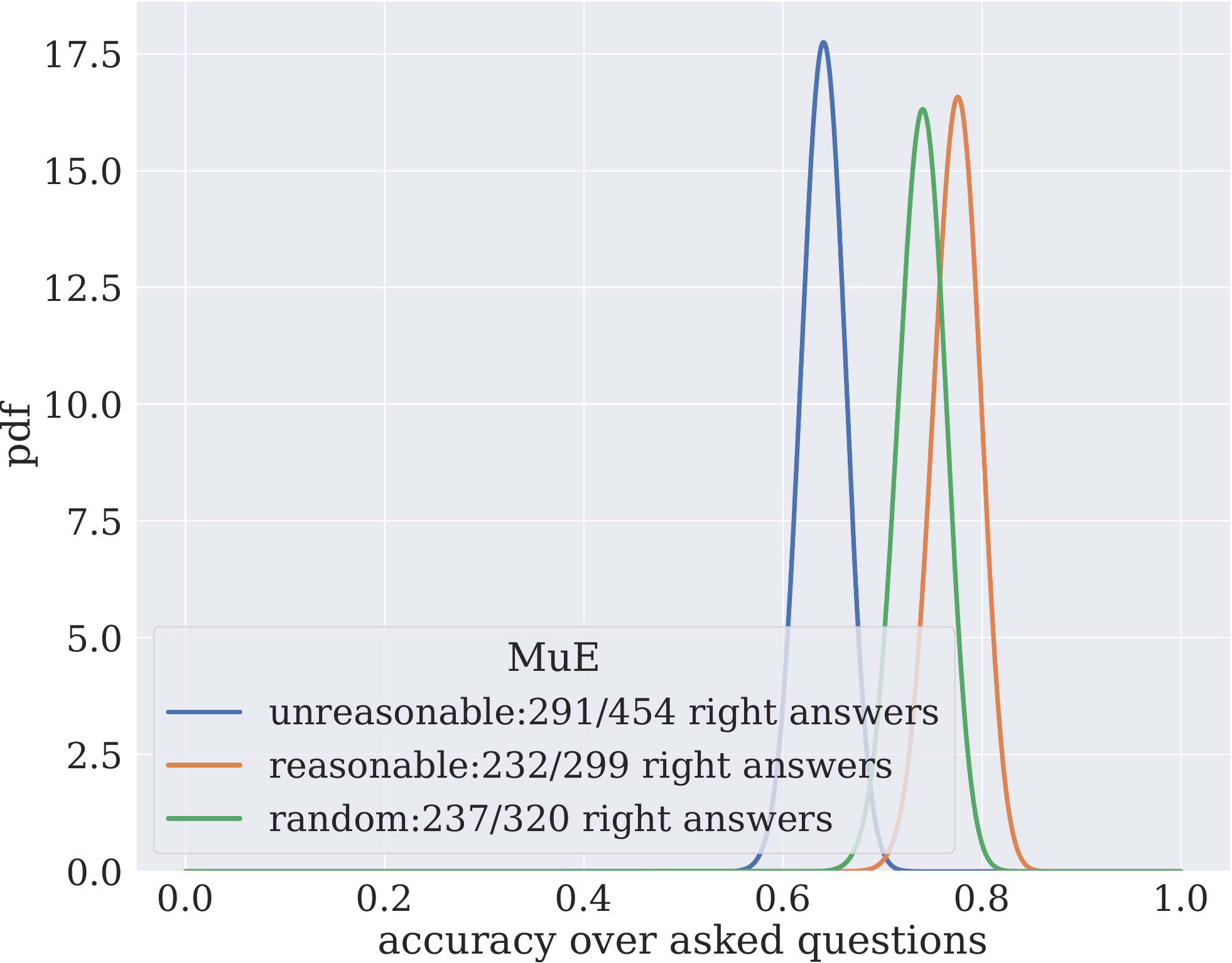}}
	\qquad
	\subfloat[{Q-learning QS.}]
	{\includegraphics[width=.4\textwidth]{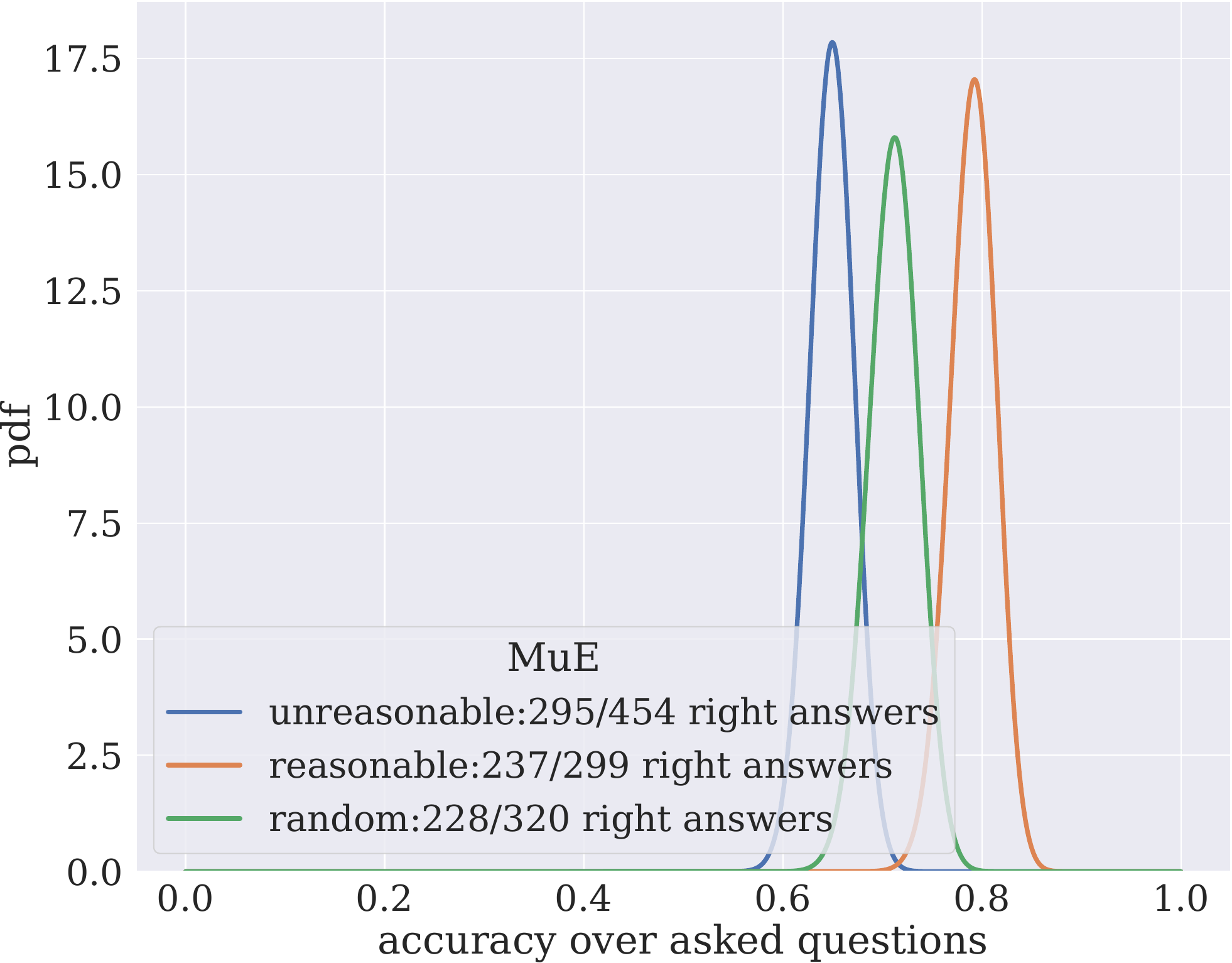}}
	\caption{Distributions of accuracy on asked questions for each QS and MuE, using the proposed beta distribution approximation. All MuEs have an accuracy of $70\%$ over the entire dataset.}
	\label{fig:mue_dist_sep}
\end{figure*}

In the next experiment, we used the {\bf random}, {\bf reasonable} and {\bf unreasonable} \gls{mue}s, all with a $70\%$ average accuracy over the entire question set $\bm{\mathcal{A}}$, to infer the distribution of responses as described in \autoref{sec:res_evaluation}. This allows us to compute a beta distribution per \gls{mue} per \gls{qs}.

We present the information radius $R_\text{qs}$ in \autoref{table:information_radius_only_answer_acc}, and the final state of the beta distributions in \autoref{fig:mue_dist_sep}. It can be seen that the {\bf Textbook QS}, as well as both versions of {\bf RL QS}, lead to distinguishable distributions between the \gls{mue}s, something that the {\bf Random QS} fails to do. This is an indication that certain questioning strategies can see beyond the common accuracy, and distinguish between \gls{mue}s with different behaviors, by asking appropriate questions. Those results confirm the value of selecting which questions to pose to a \gls{mue}, instead of posing all possible questions. If we posed all questions, the average acurracy would be the same for the three \gls{mue}s, and the one with the more desirable behavior would not stand out. 

\begin{table}[!h]
	\small
	\begin{center}	\caption{\label{table:information_radius_only_answer_acc}Information radius for different experiments. Dissimilarity between {\bf random, reasonable} and {\bf  unreasonable} MuEs, all having an accuracy of $70\%$ over the entire dataset. Higher values indicate better separation. }
		\begin{tabularx}{0.45\textwidth}{lYY}
			\toprule
			& \textbf{simple-A} & \textbf{extra-U-A}\\
			\midrule
			\textbf{Textbook QS (gold standard)} & 1.104 &  0.855\\
			\cmidrule{1-1}
			\textbf{Random QS} & 0.259 & 0.413 \\ 
			\textbf{DT-RB QS} & 0.709 &  0.591 \\
			\textbf{DT-TB QS} & 0.815 &  1.039\\ 
			\textbf{RL QS (MC learning)} & 1.06 &  \textbf{1.370}\\
			\textbf{RL QS (Q-learning)} & \textbf{1.235} & 0.896\\ 
			\bottomrule
		\end{tabularx}
	\end{center}
\end{table}

\section{Conclusions and future work}
\label{sec:conc}
In this work, we focused on determining if asking the right questions impact the evaluation of a \gls{vqa} method. To do so, we devised a trainable \gls{vtt} method that adaptively poses closed-ended questions to a \gls{vqa} method, with the intention of exposing its reasoning behavior. We use a reinforcement learning scheme to train an agent to act as the interrogator. We evaluated our framework in the context of \gls{dme} grading and show that our approach is able to generate question streams adequate for diagnosis in a small number of steps, highly resembling the reasoning process of a clinician. We also propose the use of a beta distribution, that is progressively updated after each question is asked, to characterize the performance of a MuE as perceived by the interrogator (QS). The results show that the beta distributions produced by the proposed \gls{qs} are better at distinguishing between a reasonable and an unreasonable responder (\gls{mue}), even if the two have the same average performance. The careful and dynamic question selection proves therefore to be a useful evaluation tool, since it quantitatively reveals differences between responder behaviors that the simple asking of all the questions does not. These results are consistent on both a simpler and a more complex set of clinical criteria, as illustrated by performances on the \textbf{simple-A} and the \textbf{extra-U-A} case. That is, in both cases, our approach learns what is important to ask.

Here we use the application of \gls{dme} grading to demonstrate our approach, since it is both tractable, and yet requires different medical image analysis subtasks to be solved. Naturally, in the future we plan to investigate how to apply this framework to other clinical applications. In particular, the generation of a questioning strategy for method validation is challenging to solve in a universal way, since the very notion of understanding a topic cannot be disconnected from its specificities (\eg  the question set should be adjusted to the medical task).

As future work, the generation - in close collaboration with clinicians - of more \gls{vqa} datasets that refer to particular medical tasks, would be of very high value. Also, adding a richer representation of the image in the state is of high interest and would broaden the framework's applicability. For example, a feature extractor could be used to extract low level features of an image, that could subsequently constitute part of the state. Another direction would be the expansion of the method to open-ended questions, or the selection of the region from a continuous instead of discrete image space. In addition, exploring the addition of noise in the state or the question would most likely help determine the \gls{mue}'s ability to provide consistent responses. Likewise, there are a number of logical inconsistencies that could be revealed through questioning (\eg the answer regarding the presence of a structure in the entire image being ``Yes'', but all answers regarding subregions being ``No''). Such inconsistencies could be exploited to further enhance the responder's explainability. Another type of inconsistency could be exposed by enhancing the question set with the final question `\emph{`What is the \gls{dme} grade?''}, and checking whether the \gls{mue}'s direct response, and the grade inferred by the intermediate responses agree. 
Finally, in an next step, an approach that integrates the \gls{qs} in the \gls{mue}'s training could help to improve its interpretability, a direction we did not explore in this work, where the \gls{mue} was treated as a black box and the emphasis was given on the \gls{qs} generation.

\section*{Acknowledgments}
This work was partly funded by the Swiss National Science Foundation grant No. 325230-141189 and the University of Bern.
Calculations were performed on UBELIX (\url{http://www.id.unibe.ch/hpc}), the HPC cluster at the University of Bern.

\bibliographystyle{model2-names.bst}
\biboptions{authoryear}
\bibliography{refs}

\newpage
\onecolumn
\section*{Supplementary Material}
\appendix
\section{Reinforcement learning training process}
\label{app:RL_training_algs}
Below we give the algorithms that describe the training process of an \gls{rl} agent, by Monte Carlo learning and by Q-learning~\citep{watkins1992q}.
\vfill
\begin{algorithm}[h!]
	\caption{Monte-Carlo (MC) learning}\label{alg:MCtrain}
	\begin{algorithmic}
		\Require Training dataset $\mathcal{D}$, number of epochs $N_{\text{ep}}$, decay factor $\epsilon_{\text{decay}}$
		\Ensure $\tilde{Q} \approx Q^*$
		\State Initialize: $\epsilon=1$, action-value function $\tilde{Q}$
		\For {epoch $= 1, \dots, N_{\text{epochs}}$}
		\For {im $\in \mathcal{D}$}
		\State Generate episode $\mathcal{E}$ with $\epsilon$-greedy policy \Comment \autoref{eq:egreedy}
		\State Update $\tilde{Q}$ with gradient update step on $\mathcal{L}_{\text{MC}}^{\mathcal{E}}$
		\EndFor
		\State $\epsilon \gets \text{max}\{\epsilon\cdot \epsilon_{\text{decay}}, 0.1\}$
		\EndFor
	\end{algorithmic}
\end{algorithm}
\vfill
\begin{algorithm}[h!]
	\caption{Q-learning with experience replay}\label{alg:Qltrain}
	\begin{algorithmic}
		\Require Training dataset $\mathcal{D}$, number of epochs $N_{\text{ep}}$, decay factor $\epsilon_{\text{decay}}$, maximum number of questions $N_{\text{max}}$, replay memory capacity $N_M$
		\Ensure $\tilde{Q} \approx Q^*$
		\State Initialize: $\epsilon=1$, action-value function $\tilde{Q}$
		\State Initialize: replay memory $\mathcal{M} \gets \emptyset$
		\For {epoch $= 1, \dots, N_{\text{epochs}}$}
		\For {im $\in \mathcal{D}$}
		\State Observe starting state $s_0$
		\State $s_t \gets s_0$
		\While {step $t< N_{\text{max}}$ \textbf{and} $s_t$ not terminal}
		\State Select question $a_t$ with  $\epsilon$-greedy policy \Comment \autoref{eq:egreedy}
		\State Pose question $a_t$
		\State Observe next state $s_{t+1}$ and reward $R_{s_t, s_{t+1}}^{a_t}$
		\If {$|\mathcal{M}|<N_M$} \Comment replay memory not full
		\State $\mathcal{M} \gets \{\mathcal{M}, (s_t, a_t, R_{s_t, s_{t+1}}^{a_t}, s_{t+1}) \}$
		\Else
		\State Replace oldest replay memory entry with \State transition $(s_t, a_t, R_{s_t, s_{t+1}}^{a_t}, s_{t+1})$
		\EndIf
		\State Sample random minibatch of transitions $\mathcal{T}\sim\mathcal{M}$
		\State Compute loss on minibatch $\mathcal{L}_{\text{Q-l}}^{\mathcal{T}}$
		\State Update $\tilde{Q}$ with gradient update step on $\mathcal{L}_{\text{Q-l}}^{\mathcal{T}}$
		\State $ t\gets t+1$
		\EndWhile
		\EndFor
		\State $\epsilon \gets \text{max}\{\epsilon\cdot \epsilon_{\text{decay}}, 0.1\}$
		\EndFor
	\end{algorithmic}
\end{algorithm}

\FloatBarrier
\clearpage

\section{DME textbook questioning strategies and assumptions}
\label{app:DME_textbook_trees}
\begin{figure*}[!h]
	\centering
	\subfloat[\small{Medical textbook decision tree.}\label{fig:DME_tree}]{\includegraphics[width=0.25\textwidth]{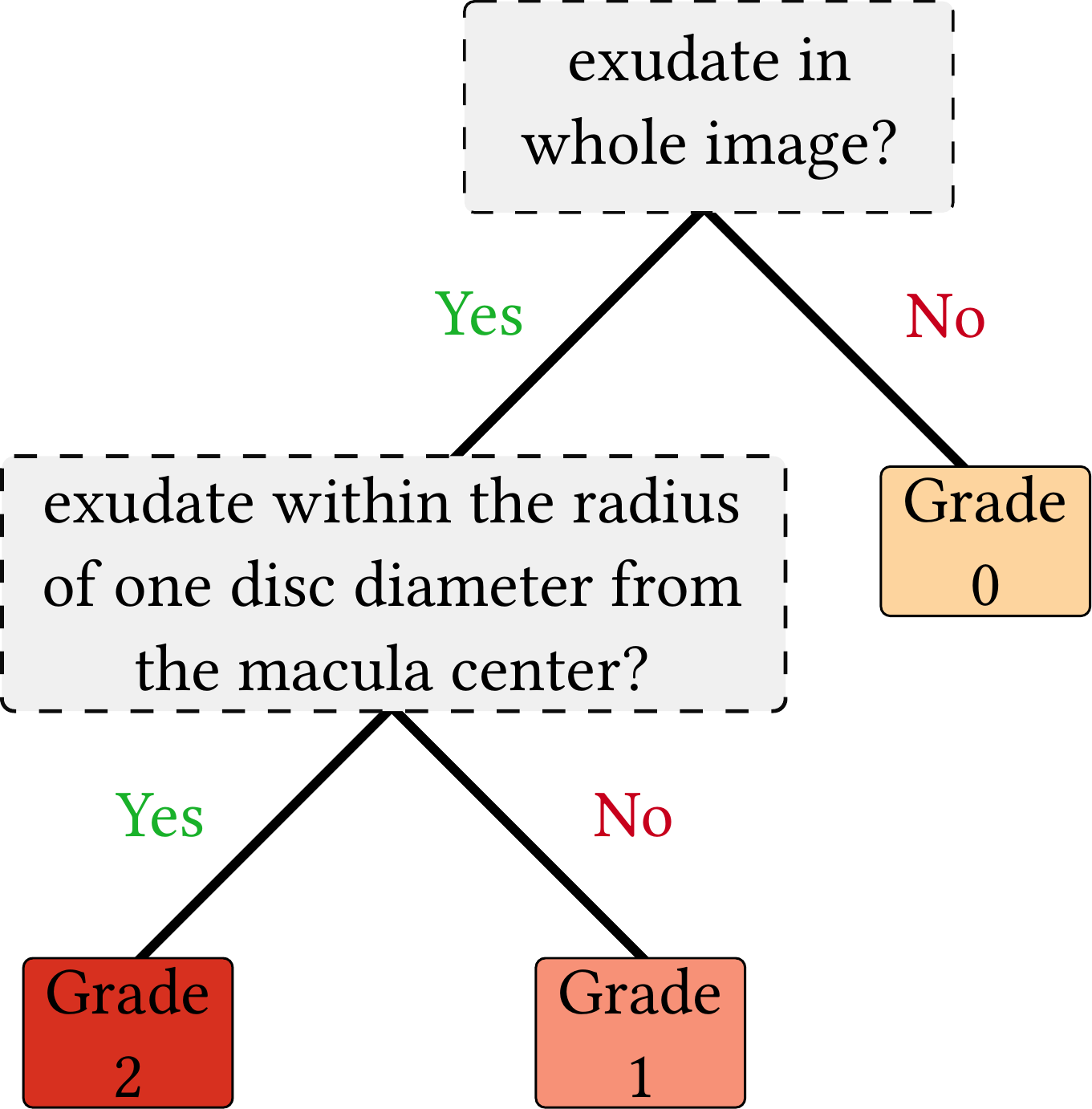}}
	\\
	\subfloat[\small{Simple-A version.}\label{fig:DME_tree_simple_questions}]{\includegraphics[height=0.4\textheight]{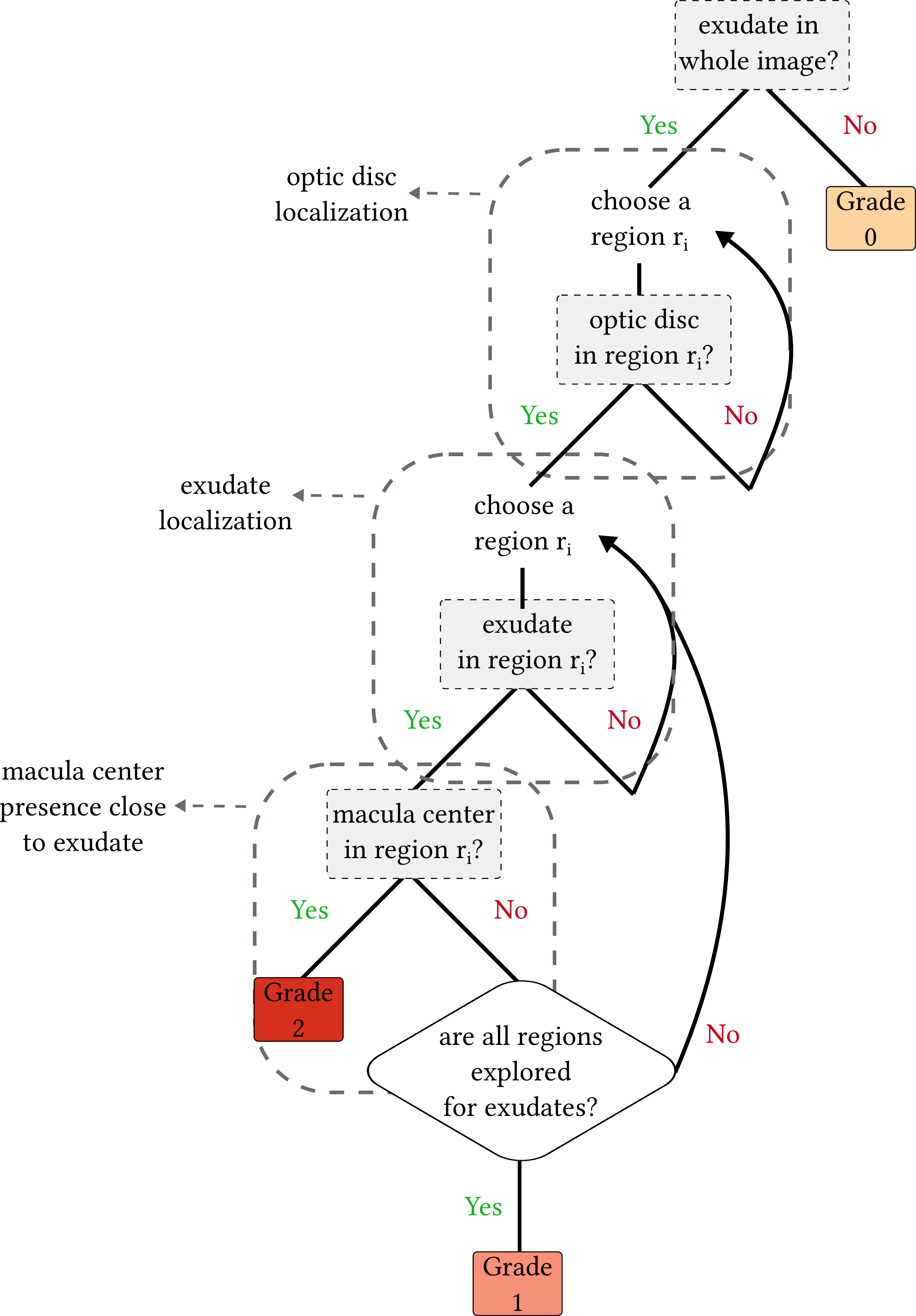}}
	\quad
	\subfloat[\small{Extra-U-A version.}\label{fig:DME_tree_simple_questions_extended}]{\includegraphics[height=0.4\textheight]{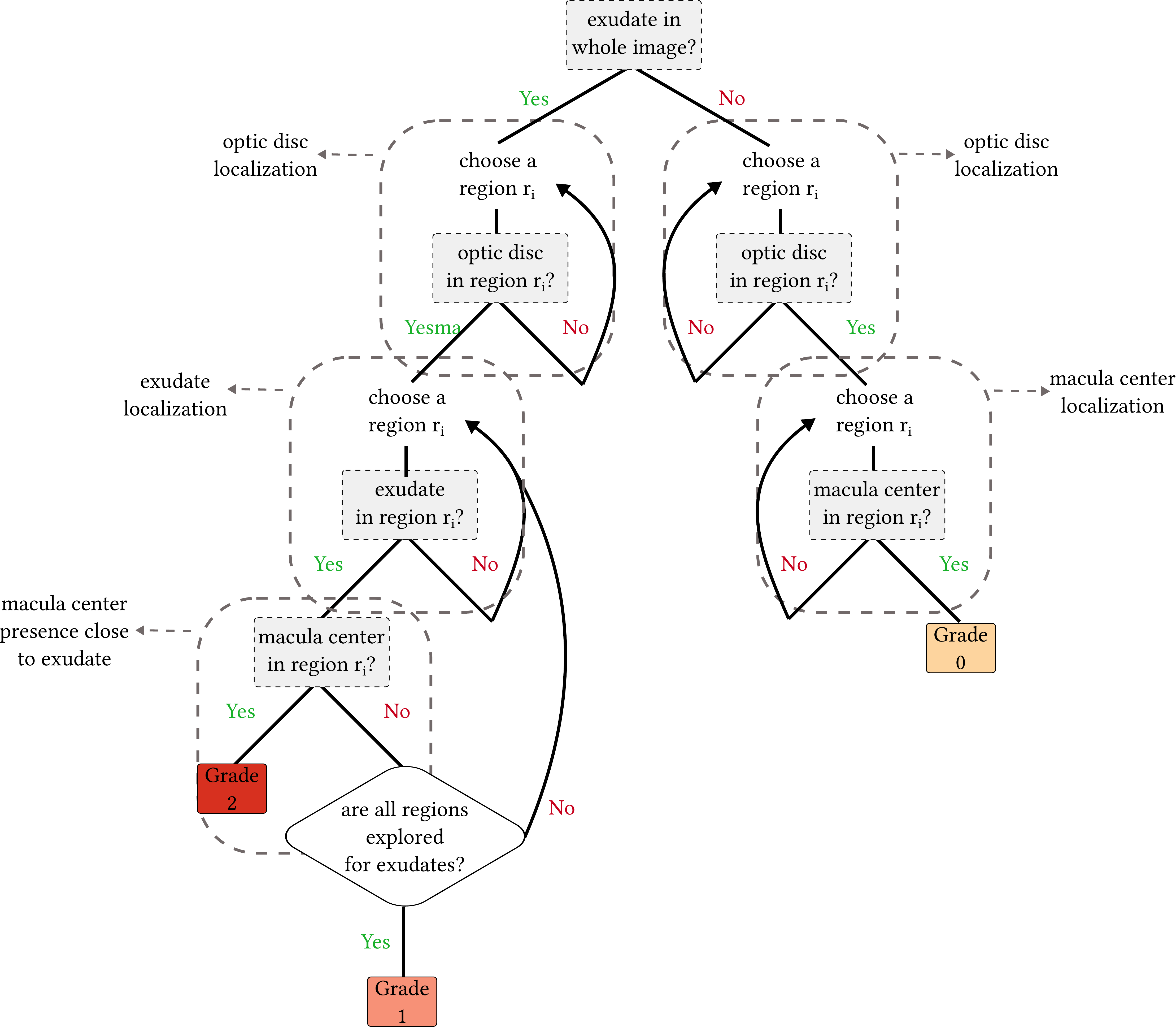}}
	\caption{\label{fig:DME_textbookTrees}\small{Decision trees for the assessment of the risk grade for DME.}}
\end{figure*}

A visualization of the textbook decision trees is shown in \autoref{fig:DME_textbookTrees}.

\begin{figure}[!h]
	\centering
	\subfloat[\small{IDRiD dataset}]{\includegraphics[width=0.25\textwidth]{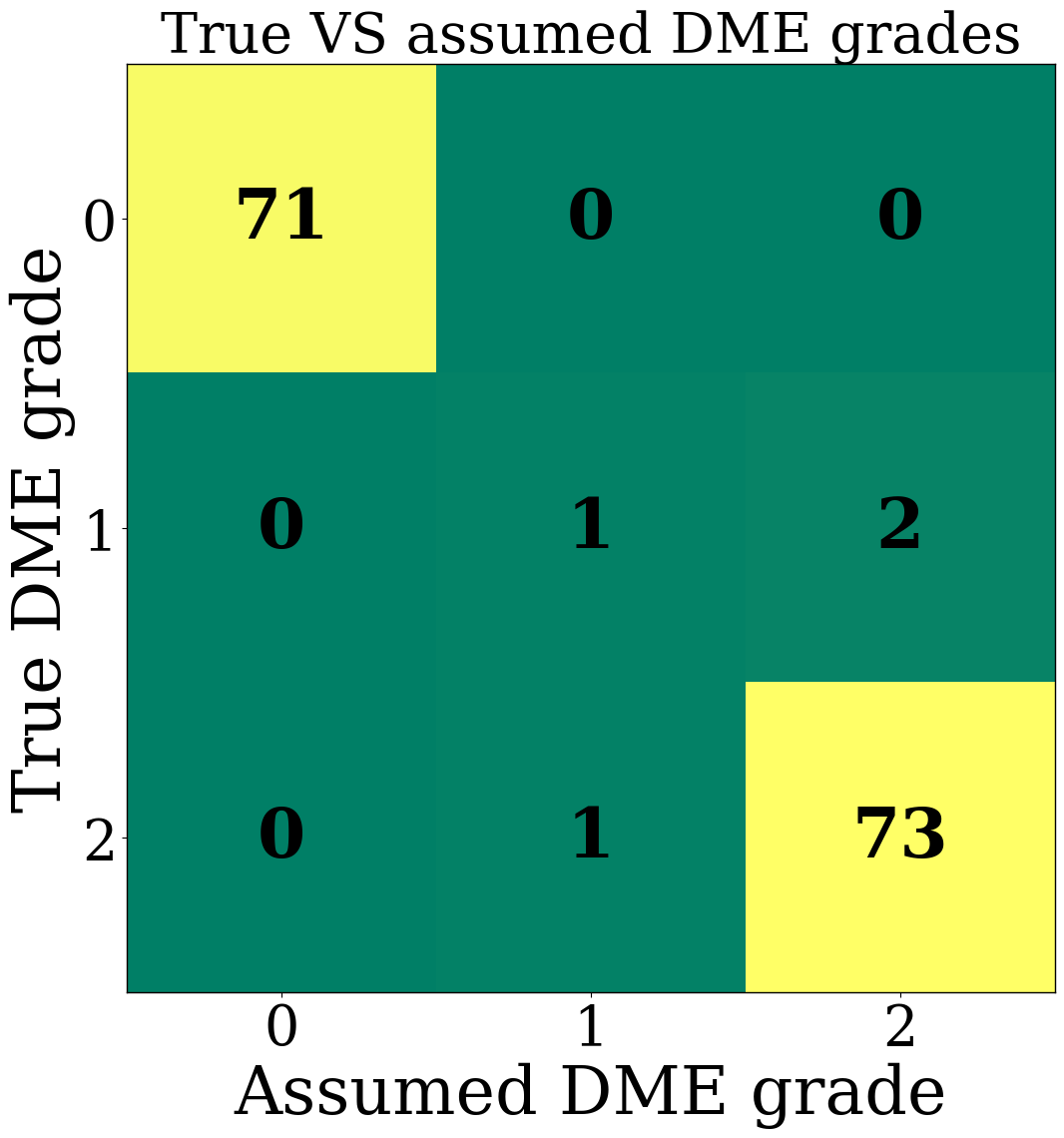}}
	~
	\subfloat[\small{eOphtha dataset}]{\includegraphics[width=0.25\textwidth]{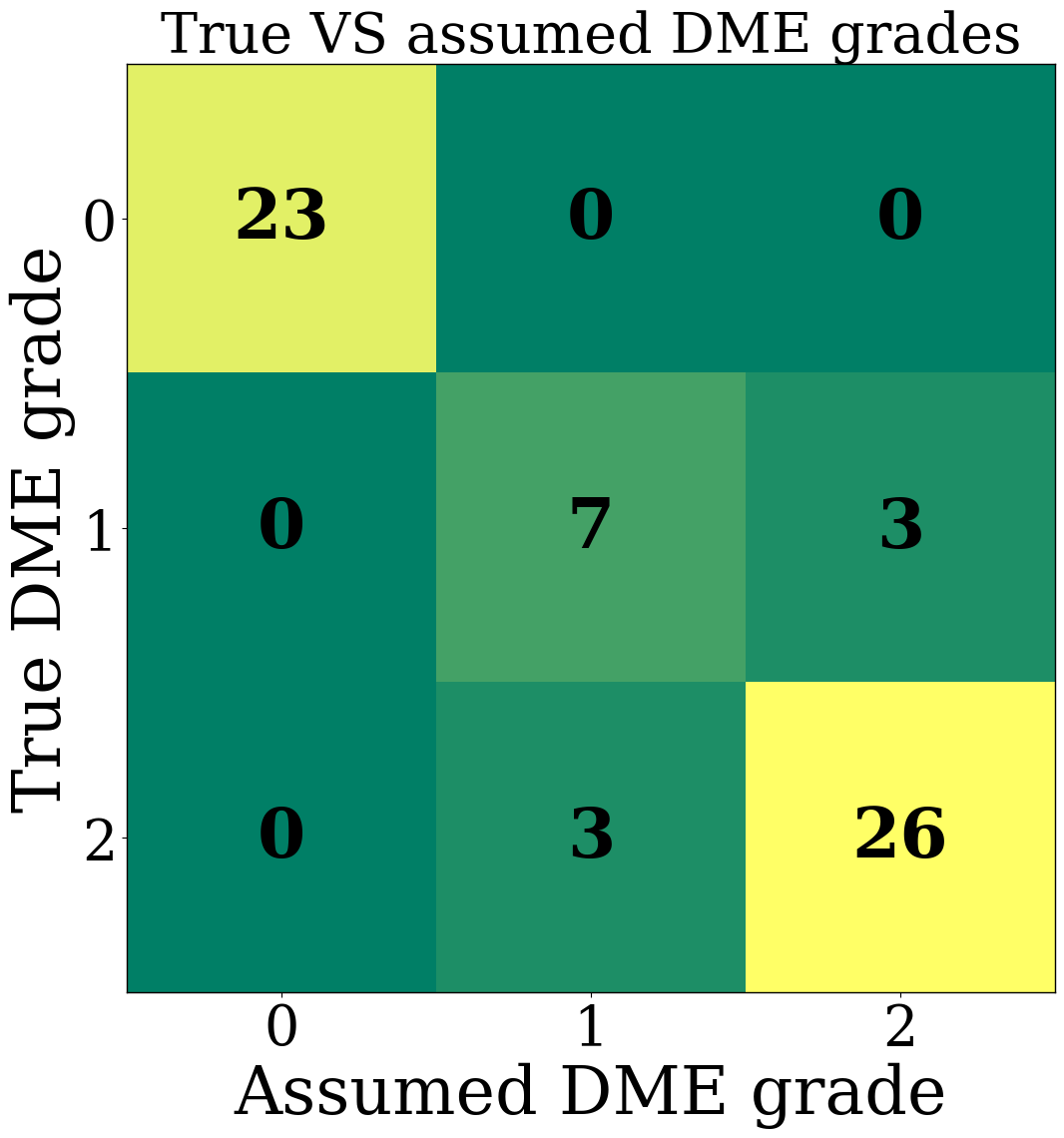}}
	~
	\subfloat[\small{Both datasets combined}]{\includegraphics[width=0.25\textwidth]{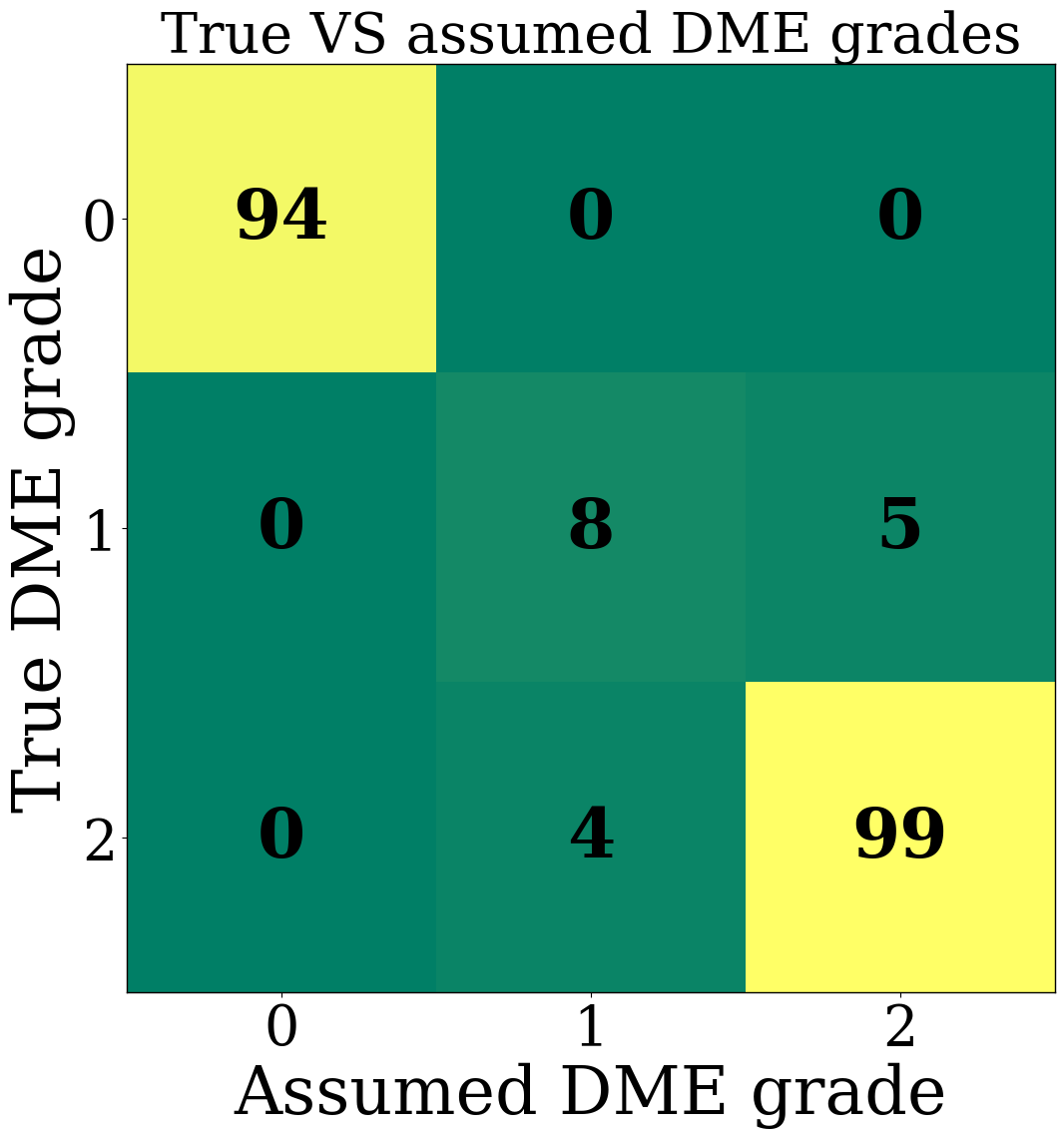}}
	\caption{\label{fig:trueVSassConfusionMatrix}\small{True VS assumed DME grades confusion matrices. True refers to the actual DME grade. Assumed refers to the DME grade that would be diagnosed by trusting the assumptions of \autoref{subsec:question_set}.}}
\end{figure}

The assumptions of \autoref{subsec:question_set} do not hold for all images in our dataset. That means, for some samples, the groundtruth \gls{dme} grade is not the same as the one assumed when taking into account the assumptions. For a given sample, the distance of exudate from the fovea center might be small enough to make it a grade 2, but the exudate and fovea center might not be at the same quadrant, for example. Note that such an error can only occur between \gls{dme} grade 1 and 2, a healthy subject can not be mistakenly diagnosed, and a unhealthy subject can not be assumed to be healthy. The number of samples for which the assumptions are not valid are presented in the form of a confusion matrix in \autoref{fig:trueVSassConfusionMatrix}, and some examples are illustrated in \autoref{fig:trueVSassExamples}.
\begin{figure}[!h]
	\centering
	\subfloat[\small]{\includegraphics[width=0.4\textwidth]{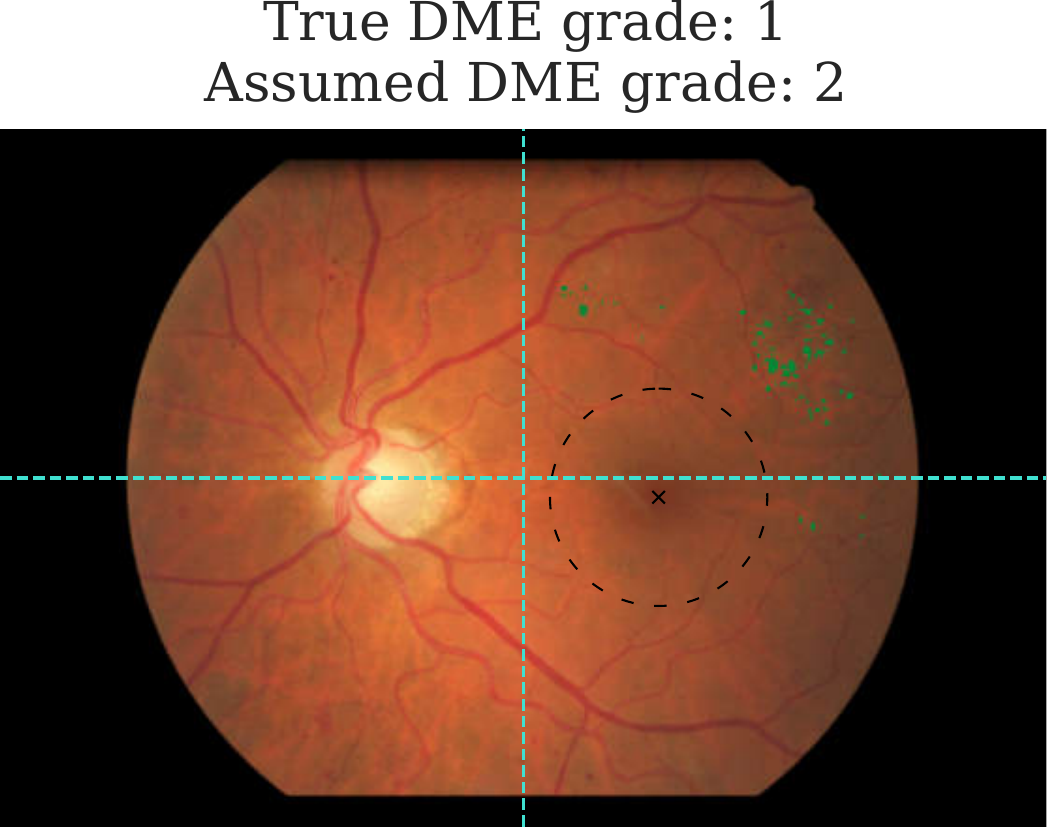}}
	~
	\subfloat[\small]{\includegraphics[width=0.4\textwidth]{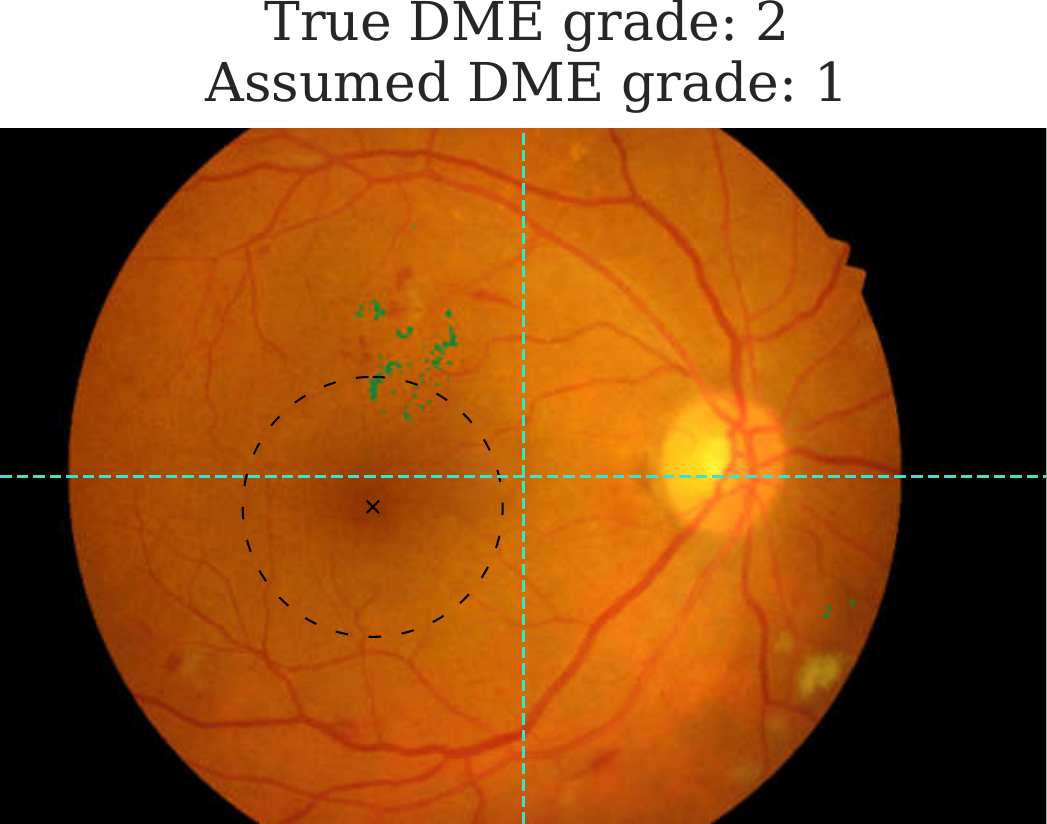}}
	\caption{\label{fig:trueVSassExamples}\small{Examples of fundus images where the assumptions do not hold.}}
	\label{fig:trueVSassExamples}
\end{figure}

\FloatBarrier
\clearpage
\section{Questioning strategy with decision trees}	
\label{app:decision_tree_learning}
The question selection process can be seen as a decision tree, where based on what is asked so far one decides what to ask next, and so on. It would therefore be only logical to try to generate this decision tree directly by training a binary tree classifier~\citep{breiman1984classification}. To do that, we assume a training set of images, and we generate for each image a series of questions, with their corresponding groundtruth answers. The questioning stops when the answers suffice for a medical diagnosis to take place (as indicated by some clinical criteria). The input to the classifier is a vector containing the entries of the transformed history sequence $\phi(\bm{H})$, as described in Eq.~\eqref{eq:fv_history}, and the target output is defined by the medical task (\eg diagnosis result, proposed treatment, etc). Our dataset therefore consists of a certain budget of feature vectors, each representing a clinically adequate question-response stream.

We compose two different budgets, differing in the question selection. In the first case, we select the next question \emph{randomly} each time, pooling from the set $\mathcal{A} \setminus \mathcal{A}_\text{asked}$ of not asked questions. In the second case, we select each time the question \emph{according to the clinical criteria (textbook)}, meaning in the same way a trained clinician would do to reach a conclusion. A \gls{qs} is then generated from the classification tree as follows: Starting from the tree root, the splitting feature is identified. Since each feature corresponds to a question in the question set $\mathcal{A}$, this question is chosen to be presented to the \gls{mue}. Depending on the answer the \gls{mue} gives, we keep traversing the tree and identify the next splitting criterion etc. During testing, the history $\bm{H}$ might not be adequate for diagnosis, however the \gls{qs} might (falsely) assume that it is, meaning that it will provide a prediction for the given history. In terms of the tree, that implies that not all tree leaves correspond to clinically sufficient feature vectors. To overcome this, we chose to keep asking random questions from the set $\mathcal{A} \setminus \mathcal{A}_{\text{asked}} $, until a sufficient history sequence is reached.

\FloatBarrier
\clearpage
\section{Questioning strategies depicted as trees for extra-U-A version of \autoref{subsec:question_set}}
\label{app:trees_extended}
\begin{figure*}[!htb]
	\centering
	\subfloat[\small{Textbook QS (gold standard).}]{\includegraphics[width=\textwidth]{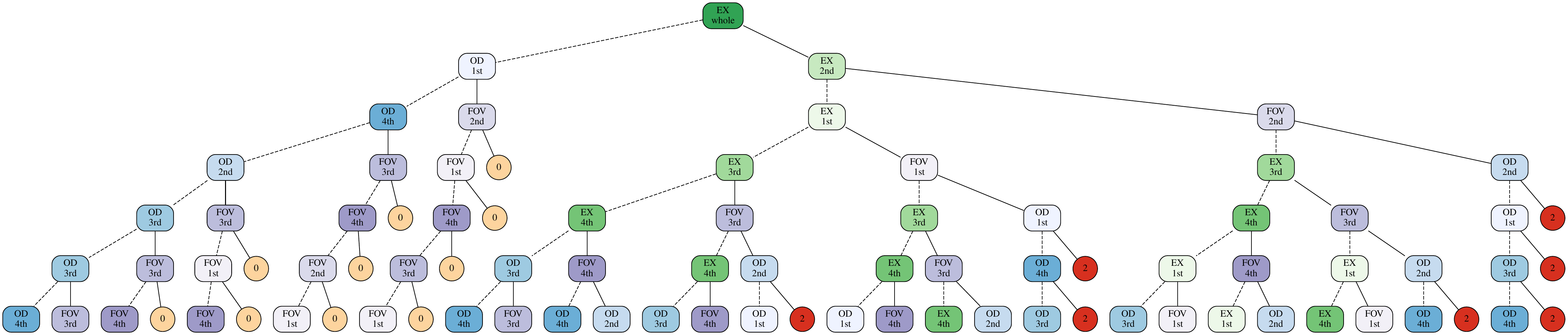}}
	\\
	\subfloat[\small{Random QS.}]{\includegraphics[width=\textwidth]{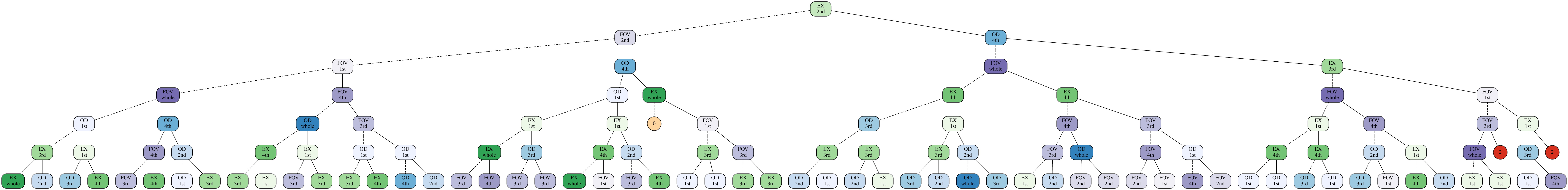}}
	\\
	\subfloat[\small{Decision tree QS (trained on random budget) (DT-RB).}]{\includegraphics[width=\textwidth]{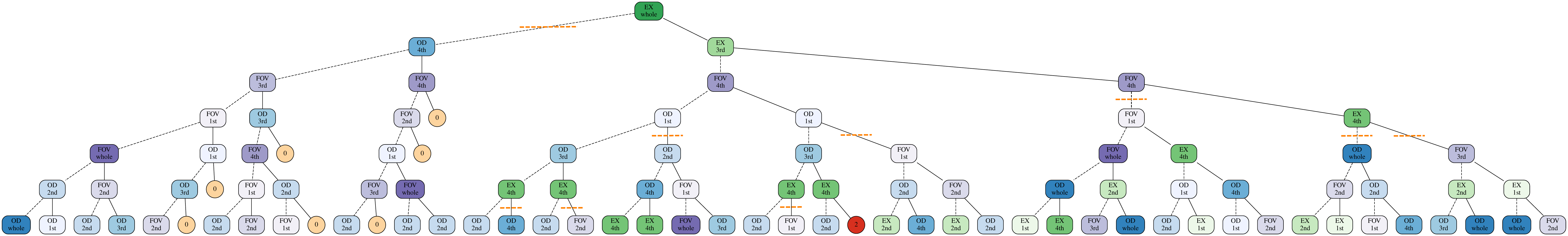}}
	\\
	\subfloat[\small{Decision tree QS (trained on textbook budget) (DT-TB).}]{\includegraphics[width=\textwidth]{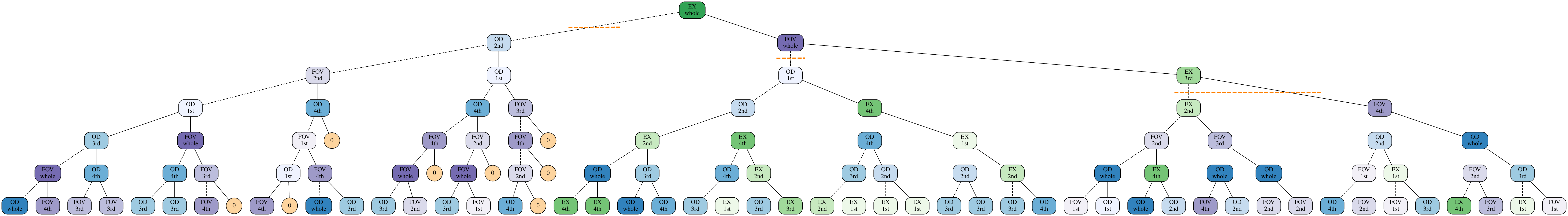}}
	\caption{\small{Decision trees for the questioning strategies for extra-U-A questioning assumptions (part 1 of 2). For clarity reasons, only the part of the tree up to depth 6 is shown. Notation in the tree nodes is as follows: \textbf{EX:} hard exudate, \textbf{OD:} optic disk, \textbf{FOV:} fovea. The region is specified in the second line. \textbf{Left} child of each node corresponds to answer ``No'' of the parent node question, while \textbf{right} child corresponds to answer ``Yes''. Circles correspond to terminal states with the number indicating the DME grade. Orange dashed lines on \textbf{(b)} and \textbf{(c)} correspond to the point that the baseline considers adequate for classification, and therefore random questions are chosen from then on.}}
	\label{fig:qs_trees_extended}
\end{figure*}

\begin{figure*}[!htb]
	\ContinuedFloat 
	\centering
	\subfloat[\small{RL QS (MC learning).}]{\includegraphics[width=\textwidth]{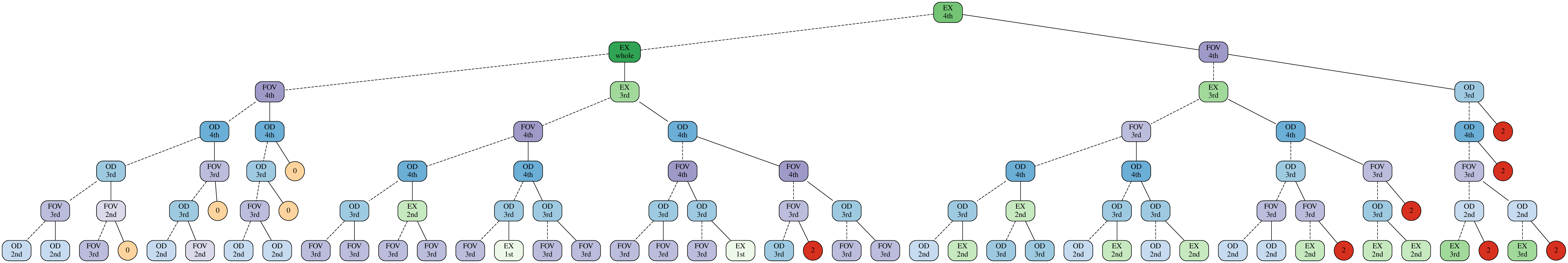}}
	\\
	\subfloat[\small{RL QS (Q-learning).}]{\includegraphics[width=\textwidth]{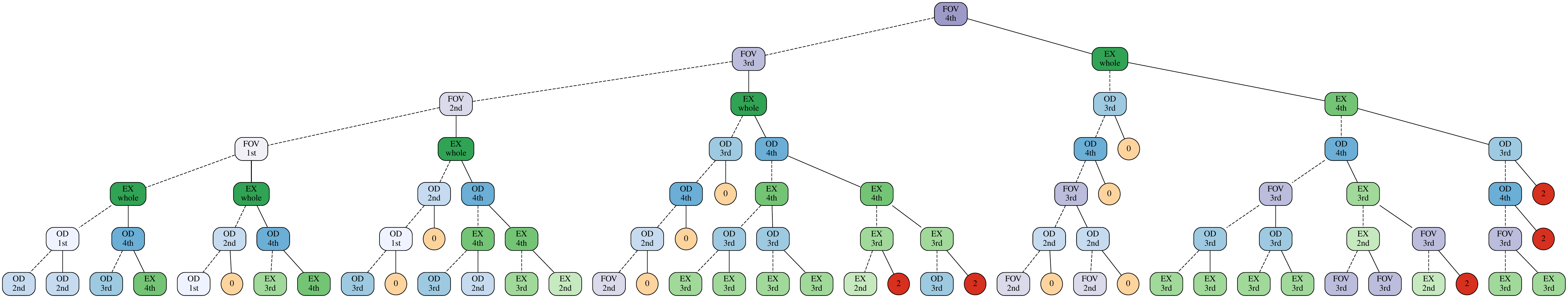}}
	\\
	\subfloat[\small{Legend for tree nodes. }]{\includegraphics[width=.6\textwidth]{tree_node_legend}}
	\caption{\small{Decision trees for the questioning strategies for extra-U-A questioning assumptions (part 2 of 2). For clarity reasons, only the part of the tree up to depth 6 is shown. Notation in the tree nodes is as follows: \textbf{EX:} hard exudate, \textbf{OD:} optic disk, \textbf{FOV:} fovea. The region is specified in the second line. \textbf{Left} child of each node corresponds to answer ``No'' of the parent node question, while \textbf{right} child corresponds to answer ``Yes''. Circles correspond to terminal states with the number indicating the DME grade. Orange dashed lines on \textbf{(b)} and \textbf{(c)} correspond to the point that the baseline considers adequate for classification, and therefore random questions are chosen from then on.}}
	\label{fig:qs_trees_extended}
\end{figure*}
\FloatBarrier
\clearpage
\section{Examples of question streams}
\label{app:qs_streams_examples}
We present below examples of generated question streams for the different questioning strategies, methods under evaluation, assumption versions (see \autoref{subsec:question_set}, page~\pageref{assumptions}) and \gls{dme} grades. The examples are selected randomly.
\begin{figure*}[!h]
	\centering
	\includegraphics[width=.9\textwidth]{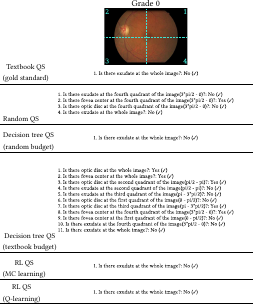}
	\caption{\small{Example of question stream from all questioning strategies, for a sample with grade 0. \textbf{Groundtruth} answers are provided to the questions. \textbf{Simple-A} version}.}
	\label{fig:gt_mue_grade0_examples}
\end{figure*}


\begin{figure*}[!h]
	\centering
	\includegraphics[width=.9\textwidth]{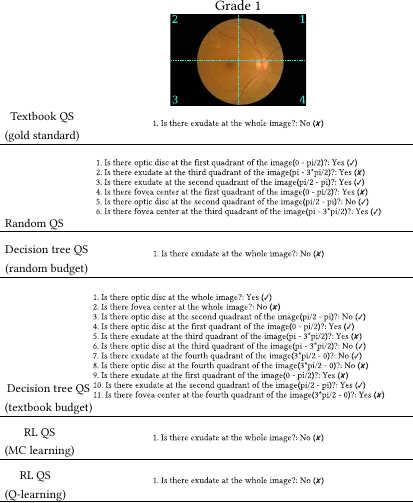}
	\caption{\small{Example of question stream from all questioning strategies, for a sample with true grade 1. Answers from a \textbf{random} MuE with a total accuracy of $70\%$ are provided to the questions. \textbf{Simple-A} version.}}
	\label{fig:random_mue_07_grade1_examples}
\end{figure*}


\begin{figure*}[!h]
	\centering
	\includegraphics[width=.9\textwidth]{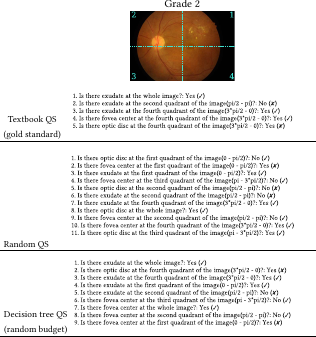}
	\caption{\small{Example of question stream from all questioning strategies, for a sample with true grade 2 (part 1 of 2). Answers from a \textbf{reasonable} MuE with a total accuracy of $70\%$ are provided to the questions. \textbf{Simple-A} version.}}
	\label{fig:reasonable_mue_07_grade2_examples}
\end{figure*}
\begin{figure*}[!h]
	\ContinuedFloat 
	\centering
	\includegraphics[width=.9\textwidth]{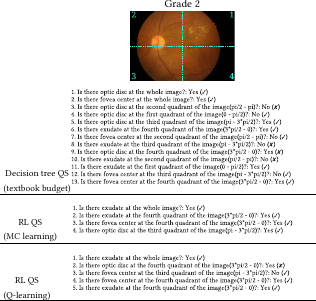}
	\caption{\small{Example of question stream from all questioning strategies, for a sample with true grade 2 (part 2 of 2). Answers from a \textbf{reasonable} MuE with a total accuracy of $70\%$ are provided to the questions. \textbf{Simple-A} version.}}
\end{figure*}

\begin{figure*}[!h]
	\centering
	\includegraphics[width=.9\textwidth]{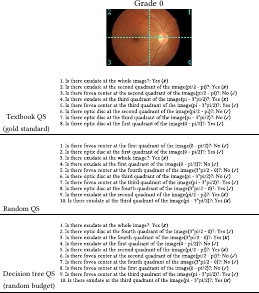}
	\caption{\small{Example of question stream from all questioning strategies, for a sample with true grade 0 (part 1 of 2). Answers from an \textbf{unreasonable} MuE with a total accuracy of $70\%$ are provided to the questions. \textbf{Simple-A} version.}}
	\label{fig:unreasonable_mue_07_grade0_examples}
\end{figure*}
\begin{figure*}[!h]
	\ContinuedFloat 
	\centering
	\includegraphics[width=.9\textwidth]{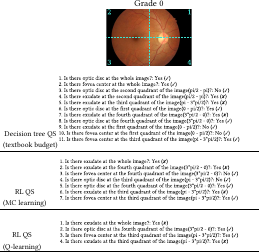}
	\caption{\small{Example of question stream from all questioning strategies, for a sample with true grade 0 (part 2 of 2). Answers from an \textbf{unreasonable} MuE with a total accuracy of $70\%$ are provided to the questions. \textbf{Simple-A} version.}}
\end{figure*}


\begin{figure*}[!h]
	\centering
	\includegraphics[width=.9\textwidth]{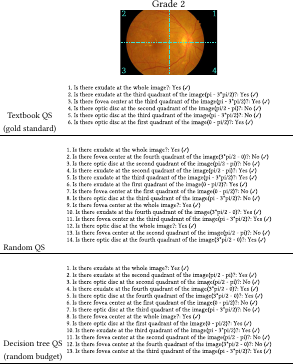}
	\caption{\small{Example of question stream from all questioning strategies, for a sample with grade 2 (part 1 of 2). \textbf{Groundtruth} answers are provided to the questions. \textbf{Extra-U-A} version}.}
	\label{fig:gt_mue_grade2_examples_extended}
\end{figure*}
\begin{figure*}[!h]
	\ContinuedFloat 
	\centering
	\includegraphics[width=.9\textwidth]{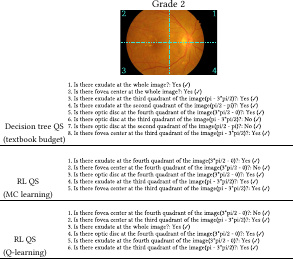}
	\caption{\small{Example of question stream from all questioning strategies, for a sample with grade 2 (part 2 of 2). \textbf{Groundtruth} answers are provided to the questions. \textbf{Extra-U-A} version}.}
\end{figure*}


\begin{figure*}[!h]
	\centering
	\includegraphics[width=.9\textwidth]{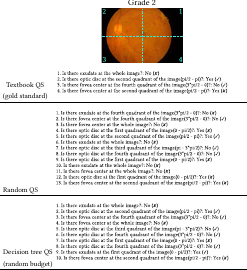}
	\caption{\small{Example of question stream from all questioning strategies, for a sample with true grade 1 (part 1 of 2). Answers from a \textbf{random} MuE with a total accuracy of $70\%$ are provided to the questions. \textbf{Extra-U-A} version.}}
	\label{fig:random_mue_07_grade1_examples_extended}
\end{figure*}
\begin{figure*}[!h]
	\ContinuedFloat 
	\centering
	\includegraphics[width=.9\textwidth]{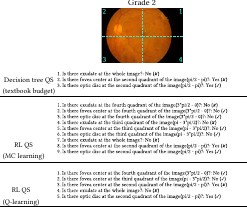}
	\caption{\small{Example of question stream from all questioning strategies, for a sample with true grade 1 (part 2 of 2). Answers from a \textbf{random} MuE with a total accuracy of $70\%$ are provided to the questions. \textbf{Extra-U-A} version.}}
\end{figure*}

\begin{figure*}[!h]
	\centering
	\includegraphics[width=.9\textwidth]{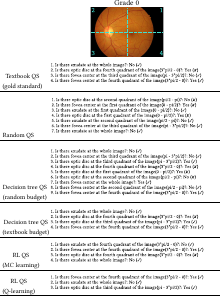}
	\caption{\small{Example of question stream from all questioning strategies, for a sample with true grade 0. Answers from a \textbf{reasonable} MuE with a total accuracy of $70\%$ are provided to the questions. \textbf{Extra-U-A} version.}}
	\label{fig:reasonable_mue_07_grade0_examples_extended}
\end{figure*}


\begin{figure*}[!h]
	\centering
	\includegraphics[width=.9\textwidth]{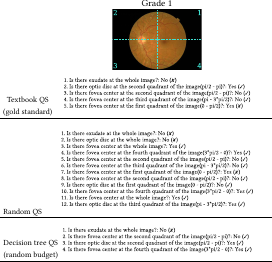}
	\caption{\small{Example of question stream from all questioning strategies, for a sample with true grade 1 (part 1 of 2). Answers from a \textbf{unreasonable} MuE with a total accuracy of $70\%$ are provided to the questions. \textbf{Extra-U-A} version.}}
	\label{fig:unreasonable_mue_07_grade1_examples_extended}
\end{figure*}
\begin{figure*}[!h]
	\ContinuedFloat 
	\centering
	\includegraphics[width=.9\textwidth]{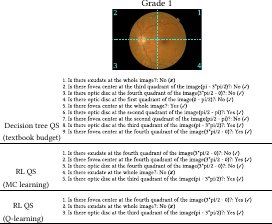}
	\caption{\small{Example of question stream from all questioning strategies, for a sample with true grade 1 (part 2 of 2). Answers from a \textbf{unreasonable} MuE with a total accuracy of $70\%$ are provided to the questions. \textbf{Extra-U-A} version.}}
\end{figure*}

\FloatBarrier
\clearpage
\section{Average reward over test set for different MuEs}
\label{app:rewards_per_MuE}
Below we present in tables the average reward for the test set, for different \gls{mue}s in term of accuracy (accuracy refers to the rate of correct questions over all possible questions) and behavior. Note that the trained questioning strategies (DT-RB, DT-TB, MC learning and Q-learning) are all trained with the groundtruth answers. This way, the strategies are exactly the same in all the tables below. 

The grade refers to the true fundus image \gls{dme} grade, and not the one assumed by the provided answers. For some cases, we see that the {\bf reasonable} \gls{mue} may have a lower average reward than a  {\bf random}, or an {\bf unreasonable} one, for the separate grades. This can occur when wrong answers for important questions lead to a history sequence $\bm{H}$ that implies an incorrect \gls{dme} grade. For example, grade 1 requires a higher number of questions for diagnosis. For a {\bf random} or {\bf unreasonable} \gls{mue}, if many questions are wrong (as compared to a {\bf reasonable} \gls{mue}), there are higher chances that a true DME grade of 1 gets misdiagnosed. hat we expect to see is that for the reasonable \gls{mue} the rewards are closer to the ones of the ``always correct'' \gls{mue} for the separate grades, which is most of the times indeed the case.

\vfill
\begin{table*}[!h]
	\small
	\begin{center}
		\caption{\label{table:rewards_random_MuE_06}Average reward over test set (in brackets \textbf{[]} is the average number of questions needed to achieve diagnosis). We consider that a \textbf{random} MuE with total accuracy $60\%$ answers the questions. In bold we highlight the best performance among all QS excluding the gold standard one.}
		\begin{tabular}{lrrrrr}
			\toprule
			\textbf{QS} & \textbf{Grade 0} & \textbf{Grade 1} & \textbf{Grade 2} && \textbf{Total} \\
			\toprule
			\emph{simple-A} &&&&&  \\
			\textbf{Textbook (gold standard)} & 0.826 [2.33] & 0.737 [3.33] & 0.523 [5.25] && 0.664 [3.9] \\ 
			\cmidrule{1-1}
			\textbf{Random} & 0.201 [9.27] & 0.129 [10.33] & 0.073 [12.17] && 0.131 [10.82]\\  
			\textbf{DT-RB} & 0.753 [4] & 0.554 [7.33] & 0.396 [9.28] && 0.558 [6.9]\\
			\textbf{DT-TB} & 0.017 [11.6] & 0.1 [11.33] & 0.005 [12.83] && 0.014 [12.23]\\
			\textbf{RL (MC learning)} & 0.780 [3.2] & \textbf{0.803 [2.33]} & 0.501 [6.61] && 0.635 [4.94] \\
			\textbf{RL (Q-learning)} & \textbf{0.799 [2.77]} & \textbf{0.803 [2.33]} & \textbf{0.518 [6.19]} && \textbf{0.653 [4.54]}\\ 
			&&&&& \\
			\emph{extra-U-A} &&&&&  \\
			\textbf{Textbook (gold standard)} & 0.214 [7.23] & 0.314 [6.37] & 0.2 [7.67] && 0.211 [7.42] \\ 
			\cmidrule{1-1}
			\textbf{Random} & 0.056 [11.85] & 0.114 [10.73] & 0.036 [12.38] && 0.048 [12.08]\\  
			\textbf{DT-RB} & 0.079 [10.86] & 0.104 [10.6] & 0.082 [11.35] && 0.082 [11.1]\\
			\textbf{DT-TB} & 0.043 [11.65] & 0.077 [11.47] & -0.012 [12.69] && 0.016 [12.09]\\
			\textbf{RL (MC learning)} & 0.246 [8.39] & \textbf{0.26 [8.23]} & \textbf{0.24 [8.75]} && 0.243 [8.57]\\
			\textbf{RL (Q-learning)} & \textbf{0.272 [8.03]} & 0.252 [8.13] & 0.233 [8.7] && \textbf{0.251 [8.38]}\\ 
			\bottomrule
		\end{tabular}
	\end{center}
\end{table*}
\vfill

\begin{table*}[!h]
	\small
	\begin{center}
		\caption{\label{table:rewards_reasonable_MuE_06}Average reward over test set (in brackets \textbf{[]} is the average number of questions needed to achieve diagnosis). We consider that a \textbf{reasonable} MuE with total accuracy $60\%$ answers the questions. In bold we highlight the best performance among all QS excluding the gold standard one.}
		\begin{tabular}{lrrrrr}
			\toprule
			\textbf{QS} & \textbf{Grade 0} & \textbf{Grade 1} & \textbf{Grade 2} && \textbf{Total} \\
			\toprule
			\emph{simple-A} &&&&&  \\
			\textbf{Textbook (gold standard)} & 0.954 [1.37] & 0.465 [6] & 0.341 [6.52] && 0.613 [4.26] \\ 
			\cmidrule{1-1}
			\textbf{Random} & 0.332 [7.13] & -0.086 [13.67] & 0.133 [10.31] && 0.21 [9.07] \\  
			\textbf{DT-RB} & \textbf{0.942 [1.63]}& 0.385 [8.67] & 0.166 [10.08] && 0.513 [6.35]\\
			\textbf{DT-TB} & 0.219 [8.6] & -0.326 [18.33] & 0.098 [11.44] && 0.132 [10.51]\\
			\textbf{RL (MC learning)} & 0.93 [1.97] & 0.306 [10.33] & 0.311 [7.27] && 0.58 [5.1]\\
			\textbf{RL (Q-learning)} & 0.931 [1.93] & \textbf{0.412 [7.67]} & \textbf{0.356 [6.39]} && \textbf{0.608 [4.51]}\\ 
			&&&&& \\
			\emph{extra-U-A} &&&&&  \\
			\textbf{Textbook (gold standard)} & 0.474 [4.57] & 0.238 [7.63] & 0.326 [6.41] && 0.387 [5.66] \\ 
			\cmidrule{1-1}
			\textbf{Random} & 0.196 [8.95] & 0.041 [12.73] & 0.158 [9.94] && 0.17 [9.63]\\  
			\textbf{DT-RB} & 0.34 [6.5] & 0.035 [12.63] & 0.158 [9.94] && 0.232 [8.56]\\
			\textbf{DT-TB} & 0.344 [6.42] & -0.11 [15.23] & 0.12 [11.11] && 0.207 [9.25]\\
			\textbf{RL (MC learning)} & 0.367 [5.96] & 0.111 [11.37] & \textbf{0.341 [6.78]} && 0.342 [6.62]\\
			\textbf{RL (Q-learning)} & \textbf{0.472 [4.71]} & \textbf{0.133 [10.73]} & 0.297 [7.23] && \textbf{0.366 [6.29]} \\ 
			\bottomrule
		\end{tabular}
	\end{center}
\end{table*}

\begin{table*}[!h]
	\small
	\begin{center}
		\caption{\label{table:rewards_unreasonable_MuE_06}Average reward over test set (in brackets \textbf{[]} is the average number of questions needed to achieve diagnosis). We consider that an \textbf{unreasonable} MuE with total accuracy $60\%$ answers the questions. In bold we highlight the best performance among all QS excluding the gold standard one.}
		\begin{tabular}{lrrrrr}
			\toprule
			\textbf{QS} & \textbf{Grade 0} & \textbf{Grade 1} & \textbf{Grade 2} && \textbf{Total} \\
			\toprule
			\emph{simple-A} &&&&&  \\
			\textbf{Textbook (gold standard)} & 0.29 [7.37] & 0.754 [3] & 0.7 [4.03] && 0.524 [5.44]  \\ 
			\cmidrule{1-1}
			\textbf{Random} & 0.124 [11.07] & -0.378 [19] & 0.175 [10.5] && 0.129 [11.12]\\  
			\textbf{DT-RB} & 0.13 [11.47] & 0.586 [7.33] & 0.618 [6.08] && 0.404 [8.48]\\
			\textbf{DT-TB} & 0.11 [11.23] & -0.215 [16] & -0.082 [12.92] && -0.004 [12.32] \\
			\textbf{RL (MC learning)} & 0.261 [8.33] & 0.711 [4] & 0.657 [5.25] && 0.487 [6.54]\\
			\textbf{RL (Q-learning)} & \textbf{0.293 [7.43]} & \textbf{0.737 [3.33]} & \textbf{0.688 [4.58]} && \textbf{0.518 [5.77]}\\ 
			&&&&& \\
			\emph{extra-U-A} &&&&&  \\
			\textbf{Textbook (gold standard)} & 0.125 [8.61] & 0.387 [5.67] & -0.07 [10.36] && 0.035 [9.39] \\ 
			\cmidrule{1-1}
			\textbf{Random} & -0.063 [14.17] & 0.095 [10.97] & -0.113 [14.85] && -0.082 [14.38]\\  
			\textbf{DT-RB} & 0.006 [12.72] & 0.128 [9.83] & -0.154 [14.61] && -0.072 [13.58]\\
			\textbf{DT-TB} & -0.076 [13.91] & 0.136 [9.57] & -0.177 [14.77] && -0.119 [14.17]\\
			\textbf{RL (MC learning)} & \textbf{0.177 [9.78]} & 0.298 [7.23] & \textbf{0.086 [11.64]} && \textbf{0.135 [10.64]} \\
			\textbf{RL (Q-learning)} & 0.157 [10.22] & \textbf{0.37 [6.13]} & 0.054 [11.64] && 0.113 [10.78]\\ 
			\bottomrule
		\end{tabular}
	\end{center}
\end{table*}

\begin{table}[!t]
	\small
	\begin{center}
		\caption{\label{table:rewards_random_MuE_07} 
			Average reward over the test set (\textbf{[]} show the average number of questions needed to achieve diagnosis). We consider that a \textbf{random} MuE with total accuracy $70\%$ answers the questions. In bold we highlight the best performance among all QS excluding the gold standard one.}
		\begin{tabular}{lrrrrr} %
			\toprule
			\textbf{QS} & \textbf{Grade 0} & \textbf{Grade 1} & \textbf{Grade 2} && \textbf{Total} \\
			\toprule
			\emph{simple-A} &&&&&  \\
			\textbf{Textbook} (gold standard) & 0.81 [2.5] & 0.27 [7.3] & 0.47 [5.7] && 0.61 [4.4] \\ 
			\cmidrule{1-1}
			\textbf{Random} & 0.25 [8.3] & 0.08 [12.3] & 0.06 [11.9] && 0.14 [10.4]\\  
			\textbf{DT-RB} & 0.75 [4.3] & 0.09 [12.3] & 0.38 [8.4] && 0.53 [6.8]\\
			\textbf{DT-TB} & 0.74 [4.4] & 0.08 [12.7] & 0.36 [9.2] && 0.51 [7.3]\\
			\textbf{RL (MC)} & 0.77 [3.6] & 0.11 [11] & 0.48 [6.4] && 0.59 [5.4]\\
			\textbf{RL (Q)} & \textbf{0.79 [3.2]} & \textbf{0.12 [11]} & \textbf{0.50 [5.9]} &&  \textbf{0.60 [4.9]}\\ 
			&&&&& \\
			\emph{extra-U-A} &&&&&  \\
			\textbf{Textbook} (gold standard) & 0.21 [7.6] & 0.33 [6] & 0.24 [7.9] && 0.23 [7.7] \\ 
			\cmidrule{1-1}
			\textbf{Random} & 0.04 [12.5] & 0.13 [10.3] & 0.02 [12.5] && 0.03 [12.4]\\  
			\textbf{DT-RB} & 0.05 [11.6] & -0.01 [13] & 0.03 [12.8] && 0.04 [12.3]\\
			\textbf{DT-TB} & -0.03 [12.9] & -0.11 [14.3] & 0 [13.2] && -0.02 [13.1]\\
			\textbf{RL (MC)} & 0.24 [8.8] & 0.26 [8] & 0.16 [10.1] && 0.20 [9.5] \\
			\textbf{RL (Q)} & \textbf{0.28 [8.4]} & \textbf{0.27 [7.7]} & \textbf{0.17 [10.1]} && \textbf{0.22 [9.3]}\\ 
			\bottomrule
		\end{tabular}
\end{center}
\end{table}

\begin{table}[!h]
\small
\begin{center}
	\caption{\label{table:rewards_reasonable_MuE_07} 
		Average reward over the test set (\textbf{[]} show the average number of questions needed to achieve diagnosis). We consider that a \textbf{reasonable} MuE with total accuracy $70\%$ answers the questions. In bold we highlight the best performance among all QS excluding the gold standard one.
	}
		\begin{tabular}{lrrrrr}
			\toprule
			\textbf{QS} & \textbf{Grade 0} & \textbf{Grade 1} & \textbf{Grade 2} && \textbf{Total} \\
			\toprule
			\emph{simple-A} &&&&&  \\
			\textbf{Textbook} (gold standard) & 0.98 [1.1] & 0.31 [6.3] & 0.29 [7] && 0.59 [4.4]\\ 
			\cmidrule{1-1}
			\textbf{Random} & 0.32 [7.1] & 0.05 [12.3] & 0.14 [10.4] && 0.22 [9.1]\\  
			\textbf{DT-RB} & 0.97 [1.4] & -0.18 [16] & 0.13 [11.1] && 0.48 [7.1]\\
			\textbf{DT-TB} & 0.97 [1.4] & -0.03 [12.7] & 0.13 [11.1] && 0.49 [6.9]\\
			\textbf{RL (MC)} & 0.97 [1.3] & 0.09 [11.7] & 0.27 [7.8] && 0.57 [5.2]\\
			\textbf{RL (Q)} & \textbf{0.97 [1.3]} & \textbf{0.11 [11]} & \textbf{0.33 [6.9]} && \textbf{0.60 [4.6]} \\ 
			&&&&& \\
			\emph{extra-U-A} &&&&&  \\
			\textbf{Textbook} (gold standard) & 0.41 [5.1] & 0.18 [8.2] & 0.32 [6.6] && 0.35 [6] \\ 
			\cmidrule{1-1}
			\textbf{Random} & 0.17 [9.6] & -0.04 [14.1] & 0.16 [10.2] && 0.15 [10.1]\\  
			\textbf{DT-RB} & 0.30 [7] & 0.06 [12.2] & 0.14 [10.5] && 0.21 [9]\\
			\textbf{DT-TB} & 0.30 [7.2] & -0.10 [15.1] & 0.12 [11.1] && 0.19 [9.5]\\
			\textbf{RL (MC)} & 0.44[5.3] & \textbf{0.09 [11.6]} & \textbf{0.31 [7]} && 0.35 [6.5]\\
			\textbf{RL (Q)} & \textbf{0.45 [5.1]} & \textbf{0.09 [11.6]} & \textbf{0.31 [7.1]} && \textbf{0.36 [6.5]}\\ 
			\bottomrule
		\end{tabular}
\end{center}
\end{table}

\begin{table}[!h]
\small
\begin{center}
	\caption{\label{table:rewards_unreasonable_MuE_07} 
		Average reward over the test set (\textbf{[]} show the average number of questions needed to achieve diagnosis). We consider that a \textbf{unreasonable} MuE with total accuracy $70\%$ answers the questions. In bold we highlight the best performance among all QS excluding the gold standard one.
	}
		\begin{tabular}{lrrrrr}
			\toprule
			\textbf{QS} & \textbf{Grade 0} & \textbf{Grade 1} & \textbf{Grade 2} && \textbf{Total} \\
			\toprule
			\emph{simple-A} &&&&&  \\
			\textbf{Textbook} (gold standard) &0.39 [6.3] & 0.74 [3.3] & 0.67 [4.1] && 0.55 [5]\\ 
			\cmidrule{1-1}
			\textbf{Random} & 0.07 [12.2] & 0.04 [10.3] & 0.01 [13.1] && 0.04 [12.6]\\  
			\textbf{DT-RB} & 0.19 [10.8] & 0.69 [5] & 0.54 [6.9] && 0.39 [8.5]\\
			\textbf{DT-TB} & 0.21 [10.5] & 0.69 [5] & 0.54 [6.7] && 0.40 [8.3]\\
			\textbf{RL (MC)} & 0.32 [8] & \textbf{0.71 [4]} & 0.62 [5.5] && 0.50 [6.5]\\
			\textbf{RL (Q)} & \textbf{0.34 [7.5]} & \textbf{0.71 [4]} & \textbf{0.64 [5.1]} && \textbf{0.51 [6.1]}\\ 
			&&&&& \\
			\emph{extra-U-A} &&&&&  \\
			\textbf{Textbook} (gold standard) & 0.12 [8.5] & 0.31 [6.7] & 0.09 [9.4] && 0.11 [8.9]\\ 
			\cmidrule{1-1}
			\textbf{Random} & -0.03 [13.6] & 0.07 [11.6] & -0.05 [13.8] && -0.03 [13.6]\\  
			\textbf{DT-RB} & 0.01 [12.4] & 0.09 [11.4] & -0.04 [13.6] && -0.01 [13] \\
			\textbf{DT-TB} & -0.04 [13] & 0.07 [11.6] & -0.09 [14.3] && -0.06 [13.6]\\
			\textbf{RL (MC)} & 0.16 [9.9] & \textbf{0.27 [8.1]} & \textbf{0.13 [10.6]} && 0.15 [10.2]\\
			\textbf{RL (Q)} & \textbf{0.18 [9.7]} & \textbf{0.27 [8.2]} & \textbf{0.13 [10.6]} && \textbf{0.16 [10.1]}\\ 
			\bottomrule
		\end{tabular}
\end{center}
\end{table}

\begin{table*}[!h]
	\small
	\begin{center}
		\caption{Average reward over test set (in brackets \textbf{[]} is the average number of questions needed to achieve diagnosis). We consider that a \textbf{random} MuE with total accuracy $90\%$ answers the questions. In bold we highlight the best performance among all QS excluding the gold standard one.}
		\begin{tabular}{lrrrrr}
			\toprule
			\textbf{QS} & \textbf{Grade 0} & \textbf{Grade 1} & \textbf{Grade 2} && \textbf{Total} \\
			\toprule
			\emph{simple-A} &&&&&  \\
			\textbf{Textbook (gold standard)} & 0.927 [1.57] & 0.284 [7.27] & 0.299 [7.46] && 0.571 [4.89] \\ 
			\cmidrule{1-1}
			\textbf{Random} & 0.271 [8.27] & 0.090 [11.9] & 0.125 [10.99] && 0.187 [9.85]\\  
			\textbf{DT-RB} & 0.901 [2.23] & 0.091 [12.8] & 0.134 [11.68] && 0.466 [7.62]\\
			\textbf{DT-TB} & 0.901 [2.25] & 0.127 [11.97] & 0.136 [11.71] && 0.468 [7.61]\\
			\textbf{RL (MC learning)} & 0.911 [1.94] & 0.147 [11.07] & 0.313 [7.81] && 0.566 [5.4]\\
			\textbf{RL (Q-learning)} & \textbf{0.913 [1.86]} & \textbf{0.175 [10.07]} & \textbf{0.335 [7.29]} && \textbf{0.580 [5.05]}\\ 
			&&&&& \\
			\emph{extra-U-A} &&&&&  \\
			\textbf{Textbook (gold standard)} & 0.273 [6.72] & 0.249 [7.63] & 0.241 [7.91] && 0.255 [7.38] \\ 
			\cmidrule{1-1}
			\textbf{Random} & 0.106 [11.08] & 0.069 [12.23] & 0.106 [11.27] && 0.104 [11.23]\\  
			\textbf{DT-RB} & 0.205 [8.87] & 0.126 [11] & 0.119 [11.05] && 0.157 [10.1]\\
			\textbf{DT-TB} & 0.196 [9.07] & 0.045 [12.93] & 0.057 [12.57] && 0.117 [11.06]\\
			\textbf{RL (MC learning)} & 0.249 [7.99] & 0.128 [11.03] & \textbf{0.283 [7.83]} && 0.261 [8.04]\\
			\textbf{RL (Q-learning)} & \textbf{0.355 [6.49]} & \textbf{0.158 [10.2]} & 0.272 [8.03] && \textbf{0.303 [7.45]}\\ 
			\bottomrule
		\end{tabular}
	\end{center}
\end{table*}

\begin{table*}[!h]
	\small
	\begin{center}
		\caption{Average reward over test set (in brackets \textbf{[]} is the average number of questions needed to achieve diagnosis). We consider that a \textbf{reasonable} MuE with total accuracy $90\%$ answers the questions. In bold we highlight the best performance among all QS excluding the gold standard one.}
		\begin{tabular}{lrrrrr}
			\toprule
			\textbf{QS} & \textbf{Grade 0} & \textbf{Grade 1} & \textbf{Grade 2} && \textbf{Total} \\
			\toprule
			\emph{simple-A} &&&&&  \\
			\textbf{Textbook (gold standard)} & 0.976 [1.16] & 0.311 [7.17] & 0.307 [7.19] && 0.598 [4.57] \\ 
			\cmidrule{1-1}
			\textbf{Random} & 0.337 [7.28] & 0.072 [11.97] & 0.129 [10.71] && 0.217 [9.27]\\  
			\textbf{DT-RB} & 0.964 [1.44] & 0.145 [11.63] & 0.143 [11.13] && 0.5 [6.94]\\
			\textbf{DT-TB} & 0.964 [1.43] & 0.149 [11.7] & 0.139 [11.23] && 0.498 [6.99]\\
			\textbf{RL (MC learning)} & 0.966 [1.36] & 0.207 [9.93] & 0.323 [7.31] && 0.598 [4.84]\\
			\textbf{RL (Q-learning)} & \textbf{0.968 [1.31]} & \textbf{0.234 [9.17]} & \textbf{0.346 [6.75]} && \textbf{0.612 [4.49]}\\ 
			&&&&& \\
			\emph{extra-U-A} &&&&&  \\
			\textbf{Textbook (gold standard)} & 0.311 [6.16] & 0.172 [8.27] & 0.274 [7.4] && 0.286 [6.9] \\ 
			\cmidrule{1-1}
			\textbf{Random} & 0.154 [10.07] & 0.066 [12.53] & 0.125 [10.9] && 0.135 [10.61]\\  
			\textbf{DT-RB} & 0.223 [8.32] & 0.060 [12.63] & 0.149 [10.35] && 0.178 [9.56] \\
			\textbf{DT-TB} & 0.218 [8.49] & -0.032 [14.73] & 0.101 [11.66] && 0.146 [10.42]\\
			\textbf{RL (MC learning)} & 0.266 [7.59] & 0.065 [12.33] & \textbf{0.329 [6.9]} && 0.29 [7.44]\\
			\textbf{RL (Q-learning)} & \textbf{0.379 [6.01]} & \textbf{0.087 [11.37]} & 0.307 [7.16] && \textbf{0.329 [6.84]} \\ 
			\bottomrule
		\end{tabular}
	\end{center}
\end{table*}

\begin{table*}[!h]
	\small
	\begin{center}
		\caption{Average reward over test set (in brackets \textbf{[]} is the average number of questions needed to achieve diagnosis). We consider that an \textbf{unreasonable} MuE with total accuracy $90\%$ answers the questions. In bold we highlight the best performance among all QS excluding the gold standard one.}
		\begin{tabular}{lrrrrr}
			\toprule
			\textbf{QS} & \textbf{Grade 0} & \textbf{Grade 1} & \textbf{Grade 2} && \textbf{Total} \\
			\toprule
			\emph{simple-A} &&&&&  \\
			\textbf{Textbook (gold standard)} & 0.856 [2.17] & 0.527 [4.9] & 0.312 [7.41] && 0.558 [5.02] \\ 
			\cmidrule{1-1}
			\textbf{Random} & 0.274 [8.64] & 0.231 [9.63] & 0.109 [11.4] && 0.186 [10.12]\\  
			\textbf{DT-RB} & 0.811 [3.34] & 0.396 [8.17] & 0.142 [11.64] && 0.444 [7.88]\\
			\textbf{DT-TB} & 0.812 [3.31] & 0.404 [8.2] & 0.143 [11.73] && 0.446 [7.92]\\
			\textbf{RL (MC learning)} & 0.823 [2.92] & 0.461 [6.6] & 0.314 [7.9] && 0.542 [5.68]\\
			\textbf{RL (Q-learning)} & \textbf{0.829 [2.75]} & \textbf{0.474 [6.23]} & \textbf{0.344 [7.31]} && \textbf{0.560 [5.28]}\\ 
			&&&&& \\
			\emph{extra-U-A} &&&&&  \\
			\textbf{Textbook (gold standard)} & 0.24 [6.97] & 0.257 [7.47] & 0.189 [8.81] && 0.214 [7.95] \\ 
			\cmidrule{1-1}
			\textbf{Random} & 0.101 [11.05] & 0.082 [11.73] & 0.073 [12.11] && 0.085 [11.63] \\  
			\textbf{DT-RB} & 0.16 [9.4] & 0.109 [11.1] & 0.111 [11.42] && 0.132 [10.53]\\
			\textbf{DT-TB} & 0.157 [9.52] & 0.075 [12.1] & 0.043 [13.05] && 0.094 [11.47] \\
			\textbf{RL (MC learning)} & 0.252 [7.98] & 0.16 [10.3] & \textbf{0.248 [8.53]} && 0.246 [8.37]\\
			\textbf{RL (Q-learning)} & \textbf{0.348 [6.61]} & \textbf{0.202 [9.2]} & 0.233 [8.81] && \textbf{0.282 [7.87]} \\ 
			\bottomrule
		\end{tabular}
	\end{center}
\end{table*}
\FloatBarrier
\clearpage

\end{document}